\documentclass{article}

\usepackage[main, final]{neurips_2026}

\makeatletter
\renewcommand{\@notice}{}
\makeatother

\usepackage[utf8]{inputenc} 
\usepackage[T1]{fontenc}    
\usepackage{hyperref}       
\usepackage{url}            
\usepackage{booktabs}       
\usepackage{amsfonts}       
\usepackage{nicefrac}       
\usepackage{microtype}     
\usepackage{xcolor}         

\usepackage{amsmath}
\usepackage{wrapfig}
\usepackage{amssymb}
\usepackage{mathtools}
\usepackage{amsthm}
\usepackage{adjustbox}
\usepackage{listings}
\usepackage{xcolor}
\usepackage{colortbl}
\usepackage{booktabs}
\usepackage{multirow}
\usepackage{makecell}
\usepackage{graphicx}
\usepackage{hyperref}
\usepackage{url}
\usepackage{hyperref}
\usepackage{url}
\usepackage{booktabs, multirow, array}
\usepackage{xcolor}
\usepackage{colortbl}
\usepackage{graphicx}
\usepackage{booktabs}
\usepackage{tabularx}
\usepackage{multirow}
\usepackage{threeparttable}
\usepackage{array}
\usepackage{booktabs}
\usepackage{multirow}
\usepackage{makecell}
\usepackage{comment}
\usepackage{makecell}
\usepackage{placeins}
\usepackage{listings}
\usepackage{xcolor}
\usepackage{enumitem}

\usepackage{listings}
\usepackage{xcolor}

\lstdefinestyle{compactpython}{
    language=Python,
    basicstyle=\ttfamily\footnotesize,
    breaklines=true,
    columns=fullflexible,
    keepspaces=true,
    showstringspaces=false,
    frame=single
}

\usepackage[utf8]{inputenc} 
\usepackage[T1]{fontenc}    
\usepackage{hyperref}       
\usepackage{url}            
\usepackage{booktabs}      
\usepackage{amsfonts}      
\usepackage{nicefrac}      
\usepackage{microtype}      
\usepackage{xcolor}         
\usepackage{placeins}
\usepackage[table]{xcolor}
\usepackage{graphicx}
\usepackage{amsmath,amssymb}   
\usepackage{graphicx}        
\usepackage{xspace}
\usepackage{float}
\usepackage{xcolor}         
\usepackage{bbm}
\usepackage{algpseudocode}
\usepackage{titletoc}

\usepackage{microtype}
\usepackage[ruled,vlined]{algorithm2e}
\usepackage{enumitem}
\usepackage{threeparttable}
\usepackage{tabularx}
\usepackage{listings}

\theoremstyle{plain}

\theoremstyle{definition}

\theoremstyle{remark}

\newcommand{\merge}{\mathcal{M}_\phi}
\usepackage[textsize=tiny]{todonotes}

\newcommand{\namem}{UMIM}

\usepackage{multibib}
\newcites{appendix}{Appendix References}

\title{Distilling Sequential Computation\\in Transformer Language Models}

\author{
Zixuan Lan\textsuperscript{1} \quad
Jessica Yang\textsuperscript{2} \quad
Yanhong Li\textsuperscript{3} \quad
Karen Livescu\textsuperscript{2} \quad
Jiawei Zhou\textsuperscript{4}
\\[0.6em]
\textsuperscript{1}The University of Chicago \quad
\textsuperscript{2}Toyota Technological Institute at Chicago
\\
\textsuperscript{3}Independent Researcher \quad
\textsuperscript{4}Stony Brook University
\\[0.5em]
\texttt{zixuanlan@uchicago.edu} \quad
\texttt{jiaminy@ttic.edu} \quad
\texttt{klivescu@ttic.edu}
\\
\texttt{yanhong.lbh@gmail.com} \quad
\texttt{jiawei.zhou.1@stonybrook.edu}
}

\begin{document}

\maketitle

\begin{abstract}

Transformer language models process sequences token by token in an autoregressive manner, making growing contexts increasingly expensive. Yet many adjacent token spans are highly predictable or frequently occur as stable units, suggesting that their representations may be compressible. We introduce a method for distilling sequential computation by replacing spans of input tokens with collapsed representations, computed on the fly by a lightweight merge module. This module generates a single surrogate embedding from a sequence of static token embeddings that captures the functional role of the multiple tokens, allowing pretrained models to operate on compressed inputs without architectural changes or re-training. We apply this approach during inference to compress both prompts and intermediate decoding steps, using a rollback mechanism to substitute stored multi-token KV cache entries with their single-step surrogates. Experiments across diverse models show that the merge module can be used to reduce effective sequence length by up to 40\% with minimal accuracy degradation across language modeling evaluations and downstream tasks, including question answering, summarization, commonsense reasoning, and long-form mathematical reasoning. Additional lightweight adaptation of the merge module further improves the accuracy-compression trade-off in selected settings. These results demonstrate that sequential token computation in Transformers can be effectively approximated through condensed surrogate representations that approximate the original behavior without model updating. Code and project resources are available at
\url{https://github.com/Zesearch/Umim-LLM}.
\end{abstract}

\section{Introduction}
\label{sec:introduction}

Autoregressive large language models (LLMs), typically built on Transformer architectures \citep{vaswani2017attention}, process sequences token by token, making long inputs and growing autoregressive contexts expensive to compute and cache. Yet many adjacent token spans are highly predictable or frequently recurring in a sequence but are still processed as separate tokens. For example, ``machine learning'' or ``New York'' often appear together and are easy to predict once the first token is observed. 

Prior work has explored reducing sequential computation in Transformer language models \citep{lan2026reducedmatrixmultiplicationinputadaptive} through alternative tokenization, multi-token decoding, and context reduction. Tokenization-based methods modify the input units on which models are trained, including byte- or patch-level representations \citep{pagnoni2024byte, xue2022byt5, yu2023megabyte} and larger ``superword'' units \citep{liu2025superbpe}, but typically require re-training because tokenization defines the model's input structure. Other methods target pre-trained models through extended vocabularies, drafting models, or external chunk representations \citep{lan2023copy, leviathan2023fast, lichunk}, but introduce auxiliary components or inference overhead. Context-reduction methods prune or compress inputs, hidden states, or KV caches \citep{zhang2023h2o, bolya2023token, mu2023learning, kallini2024mrt5, jiang2023llmlingua, xiaoefficient, shao2024flexibly, li2025text}, but often rely on model-internal signals, custom compression objectives, or accuracy-efficiency trade-offs. It remains unclear whether multi-step Transformer computation can be replaced at inference time with single-step surrogates without accessing internal activations or updating the pretrained backbone.

In this work, we address the above question by distilling the sequential computation of multiple tokens into a single-step input representation for pre-trained language models. Given a pre-trained language model (LM) and a span of input tokens in context during generation, we aim to construct a \textit{single surrogate embedding} that replaces the original token span \textit{at the input level}, while preserving the model’s next-token distribution. 

To handle the large number of possible token spans during inference, we introduce \textbf{U}niversal \textbf{M}ulti-step \textbf{I}nput \textbf{M}erging (\namem), a shared, lightweight merge module that computes surrogate embeddings on demand. {\namem} takes the static embeddings of the tokens in a selected span and computes a single surrogate embedding. The resulting surrogate is span-specific but context-independent: For the same $n$-grams , the module produces the same surrogate wherever it appears, while different $n$-grams  naturally produce different surrogates.

We implement the {\namem} module as a lightweight, single-layer attention network that serves as a plug-in to any pre-trained Transformer language model, applied to input embeddings \textit{prior to model computation} at inference. {\namem} is trained using a predictive distillation loss, encouraging the model’s output distribution with merged input embeddings to match that of the original uncompressed token sequences. Crucially, the pre-trained language model remains \textit{frozen}, with only the external merge module updated during training. Once trained, the same {\namem} module can be applied across different downstream tasks without re-training, offering a general and reusable mechanism for input-level sequential computation compression. Building on this general module, downstream performance can be further improved through task-specific merge-rule construction and lightweight adaptation of the merge module, while keeping the backbone language model frozen.

Using the shared {\namem} module, we compress both static prompts and autoregressive decoding contexts, with rollback replacing multi-token KV-cache entries by single-step surrogate states. Experiments across diverse models and tasks show that {\namem} directly reduces effective sequence length by up to 40\% with minimal accuracy degradation, while preserving next-token predictive behavior.

\section{Related Work}
\label{sec:related-work}

\paragraph{Tokenization Variations and Extension}

Prior work explores alternative tokenization strategies for Transformer language models. Byte-level and tokenizer-free models such as ByT5 \citep{xue2022byt5}, Megabyte \citep{yu2023megabyte}, and ByteFusion \citep{pagnoni2024byte} reduce reliance on predefined vocabularies, while SuperBPE \citep{liu2025superbpe} introduces larger phrase-level units to reduce sequence length. Other methods extend token representations for span- or chunk-level decoding, including CoG \citep{lan2023copy}, CD-LM \citep{lichunk}, and adaptive hypertoken construction such as zip2zip \citep{geng2025zip2zipinferencetimeadaptivetokenization}. However, these approaches generally require modifying the tokenizer, extending the token space, introducing auxiliary retrieval or encoding components, or retraining the model. In contrast, our method constructs surrogate embeddings for token spans post-training, without altering the tokenizer, vocabulary, or model architecture.

\paragraph{Latent Context Compression}

\begin{comment}

\end{comment}
A closely related line of work compresses long contexts into learned continuous representations, including Gist Tokens \citep{mu2023learning}, ICAE \citep{gecontext}, Extensible Tokenization \citep{shao2024flexibly}, 500xCompressor \citep{li2024500xcompressorgeneralizedpromptcompression}, and multimodal processing of textual information via images \citep{li2025text}. These methods encode inputs into compact vectors or special-token representations that are later consumed by the language model. While effective, they typically require an additional encoding stage and treat compression as a separate interface. By contrast, our approach stays within the original token-based generation pipeline: It constructs surrogate embeddings for token spans directly at the input level and replaces multi-step computation with single-step surrogates during inference.
\paragraph{Prompt Compression and Context Reduction}

Prompt and context compression methods reduce inference-time computation through task-aware or task-agnostic content reduction. Task-aware methods such as LongLLMLingua \citep{jiang2024longllmlingua}, Recomp \citep{xu2024recomp}, and also CPC \citep{liskavets2024promptcompressioncontextawaresentence} retain content relevant to downstream questions or tasks. Task-agnostic methods, including ProCut \citep{xu2025procutllmpromptcompression}, EHPC \citep{fei2025efficientpromptcompressionevaluator}, LLMLingua \citep{jiang2023llmlingua}, LLMLingua2 \citep{pan2024llmlingua2datadistillationefficient}, and self-information filtering \citep{li2023unlocking}, estimate token or segment importance to remove less informative content. Other work studies abstractive prompt compression, where prompts are rewritten into shorter natural-language forms, as in Nano-Capsulator \citep{chuang2024learningcompresspromptnatural} and Cmprsr \citep{zakazov2026cmprsrabstractivetokenlevelquestionagnostic}. In contrast, our approach is also task-agnostic but does not delete, filter, or rewrite input content.

\paragraph{KV Cache Management, Token Pruning and Merging}

Recent work reduces inference memory and computation through KV cache management, token pruning, or token merging. KV cache methods either statically filter tokens during prefill or dynamically retain, evict, or offload tokens during decoding, as in H$_2$O \citep{zhang2023h2o}, StreamingLLM \citep{xiaoefficient}, KeyFormer \citep{adnan2024keyformer}, OmniKV \citep{hao2025omnikv}, PagedAttention \citep{kwon2023efficient}, and Blockwise Caching \citep{dao2022flashattention}. Other methods reduce context length through token pruning or merging, including MRT5 \citep{kallini2024mrt5} and hidden-state-based merging in language \citep{yuan2024efficient}, vision \citep{bolya2023token, shin2025atom}, video \citep{ryoo2021tokenlearner}. In contrast, our method performs merging before Transformer computation rather than by pruning existing KV entries or combining hidden states during model execution. It replaces multi-token spans with single surrogate input embeddings and substitutes their corresponding multi-step KV cache entries with single-step surrogate states, preserving the functional role of the original tokens while reducing both effective sequence length and KV cache usage.

\section{Methodology}
\label{sec:method}

We develop {\namem} as an external plug-in for pre-trained Transformer language models, which maps selected contiguous token spans to single surrogate embeddings computed from their static input embeddings. Formally, let $\mathbf{x} = (x_1, \ldots, x_T)$ be a sequence of input tokens, and let $e(\cdot)$ denote the token embedding lookup, applied element-wise so that $e(x_i)$ is a single embedding and $e(x_{a:b})$ denotes the corresponding sequence of embeddings. A Transformer language model with parameters $\theta$ processes the embedded prefix $e(x_{<t})$ and outputs the next-token distribution $p_\theta^t(\cdot \mid e(x_{<t}))$. We learn a merge module $\merge$, parameterized by $\phi$, that maps a span $x_{t:t+n}$ to a surrogate embedding $\tilde{e} = \merge(e(x_{t:t+n}))$, which replaces the original embeddings while preserving the subsequent predictive distribution. The following sections describe merge span selection, the merge module architecture, the distillation objective, and decoding algorithms for applying {\namem} during prompt compression and generation.

\subsection{Merge Rule}
\label{sec:merge_rule}

To determine which token spans to merge, we construct a frequency-based merging set from a general corpus $\mathcal{C}$. Specifically, we collect unique contiguous $n$-grams  of length $2 \le n \le 4$ whose frequency exceeds a threshold $\tau$, forming the merging set $\mathcal{R}$. High-frequency spans are likely to recur during inference, enabling more frequent compression, and also provide sufficient training instances for learning surrogate embeddings. The maximum of $n=4$ mitigates data sparsity while still allowing meaningful multi-token groupings. When multiple candidates overlap, we apply a longest-match strategy, giving precedence to the longest valid span at each position. During training and inference, spans that match entries in $\mathcal{R}$ are treated as merge candidates. The merge rule is independent of the model architecture and depends only on the tokenizer and corpus statistics. The corpus $\mathcal{C}$ used to construct $\mathcal{R}$ is task-independent, though in-domain data can further improve compression coverage by capturing domain-specific frequent spans. See Appendix~\ref{sec:Data Preparation} for details.

\subsection{Lightweight Merge Module Architecture}
\label{sec:merge_module}

\begin{figure*}[t]
  \centering
  \includegraphics[width=\textwidth]{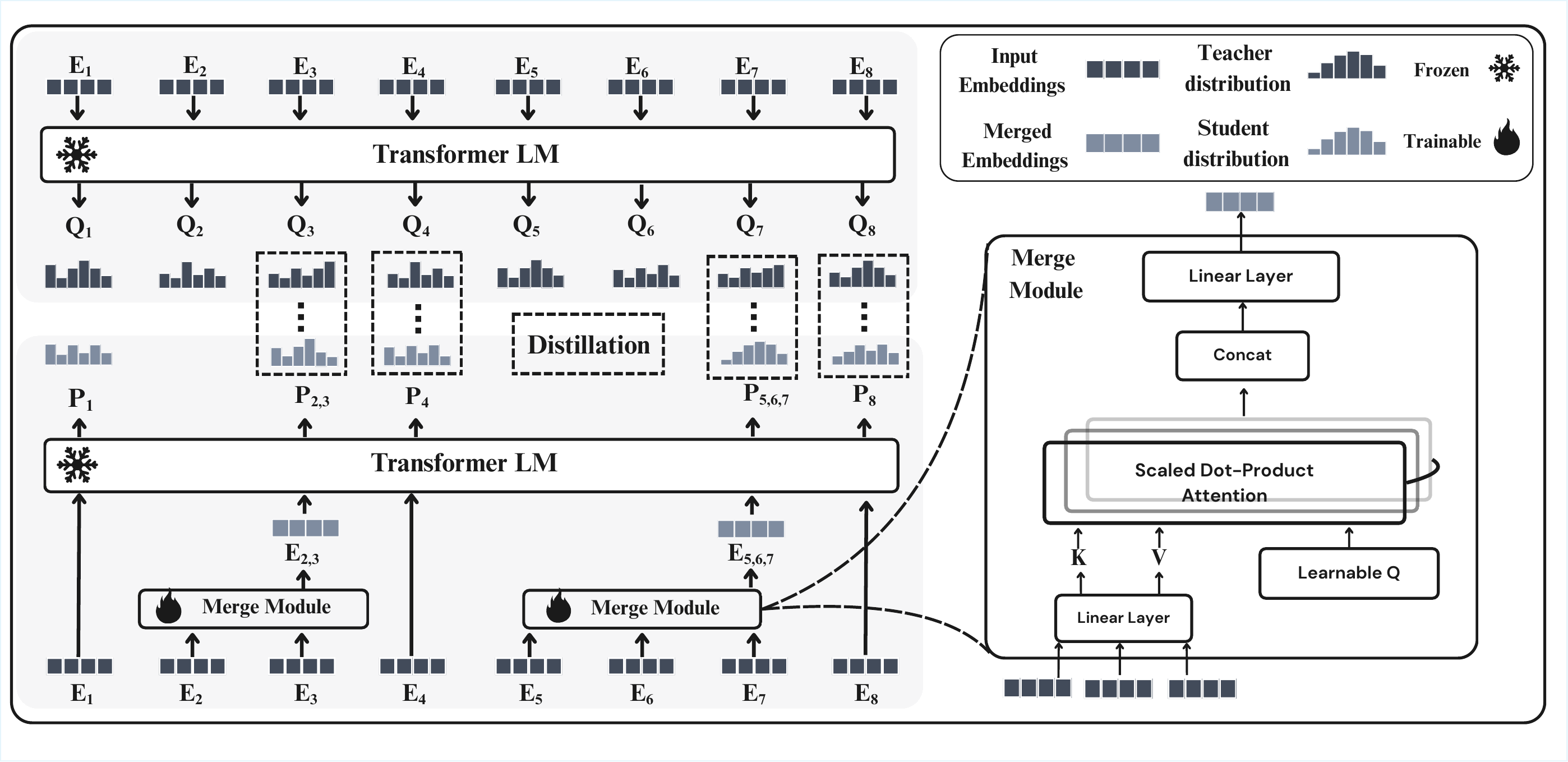}
  \vspace{-20pt}
  \caption{Training process of the {\namem} module, with token-step aligned distillation loss and merge module architecture.}
  \label{fig:Training}
\end{figure*}

The {\namem} is a lightweight single-layer multi-head attention pooling network that operates solely on static input embeddings, prior to any Transformer computation. It serves as a universal replacement for contiguous token spans, compressing variable-length inputs into a single surrogate embedding via a \textit{learnable query} mechanism.

Formally, let $d$ denote the embedding dimension of the target language model, and $h$ the number of attention heads. Each head has dimension $d_h = d / h$. Given the token span embeddings $e(x_{t:t+n}) \in \mathbb{R}^{d \times n}$, we use a learnable query parameter $Q \in \mathbb{R}^{h \times d_h}$, where the $i^\textrm{th}$ row $q_i \in \mathbb{R}^{d_h}$ is the query vector for head $i$. We use a shared projection $W^{kv}$ for span representations before attention pooling. After pooling, an output projection $W^o$ maps the concatenated head outputs back to the model embedding space. The output surrogate embedding is computed as:
\vspace{-.1in}

\vspace{-10pt}
\begin{equation}
M_\phi(e(x_{t:t+n})) =
W^o \left[
\mathrm{o}_1;\cdots;\mathrm{o}_h
\right]^\top,
\quad
\mathrm{o}_i =
\mathrm{softmax}\!\left(
\frac{q_i^\top W^{kv} e(x_{t:t+n})}{\sqrt{d_h}}
\right)
\left(W^{kv} e(x_{t:t+n})\right)^\top .
\end{equation}

\subsection{{\namem} Learning by Distilling Sequential computation}
\label{sec:training_phase}

We train the {\namem} to produce single-step surrogate embeddings that can replace multi-tokens, while preserving the language model’s next-token distributions. The language model parameters $\theta$ are frozen throughout training. To guide learning, we introduce a \textit{sequential computation distillation loss}, aligning predictions made on merged inputs with those from the original token-by-token computation. Let $\merge(e(x_{1:T}))=(e'_s)_{s\in \mathcal{S}}$ denote the compressed sequence of token embeddings after applying the merge module to all eligible token spans according to the merge rule $\mathcal{R}$. Here, $\mathcal{S}$ represents the set of \textit{retained token positions} after merging. For instance, if tokens $x_{t:t+n}$ are merged into a single surrogate embedding $\merge(e(x_{t:t+n}))$, then token steps $t$ through $t+n-1$ are collapsed, and only step $t+n-1$ is retained in $\mathcal{S}$ (e.g., $\mathcal{S} = (\cdots, t-1, t+n-1, \cdots)$).

To align the behavior of the compressed input with the original, we minimize the KL divergence between the output distributions of the language model on unmerged vs.~merged input prefixes:

\vspace{-15pt}
\begin{equation}
\mathcal{L}_{\text{distill}}
= D_{\mathrm{KL}}\!\left(
p_\theta(\cdot \mid e(x_{\le s})) \;\|\; p_\theta(\cdot \mid M_\phi(e(x_{\le s})))
\right)
\end{equation}

This loss encourages the merged representation to produce predictive distributions that match those from the full token sequence. An illustration of the training procedure is shown in Figure~\ref{fig:Training}.

\subsection{Model Inference with Runtime Merging}
\label{sec:decoding_phase}

\begin{figure*}[t]
  \centering
  \includegraphics[width=\textwidth]{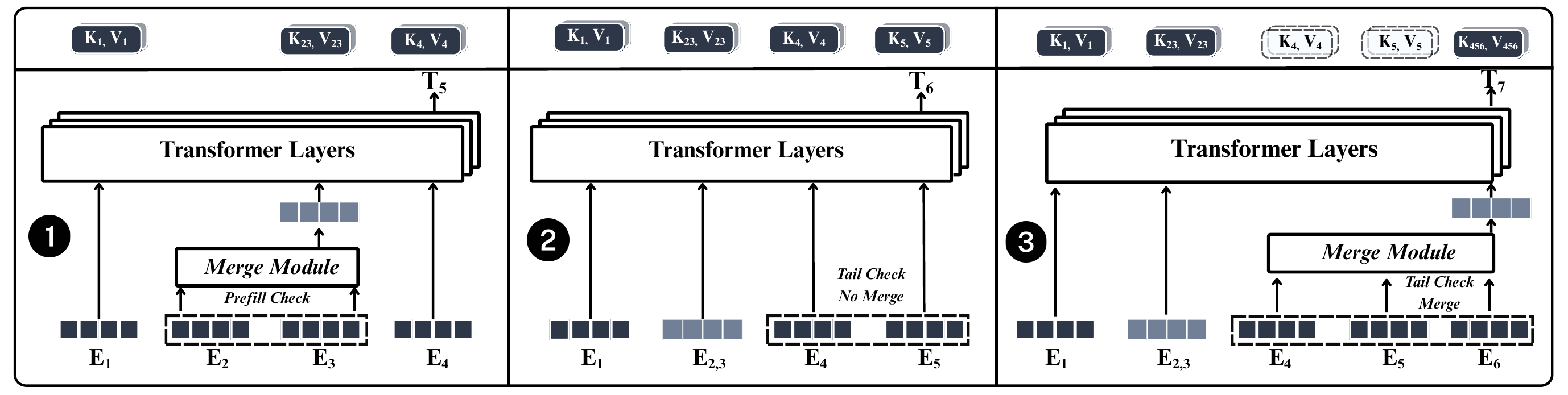}
  \vspace{-14pt}
  \caption{{\namem} Decoding Process: \textbf{E} is embedding, \textbf{T} is token, \textbf{K,V} is KV cache (1) During prefill, mergeable spans are replaced by surrogate embeddings. (2) During autoregressive decoding, each new token is checked against the recent suffix; if no valid merge is found, decoding proceeds normally by appending a new KV cache. (3) If the suffix forms a mergeable span, {\namem} rolls back the matched span and replaces its multiple KV entries with a single merged KV cache.}
  \label{fig:fig-decoding}
\end{figure*}

Once trained, the {\namem} module can be integrated into inference with the compatible pre-trained language model. During inference, {\namem} reduces the effective sequence length by applying runtime merging in both prompt prefilling and autoregressive decoding as shown in Figure~\ref{fig:fig-decoding}. Let $z = (z_1, \cdots, z_m)$ denote the prompt tokens, and $y = (y_1, y_2, \cdots)$ the generated tokens. During prefill, we scan the prompt for eligible $n$-grams in $\mathcal{R}$ and replace each matched span with its surrogate embedding. During decoding, after each new token is generated, we check whether the recent suffix forms an eligible $n$-gram; if so, we roll back the corresponding KV entries and replace them with a single merged KV state. The checking and rollback overhead is amortized over long sequences, while the reduced sequence length directly lowers KV-cache size and attention computation.

By default, we apply the merge rule underlying the training data $\mathcal{R}$ for inference. This, however, can be relaxed by adapting the merge rule further to a specific downstream task based on in-domain data, facilitating a wider range of compression ratio and domain generalization. See Section~\ref{sec:Task_Specific} for details.

\section{Experiments}
\label{sec:experiments}

\subsection{Experimental Setup}
\label{sec:experimental_setup}

We train {\namem} modules for LMs including Llama 3.1 8B \cite{grattafiori2024llama3herdmodels}, Llama 3.2 1.5B, and GPT2-XL 1.5B \cite{radford2019language} using the WikiText-103 \citep{merity2016pointersentinelmixturemodels} corpus, and DeepScaleR-1.5B-Preview\footnote{\url{https://huggingface.co/agentica-org/DeepScaleR-1.5B-Preview}.} using a long reasoning trajectory corpus we collected. Each merge module uses $h=4$ attention heads and contains 7M--50M learnable parameters, depending on the backbone size.
To build the merging rule $\mathcal R$, we extract $n \in \{2,3,4\}$-grams with threshold $\tau=5$, yielding over 2M unique token spans per model. More details can be found in Appendix~\ref{sec:Data Preparation}

We first measure the distribution alignment quality after applying the embedding merges compared with the original token-based LLMs. The metrics include: Top-1 accuracy---percentage of token positions after merging where the top-1 tokens in the predictive distribution exactly match, Top-3/10 overlap---average number of overlapped tokens in the top-3/10 tokens from corresponding distributions at aligned token positions, Top-p overlap---the overlap between tokens whose cumulative probability mass reaches p in each distribution, and Mean Reciprocal Rank (MRR)---the average reciprocal rank of the teacher (original LLM)'s top-1 token within the distribution after span merging at corresponding token positions. 

Further, we measure the language modeling performance using perplexity (PPL) with {\namem} on different test datasets (Sec.~\ref{sec:generation_eval_main}), including WikiText-103 which the module is trained on, as well as BookCorpus \citep{zhu2015aligning} and OpenWebText \citep{Gokaslan2019OpenWeb} with the same merge module without re-training.
To quantify compression, we report token reduction (TR, \%), which reflects the reduction in sequence length and KV cache usage.

Finally, we assess the \textbf{generalization} ability of {\namem} on downstream tasks including question answering (QA), document summarization, and math reasoning (Sec.~\ref{sec:downstream_tasks}). In the \textbf{task-agnostic} setting, a merge module trained on WikiText-103 is directly applied to all downstream tasks without task-specific supervision or fine-tuning, evaluating its transfer across unseen domains. We further study a \textbf{downstream adaptation} setting, where the pretrained merge module is specialized to target tasks through task-specific merge-rule construction, fine-tuning, and preference optimization, while keeping the backbone language model frozen. We compare with representative context reduction and prompt compression baselines, including Select Context \citep{li2023compressing}, Llmlingua \citep{jiang2023llmlingua}, Llmlingua2 \citep{pan2024llmlingua2datadistillationefficient}, H2O \citep{zhang2023h2o}, StreamingLLM \citep{xiaoefficient}, CPC \citep{liskavets2024promptcompressioncontextawaresentence}, and EHPC \citep{fei2025efficientpromptcompressionevaluator}. We also show more experimental results about the long context task in the Appendix \ref{sec:Additional Experiments and Evaluation}.

\begin{table}[t]
\centering
\small
\caption{Distribution alignment between the original LLMs and the merged-input models. We report all in percentage (\%). See Sec.~\ref{sec:experimental_setup} for metric definitions.}
\label{tab:trainingmetric1}
\setlength{\tabcolsep}{8pt} 
\begin{tabular}{lccccc}
\toprule
\textbf{Model} & \textbf{TR} & \textbf{Top-1} & \textbf{Top-3} & \textbf{Top-$p$} & \textbf{MRR} \\
\midrule
Llama-3.1-8B     & 37.8 & 79.2 & 77.6 & 92.9 & 87.1 \\
Llama-3.2-1.5B   & 37.8 & 74.1 & 74.1 & 91.6 & 83.3 \\
GPT2-XL          & 35.7 & 69.6 & 71.3 & 90.7 & 79.3 \\
DeepScaleR-1.5B  & 54.0 & 77.0 & 69.3 & 91.0 & 86.0 \\
\bottomrule
\end{tabular}
\end{table}

\subsection{Predictive Distribution Alignment}

Table~\ref{tab:trainingmetric1} reports predictive distribution alignment results on the corresponding \textit{test} split after training. Higher values indicate better agreement between the predictive distributions produced after token merging and those of the original token-based LLM at aligned positions. Across all backbones, {\namem} achieves strong alignment scores while reducing the effective sequence length by 35.7\%--54.0\%. These results validate that the proposed merge module can preserve much of the original next-token behavior using only static input embeddings, providing a strong foundation for downstream compression and generation.

\subsection{Language Modeling and Generation}
\label{sec:generation_eval_main}

Table~\ref{tab:ppl} reports language modeling performance. All merge modules are trained on the WikiText-103 training split, and the merge rules are derived from the same corpus. Overall, {\namem} preserves language modeling behavior across both in-domain and out-of-domain data. Llama-3.1-8B remains close to the original token-based model across all three datasets while achieving substantial token reduction, and Llama-3.2-1.5B shows the same trend with slightly larger degradation. GPT2-XL is the main exception, exhibiting a larger PPL increase after merging, likely due to its weaker backbone and smaller merge module; nevertheless, it remains stronger than external prompt-compression baselines at comparable compression rates. In terms of compression, the WikiText-103 test split, which matches the training domain of the merge module, naturally exhibits the highest token reduction, with nearly 40\% of tokens removed. More importantly, {\namem} still achieves non-trivial compression on unseen domains, yielding 12--15\% token reduction on BookCorpus and OpenWebText without any retraining. This result suggests that the learned merge operator generalizes beyond the training corpus not only in predictive behavior, but also in its ability to induce effective sequence compression.

\begin{table*}[t]
  \centering
  \caption{\textbf{Language modeling performance under token reduction.}
We report perplexity (PPL; lower is better) and token reduction (TR, \%; higher means more tokens are merged). Original denotes the base model.}
  \label{tab:ppl}
  \small
  \setlength{\tabcolsep}{2pt}
  \renewcommand{\arraystretch}{0.90}
  \begin{threeparttable}
    \begin{tabularx}{\textwidth}{ll
      >{\centering\arraybackslash}X
      >{\centering\arraybackslash}X
      >{\centering\arraybackslash}X
      >{\centering\arraybackslash}X
      >{\centering\arraybackslash}X
      >{\centering\arraybackslash}X
      >{\centering\arraybackslash}X}
      \toprule
      \multirow{2}{*}{\textbf{Backbone}} & \multirow{2}{*}{\textbf{Method}}
        & \multicolumn{2}{c}{\textbf{WikiText-103}}
        & \multicolumn{2}{c}{\textbf{BookCorpus}}
        & \multicolumn{2}{c}{\textbf{OpenWebText}}
        & \multirow{2}{*}{\textbf{PPL Avg.}} \\
      \cmidrule(lr){3-4} \cmidrule(lr){5-6} \cmidrule(lr){7-8}
        & & \textbf{PPL} & \textbf{TR (\%)}
          & \textbf{PPL} & \textbf{TR (\%)}
          & \textbf{PPL} & \textbf{TR (\%)}
          &  \\
      \midrule
      \multirow{4}{*}{Llama-3-8B}
        & Original      & 13.4 & --   & 15.3 & --   & 9.2  & --   &  \\
        & SelectContext & 65.0 & 36.0 & 46.4 & 14.0 & 18.5 & 13.0 & 43.3 \\
        & LLMLingua-2   & 76.5 & 36.0 & 75.5 & 14.0 & 22.9 & 13.0 & 58.3 \\
        \rowcolor{gray!10}
        & Merge Module  & 13.8 & 36.1 & 15.3 & 14.8 & 9.8  & 13.8 & 13.0 \\
      \midrule
      \multirow{4}{*}{Llama-3.2-1B}
        & Original      & 20.1 & --   & 21.2 & --   & 13.4 & --   &  \\
        & SelectContext & 130.4 & 36.0 & 67.0 & 14.0 & 30.7 & 13.0 & 76.0 \\
        & LLMLingua-2   & 129.0 & 36.0 & 101.8 & 14.0 & 37.3 & 13.0 & 89.4 \\
        \rowcolor{gray!10}
        & Merge Module  & 23.4 & 36.1 & 22.0 & 14.8 & 14.7 & 13.8 & 20.7 \\
      \midrule
      \multirow{4}{*}{GPT2-XL}
        & Original      & 28.6 & --   & 27.1 & --   & 13.1 & --   &  \\
        & SelectContext & 202.5 & 35.1 & 93.4 & 14.0 & 39.2 & 12.0 & 111.7 \\
        & LLMLingua-2   & 250.3 & 35.1 & 124.2 & 14.0 & 41.5 & 12.0 & 138.7 \\
        \rowcolor{gray!10}
        & Merge Module  & 37.9 & 35.1 & 30.0 & 14.1 & 14.4 & 12.5 & 27.4 \\
      \bottomrule
    \end{tabularx}
  \end{threeparttable}
\end{table*}
We also provide qualitative generation examples in Table~\ref{tab:qualitative-generation}, using greedy decoding with Llama 3.1 8B. In several cases, token merging leaves the generated output unchanged. In others, the output differs slightly from the original model, but remains coherent and semantically plausible. These examples complement the quantitative PPL results by showing that {\namem} can reduce the effective input length while largely preserving generation quality. More examples are provided in Appendix~\ref{sec:Additional Experiments and Evaluation}.

\begin{table}[t]
  \centering
\caption{\textbf{Qualitative generations with dynamic token-span merging.}
We compare continuations produced by the base model and by {\namem} under the same prompt.
Bold bracketed spans indicate token spans that are merged into a single surrogate embedding, either in the prompt or during autoregressive generation.}
  \label{tab:qualitative-generation}
  \small
  \setlength{\tabcolsep}{3pt}
  \renewcommand{\arraystretch}{0.92}

  \begin{tabularx}{\linewidth}{@{}
    >{\raggedright\arraybackslash}p{0.34\linewidth}
    >{\raggedright\arraybackslash}p{0.25\linewidth}
    >{\raggedright\arraybackslash}p{0.37\linewidth}
  @{}}
    \toprule
    \textbf{Prompt} & \textbf{Base Model} & \textbf{{\namem}} \\
    \midrule
    Traditional Chinese literary criticism emphasized \textbf{[the life of the]} author when interpreting a work
    & The author's life was seen as the key to understanding the work.
    & The author’s life \textbf{[was seen as]} the key \textbf{[to understanding the work]}. \\
    \midrule
    In the spring, \textbf{[his youngest son]}, Ian, was born.
    & He was a beautiful baby boy, and he was the light of his father’s life.
    & He \textbf{[was a beautiful baby]}, \textbf{[with a head of thick]}, dark \textbf{[hair and]} a round, chubby face. \\
    \bottomrule
  \end{tabularx}
\end{table}

\subsection{Task-Agnostic Generalization to Downstream Tasks}
\label{sec:downstream_tasks}

\paragraph{QA and Summarization}

With the \textbf{same} merge modules trained on WikiText-103, we directly apply them to downstream QA and summarization tasks \emph{without any task-specific tuning}. We evaluate zero-shot QA on PIQA \citep{bisk2020piqa}, Copa \citep{gordon-etal-2012-semeval}, OpenBookQA \citep{mihaylov2018suitarmorconductelectricity}, ARC\_Easy, and ARC\_Challenge \citep{clark2018thinksolvedquestionanswering}, and summarization on CNN/DailyMail \citep{nallapati-etal-2016-abstractive} using Rouge \citep{ganesan2018rouge20updatedimproved} and BERTScore \citep{zhang2020bertscoreevaluatingtextgeneration}. For fair comparison, the token reduction (TR) of all baselines is matched to that of our method. Additional results, including code generation, are provided in Appendix~\ref{sec:Additional Experiments and Evaluation}. Across QA benchmarks, {\namem} consistently provides the strongest or most robust accuracy--compression trade-off among the compared methods. This holds across all three backbones and across tasks entirely unseen during training, showing that the same pretrained merge module can transfer directly to unseen downstream tasks without task-specific tuning. \begin{wraptable}{r}{0.52\textwidth}
\vspace{-12pt}
\centering
\caption{Performance of {\namem} against compression baselines. Results are reported as correct/total problems. TR (\%) measures tokens removed.}
\label{tab:Reasoning}
\setlength{\tabcolsep}{7pt}
%\fontsize{7}{8.5}\selectfont
\fontsize{8}{9.5}\selectfont
\begin{tabular}{lcccc}
\toprule
\textbf{Method} & \textbf{AIME} & \textbf{TR(\%)} & \textbf{AMC} & \textbf{TR(\%)} \\
\midrule
DeepScaleR      & 15/30 & --   & 37/40 & --   \\
\rowcolor{gray!20}
{\namem}        & \textbf{13/30} & 40.0 & \textbf{32/40} & 42.4 \\
SelectContext   & \phantom{1}8/30  & 40.0 & 21/40 & 42.4 \\
LLMLingua2      & 11/30 & 40.0 & 28/40 & 42.4 \\
\bottomrule
\end{tabular}
\vspace{-14pt}
\end{wraptable}A similar pattern is observed on CNN/DailyMail summarization. Under the same controlled token reduction, {\namem} remains substantially closer to the original model than all compared baselines, achieving the best Rouge and BERTScore among compressed variants. This further suggests that the learned merge operator preserves generation quality beyond classification-style QA tasks.

\begin{table*}[t]
  \centering
  \small
  \setlength{\tabcolsep}{4pt}
  \renewcommand{\arraystretch}{0.9}
\caption{QA task performance of our Merge Module against compression baselines across three backbones. Acc (\%) measures accuracy on each benchmark; TR (\%) measures the percentage of tokens removed from the input. Bold denotes the best result among compression methods.}
  \label{tab:QA}
  \begin{tabularx}{\textwidth}{l *{10}{>{\centering\arraybackslash}X} c}
    \toprule
    \multirow{2}{*}{\textbf{Methods}}
      & \multicolumn{2}{c}{\textbf{PIQA}}
      & \multicolumn{2}{c}{\textbf{Copa}}
      & \multicolumn{2}{c}{\textbf{OpenBookQA}}
      & \multicolumn{2}{c}{\textbf{ARC-Easy}}
      & \multicolumn{2}{c}{\textbf{ARC-Challenge}}
      & \multirow{2}{*}{\textbf{Avg (\%)}} \\
    \cmidrule(lr){2-3} \cmidrule(lr){4-5} \cmidrule(lr){6-7} \cmidrule(lr){8-9} \cmidrule(lr){10-11}
      & \textbf{Acc (\%)} & \textbf{TR (\%)}
      & \textbf{Acc (\%)} & \textbf{TR (\%)}
      & \textbf{Acc (\%)} & \textbf{TR (\%)}
      & \textbf{Acc (\%)} & \textbf{TR (\%)}
      & \textbf{Acc (\%)} & \textbf{TR (\%)} & \\
    \midrule
    Llama-3.1-8B
      & 79.6 & --  & 76.8 & --  & 43.6 & --  & 76.5 & --  & 49.5 & --  & 65.2 \\
    \rowcolor{gray!20}
    Merge Module
      & \textbf{78.4} & 10 & \textbf{76.8} & 6 & \textbf{42.4} & 10 & \textbf{73.7} & 14 & \textbf{44.8} & 14 & \textbf{63.2} \\
    Select Context
      & 73.3 & 10 & 72.6 & 6 & 40.8 & 10 & 57.2 & 14 & 43.5 & 14 & 53.5 \\
    LLMLingua-2
      & 75.1 & 10 & 72.2 & 6 & 32.8 & 10 & 44.0 & 14 & 32.8 & 14 & 51.4 \\
    CPC
      & 68.7 & 10 & 73.4 & 6 & 32.0 & 10 & 42.5 & 14 & 33.8 & 14 & 50.1 \\
    EHPC
      & 72.7 & 10 & 64.6 & 6 & 28.2 & 10 & 38.3 & 14 & 28.8 & 14 & 46.5 \\
    \midrule
    Llama-3.2-1B
      & 75.5 & --  & 70.6 & --  & 36.0 & --  & 58.8 & --  & 35.5 & --  & 56.3 \\
    \rowcolor{gray!20}
    Merge Module
      & \textbf{72.2} & 10 & 68.8 & 6 & 31.2 & 10 & \textbf{55.3} & 14 & \textbf{33.4} & 14 & \textbf{52.3} \\
    Select Context
      & 68.8 & 10 & 65.8 & 6 & \textbf{35.0} & 10 & 46.5 & 14 & 32.8 & 14 & 49.7 \\
    LLMLingua-2
      & 69.1 & 10 & \textbf{69.2} & 6 & 31.2 & 10 & 37.9 & 14 & 26.4 & 14 & 46.8 \\
    CPC
      & 52.7 & 10 & 68.6 & 6 & 29.4 & 10 & 53.2 & 14 & 31.4 & 14 & 47.1 \\
    EHPC
      & 69.4 & 10 & 60.4 & 6 & 27.0 & 10 & 35.8 & 14 & 25.1 & 14 & 43.5 \\
    \midrule
    GPT2-XL
      & 70.3 & --  & 65.2 & --  & 28.0 & --  & 47.7 & --  & 29.8 & --  & 48.2 \\
    \rowcolor{gray!20}
    Merge Module
      & \textbf{68.8} & 9 & 62.0 & 9 & \textbf{29.0} & 10 & \textbf{43.3} & 14 & \textbf{26.8} & 14 & \textbf{46.0} \\
    Select Context
      & 64.8 & 9 & 61.0 & 9 & \textbf{29.0} & 10 & 40.9 & 14 & 25.4 & 14 & 44.2 \\
    LLMLingua-2
      & 66.6 & 9 & \textbf{63.8} & 9 & 26.4 & 10 & 30.7 & 14 & 25.8 & 14 & 42.6 \\
    CPC
      & 63.2 & 9 & 61.8 & 9 & 28.0 & 10 & 34.6 & 14 & 27.8 & 14 & 43.1 \\
    EHPC
      & 65.7 & 9 & 57.0 & 9 & 25.8 & 10 & 32.8 & 14 & 20.1 & 14 & 40.3 \\
    \bottomrule
  \end{tabularx}
\end{table*}

\begin{table*}[t]
  \centering
  \normalsize
  \setlength{\tabcolsep}{3.5pt}
  \renewcommand{\arraystretch}{0.9}
  \caption{CNN/DailyMail summarization performance of our Merge Module against compression baselines on Llama-3.1-8B. TR (\%) measures the percentage of tokens removed from the input. R-1, R-2, R-L, and R-Lsum denote ROUGE-1, ROUGE-2, ROUGE-L, and ROUGE-Lsum scores respectively; BERTScore measures semantic similarity. All metrics are higher-is-better. Bold denotes the best result among compression methods.}
  \label{tab:summary}
  \begin{tabular}{lcccccccc}
    \toprule
    \textbf{Method} & \textbf{TR (\%)}
      & \textbf{R-1} & \textbf{R-2}
      & \textbf{R-L} & \textbf{R-Lsum}
      & \textbf{Avg} & \textbf{BERTScore} \\
    \midrule
    Llama-3.1-8B   & --  & 37.4 & 15.6 & 24.3 & 31.3 & 27.2 & 86.8 \\
    \rowcolor{gray!20}
    {\namem}       & 16  & \textbf{35.5} & \textbf{13.9} & \textbf{23.1} & \textbf{29.7} & \textbf{25.5} & \textbf{86.4} \\
    SelectContext  & 16  & 33.5 & 11.2 & 21.4 & 27.6 & 23.4 & 85.1 \\
    LLMLingua      & 16  & 33.6 & 10.3 & 21.3 & 27.7 & 23.2 & 85.3 \\
    LLMLingua2     & 16  & 33.0 & 11.3 & 23.0 & 28.8 & 24.0 & 85.0 \\
    StreamingLLM   & 16  & 28.8 & 10.1 & 18.9 & 24.3 & 20.5 & 84.5 \\
    H2O            & 16  & 24.4 & 9.3  & 16.3 & 21.2 & 17.8 & 82.7 \\
    \bottomrule
  \end{tabular}
\end{table*}

\paragraph{Advanced Math Reasoning}

We further evaluate {\namem} on the AIME 2024 \citep{maa2024aime} and, we allow up to 15K generation tokens and evaluate pass@k accuracy under extended chain-of-thought decoding. As shown in Table~\ref{tab:Reasoning} AMC mathematics benchmarks using DeepScaleR, where the merge module is trained on collected long chain-of-thought reasoning trajectories. To specifically test {\namem} in a \textbf{long-form reasoning setting}, {\namem} remains clearly more robust than Select Context and LLMLingua2 under more aggressive compression, retaining strong reasoning performance at over 40\% token reduction. Together with the QA and summarization results, this shows that the same pretrained merge operator transfers effectively beyond its training objective, including challenging long-form generation settings with substantial reasoning depth.

\subsection{Task-Specific Adaptation of a Pretrained Merge Module}
\label{sec:Task_Specific}

We have shown that a merge module trained on WikiText-103 transfers well to unseen downstream tasks. We denote this module \texttt{WT}. The module itself is general, but the WikiText-derived merge rules are not equally well matched to every target task. We therefore study whether UMIM can obtain a better accuracy--compression trade-off by adapting its merge rules and lightweight merge module to the target domain, while keeping the backbone LLM frozen. We call this \textbf{Task Adaptation}. Task adaptation consists of two conceptually separate operations. First, we
re-mine high-frequency token spans from the training split of the target
task and use them to replace the original WikiText-derived merge rules.
Second, we update only the parameters of the merge module $M_\phi$, first through supervised fine-tuning and then through RL. For the experiments in this section, we re-mine task-specific bigram rules only. Consequently, the
maximum possible token reduction would be 50\%. We vary the frequency threshold used to retain bigrams, producing several operating points with different merge rates. We compare five settings on HellaSwag: \texttt{Task-trained:} we discard \texttt{WT} and train a new merge module from scratch using task-specific merge rules.; \texttt{WT + Rules:} we replace the WikiText rules with task-specific rules but keep the pretrained \texttt{WT} merge module fixed; \texttt{WT + FT:} Starting from \texttt{WT}, we apply supervised fine-tuning to $M_\phi$ using the task-specific rules; \texttt{WT + DPO:} we apply DPO directly to $M_\phi$, without the preceding supervised fine-tuning stage; and \texttt{Task-Adapted}, our full route defined as \texttt{WT + Rules + FT + DPO}. We first introduce task-specific rules, then fine-tune $M_\phi$, and finally continue its adaptation with DPO. Across all five settings, the tokenizer and backbone language model remain unchanged and frozen. Thus, the additional task-specific cost is limited to collecting bigram statistics and, when applicable, updating the small merge module.

\paragraph{HellaSwag.}

\begin{figure*}[t]
    \centering
    \includegraphics[width=0.95\textwidth]{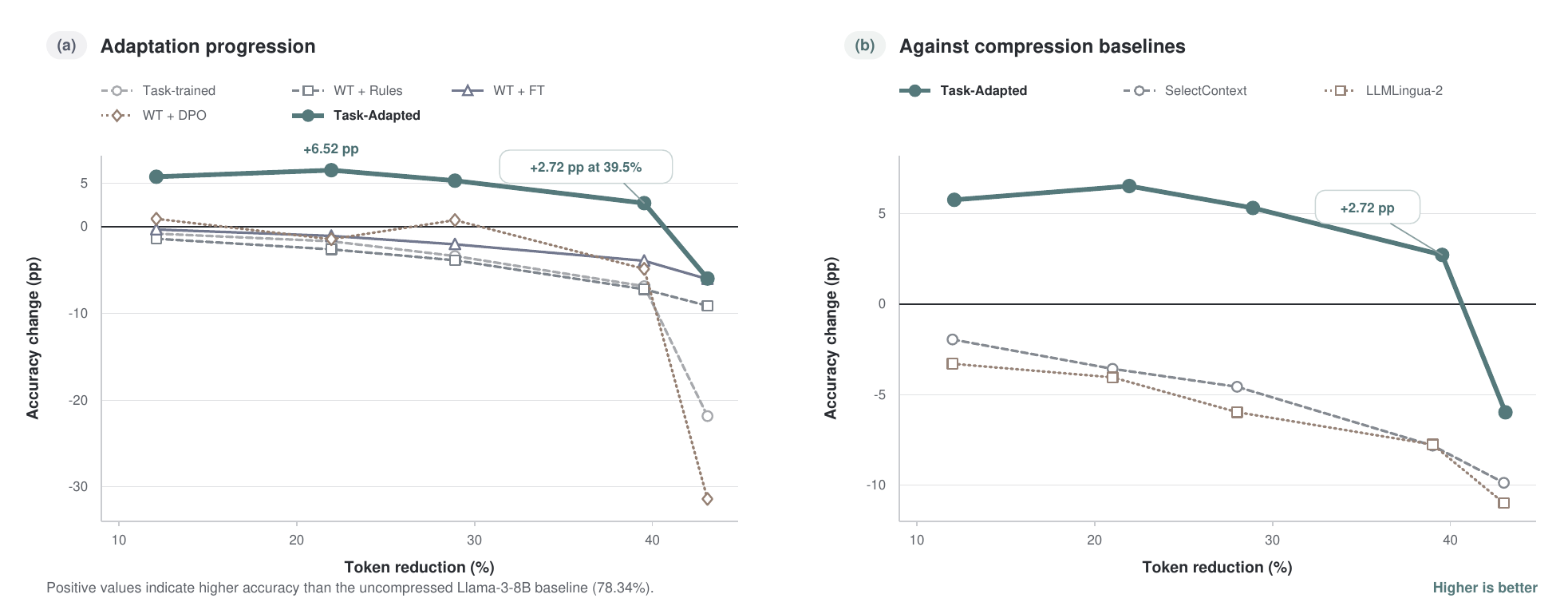}
    \vspace{-8pt}
    \caption{\textbf{Task-specific adaptation on HellaSwag.}
The y-axis reports accuracy change relative to the uncompressed model (0 = baseline).
(a) We compare five adaptation settings, four of which build on the WikiText-103 pretrained module (\texttt{WT}); \texttt{Task-Adapted} (\texttt{WT + Rules + FT + DPO}) achieves the best accuracy--compression trade-off.
(b) The strongest adapted variant also consistently outperforms external baselines at comparable merge rates.}
    \label{fig:hellaswag_progress}
\end{figure*}

Figure~\ref{fig:hellaswag_progress} separates the contributions of task-specific rules, WikiText initialization, supervised fine-tuning, and DPO. When the original WikiText rules are used directly on HellaSwag, their coverage is limited because the span distribution differs from that of WikiText-103. Re-mining bigrams from the HellaSwag training split produces operating points with merge rates from 12.10\% to 43.09\%. The \texttt{WT + Rules} setting shows that the WikiText-trained merge module can immediately operate with a newly constructed rule set, without any parameter updates. Its accuracy decreases as increasingly aggressive rules are introduced, but it provides substantially greater task-specific coverage. This result indicates that the learned merge operation is not tied to the specific spans seen during WikiText training. Supervised fine-tuning consistently improves upon \texttt{WT + Rules} across merge rates. It allows the merge module to accommodate the new span distribution while retaining the strong initialization learned from WikiText-103. In contrast, training the merge module from scratch (\texttt{Task-trained}) is less effective, showing that the transferable representation learned by \texttt{WT} remains useful even after the rules have been replaced. Applying DPO directly to \texttt{WT} is less stable, particularly at high merge rates. The complete \texttt{Task-Adapted} route instead performs supervised fine-tuning before DPO and achieves the strongest accuracy--compression trade-off. It reaches 84.86\% accuracy at a 21.94\% merge rate, compared with 78.34\% for the uncompressed backbone. Even at a 39.53\% merge rate, it obtains 81.06\% accuracy, remaining 2.72 percentage points above the uncompressed baseline. Figure~\ref{fig:hellaswag_progress}(b) also shows that \texttt{Task-Adapted} consistently outperforms other baselines at comparable merge rates.

\begin{figure}[t]
    \centering
    \includegraphics[width=\linewidth]{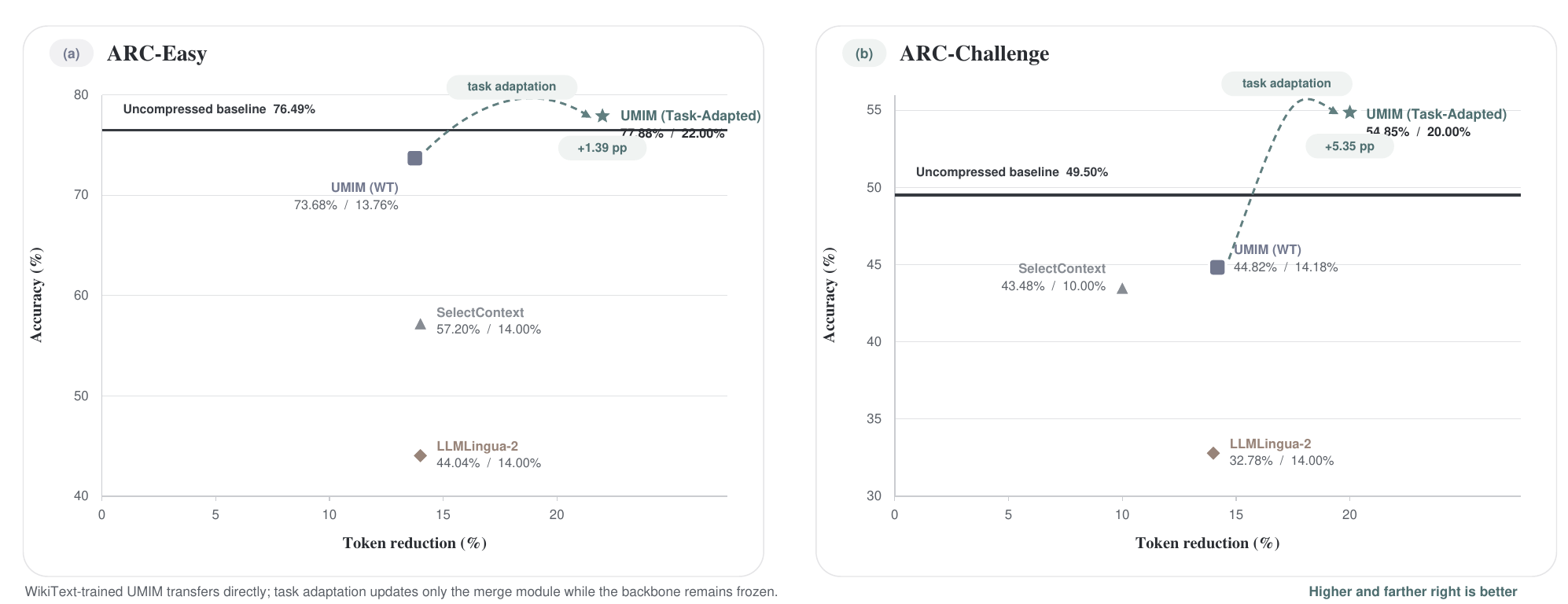}
    \caption{\textbf{Accuracy--compression trade-off on ARC-Easy and ARC-Challenge.}
Each point shows a method's accuracy (\%) vs.\ token reduction (\%). 
The dashed horizontal line indicates the uncompressed baseline accuracy.
The arrow marks the effect of downstream adaptation from \texttt{WT} to \texttt{Task-Adapted}.}
    \label{fig:arc_compar}
\end{figure}

\paragraph{ARC-Easy and ARC-Challenge.}
We further evaluate the same adaptation strategy on ARC-Easy and ARC-Challenge, as shown in Figure~\ref{fig:arc_compar}. Before task-specific training, \texttt{WT} already provides non-trivial compression on both tasks: it achieves 73.68\% accuracy with 13.76\% token reduction on ARC-Easy and 44.82\% accuracy with 14.18\% token reduction on ARC-Challenge. Adapting only the merge module substantially improves both operating points. On ARC-Easy, \texttt{Task-Adapted} reaches 77.88\% accuracy with 22.00\% token reduction, exceeding the uncompressed baseline of 76.49\%. On the more difficult ARC-Challenge benchmark, it improves from the 49.50\% uncompressed baseline to 54.85\% accuracy while reducing the number of tokens by 20.00\%. These results show that the benefit of task-specific adaptation is not limited to HellaSwag: the WikiText-trained module provides a reusable initialization, and lightweight downstream adaptation can simultaneously increase compression and recover—or even improve—task accuracy.

\subsection{Ablation Study and Efficiency Analysis}
\label{sec:ablation_study}

We conduct ablations along two dimensions: merge-rule construction and merge-module design. Overall, the results show that UMIM is not highly sensitive to a single rule-construction strategy, but it is sensitive to overly aggressive merging. They also show that the current merge module achieves a strong accuracy--compression trade-off under a lightweight parameter budget, with limited gains from further increasing capacity. For merge rules (in Appendix ~\ref{sec:alternative_rule_ablation}), we study different frequency thresholds, span construction strategies, and alternative rule-construction methods. The results show that setting the frequency threshold $\tau$ too low introduces many low-quality merges and leads to clear performance degradation. This suggests that avoiding over-merging is more important than relying on a particular rule heuristic. Across the evaluated rule-construction strategies, we do not observe large performance differences, indicating that UMIM is reasonably robust to the exact rule design as long as the resulting merges provide sufficient training signal and do not over-compress the input. For the merge module (in Appendix ~\ref{sec:ablation_study}), we compare different parameter sizes and structural configurations. The current attention-based design already performs well with a small parameter budget, while increasing capacity yields only marginal improvements. We also find that simpler linear variants or deeper but less efficient structures do not provide better trade-offs. These results suggest that the current module design is effective for learning surrogate span representations, and that UMIM does not primarily rely on brute-force scaling to achieve strong compression performance.

Following the evaluation protocol of Zip2Zip \citep{geng2025zip2zipinferencetimeadaptivetokenization}, we conduct an end-to-end throughput study on Llama-3.1-8B with a fixed 50\% merge ratio. We report tokens/sec for both prefill and decoding, with each setting in Table~\ref{tab:throughput} denoted as \texttt{prefill length + decode length}. As shown in Table~\ref{tab:throughput}, {\namem} provides substantial prefill speedups, with larger gains at longer input contexts because merging directly reduces the effective prompt length. Decoding throughput also improves consistently across settings. These results show that {\namem} improves end-to-end inference efficiency in both prompt encoding and autoregressive generation.

\section{Conclusion}
\label{sec:conclusion}

We introduced {\namem}, an token merging framework. Rather than deleting tokens or modifying the Transformer architecture, {\namem} maps a matched multi-token span to a single surrogate embedding through a lightweight merge module. In our base setting, high-frequency n-gram spans are mined from base corpus to construct the merge rules, and the merge module is trained by distilling the behavior of the original uncompressed sequence. The resulting module operates only on static input embeddings, while the backbone language model remain unchanged. At inference, {\namem} first merges eligible spans in the prompt before prefill. During decoding, it generates every token as usual and then checks whether the new suffix matches a merge rule. If a match is found, {\namem} replaces the corresponding KV entries with one merged entry, shortening the sequence and KV cache used in later steps without skipping any token generation. Our experiments show that the WikiText-trained merge module transfers directly
to unseen corpora and downstream tasks without retraining. For a new task,
simply re-mining the merge rules increases coverage, while fine-tuning followed
by DPO further improves the accuracy--compression trade-off. Only the merge
module is adapted; the backbone language model remains frozen throughout.

\section*{Limitations}
\label{Limitations and Future Work}

Our current study focuses on establishing {\namem} as a practical input-level compression framework for frozen language models across representative models and tasks. Future work may explore more adaptive merging strategies, broader model scales and domains, and the interaction between pretrained merge modules and downstream adaptation.

\section*{Acknowledgment}
We thank the Google Gemma Academic Program
for their partial support of Jiawei Zhou and for providing computational resources.
Jiawei Zhou is also supported by an Amazon Research Award and a Stony Brook
OVPR Seed Grant.
\bibliographystyle{plainnat}
\bibliography{example_paper}

@article{vaswani2017attention,
  title={Attention is all you need},
  author={Vaswani, Ashish and Shazeer, Noam and Parmar, Niki and Uszkoreit, Jakob and Jones, Llion and Gomez, Aidan N and Kaiser, {\L}ukasz and Polosukhin, Illia},
  journal={Advances in neural information processing systems},
  volume={30},
  year={2017}
}

@article{adnan2024keyformer,
  title={Keyformer: Kv cache reduction through key tokens selection for efficient generative inference},
  author={Adnan, Muhammad and Arunkumar, Akhil and Jain, Gaurav and Nair, Prashant J and Soloveychik, Ilya and Kamath, Purushotham},
  journal={Proceedings of Machine Learning and Systems},
  volume={6},
  pages={114--127},
  year={2024}
}

@article{zhang2023h2o,
  title={H2o: Heavy-hitter oracle for efficient generative inference of large language models},
  author={Zhang, Zhenyu and Sheng, Ying and Zhou, Tianyi and Chen, Tianlong and Zheng, Lianmin and Cai, Ruisi and Song, Zhao and Tian, Yuandong and R{\'e}, Christopher and Barrett, Clark and others},
  journal={Advances in Neural Information Processing Systems},
  volume={36},
  pages={34661--34710},
  year={2023}
}

@article{li2023unlocking,
  title={Unlocking context constraints of llms: Enhancing context efficiency of llms with self-information-based content filtering},
  author={Li, Yucheng},
  journal={arXiv preprint arXiv:2304.12102},
  year={2023}
}

@article{jiang2023llmlingua,
  title={Llmlingua: Compressing prompts for accelerated inference of large language models},
  author={Jiang, Huiqiang and Wu, Qianhui and Lin, Chin-Yew and Yang, Yuqing and Qiu, Lili},
  journal={arXiv preprint arXiv:2310.05736},
  year={2023}
}

@inproceedings{kwon2023efficient,
  title={Efficient memory management for large language model serving with pagedattention},
  author={Kwon, Woosuk and Li, Zhuohan and Zhuang, Siyuan and Sheng, Ying and Zheng, Lianmin and Yu, Cody Hao and Gonzalez, Joseph and Zhang, Hao and Stoica, Ion},
  booktitle={Proceedings of the 29th Symposium on Operating Systems Principles},
  pages={611--626},
  year={2023}
}

@article{dao2022flashattention,
  title={Flashattention: Fast and memory-efficient exact attention with io-awareness},
  author={Dao, Tri and Fu, Dan and Ermon, Stefano and Rudra, Atri and R{\'e}, Christopher},
  journal={Advances in neural information processing systems},
  volume={35},
  pages={16344--16359},
  year={2022}
}

@article{mu2023learning,
  title={Learning to compress prompts with gist tokens},
  author={Mu, Jesse and Li, Xiang and Goodman, Noah},
  journal={Advances in Neural Information Processing Systems},
  volume={36},
  pages={19327--19352},
  year={2023}
}

@article{kallini2024mrt5,
  title={Mrt5: Dynamic token merging for efficient byte-level language models},
  author={Kallini, Julie and Murty, Shikhar and Manning, Christopher D and Potts, Christopher and Csord{\'a}s, R{\'o}bert},
  journal={arXiv preprint arXiv:2410.20771},
  year={2024}
}

@article{lan2023copy,
  title={Copy is all you need},
  author={Lan, Tian and Cai, Deng and Wang, Yan and Huang, Heyan and Mao, Xian-Ling},
  journal={arXiv preprint arXiv:2307.06962},
  year={2023}
}

@article{xue2022byt5,
  title={Byt5: Towards a token-free future with pre-trained byte-to-byte models},
  author={Xue, Linting and Barua, Aditya and Constant, Noah and Al-Rfou, Rami and Narang, Sharan and Kale, Mihir and Roberts, Adam and Raffel, Colin},
  journal={Transactions of the Association for Computational Linguistics},
  volume={10},
  pages={291--306},
  year={2022},
  publisher={MIT Press One Broadway, 12th Floor, Cambridge, Massachusetts 02142, USA~…}
}

@article{pagnoni2024byte,
  title={Byte latent transformer: Patches scale better than tokens},
  author={Pagnoni, Artidoro and Pasunuru, Ram and Rodriguez, Pedro and Nguyen, John and Muller, Benjamin and Li, Margaret and Zhou, Chunting and Yu, Lili and Weston, Jason and Zettlemoyer, Luke and others},
  journal={arXiv preprint arXiv:2412.09871},
  year={2024}
}

@article{liu2025superbpe,
  title={Superbpe: Space travel for language models},
  author={Liu, Alisa and Hayase, Jonathan and Hofmann, Valentin and Oh, Sewoong and Smith, Noah A and Choi, Yejin},
  journal={arXiv preprint arXiv:2503.13423},
  year={2025}
}

@inproceedings{leviathan2023fast,
  title={Fast inference from transformers via speculative decoding},
  author={Leviathan, Yaniv and Kalman, Matan and Matias, Yossi},
  booktitle={International Conference on Machine Learning},
  pages={19274--19286},
  year={2023},
  organization={PMLR}
}

@inproceedings{lichunk,
  title={Chunk-Distilled Language Modeling},
  author={Li, Yanhong and Livescu, Karen and Zhou, Jiawei},
  booktitle={The Thirteenth International Conference on Learning Representations},
year={2025}
}

@article{yu2023megabyte,
  title={Megabyte: Predicting million-byte sequences with multiscale transformers},
  author={Yu, Lili and Simig, D{\'a}niel and Flaherty, Colin and Aghajanyan, Armen and Zettlemoyer, Luke and Lewis, Mike},
  journal={Advances in Neural Information Processing Systems},
  volume={36},
  pages={78808--78823},
  year={2023}
}

@inproceedings{bolya2023token,
  title={Token Merging: Your ViT But Faster},
  author={Bolya, Daniel and Fu, Cheng-Yang and Dai, Xiaoliang and Zhang, Peizhao and Feichtenhofer, Christoph and Hoffman, Judy},
  booktitle={ICLR},
  year={2023}
}

@inproceedings{xiaoefficient,
  title={Efficient Streaming Language Models with Attention Sinks},
  author={Xiao, Guangxuan and Tian, Yuandong and Chen, Beidi and Han, Song and Lewis, Mike},
  booktitle={The Twelfth International Conference on Learning Representations},
year={2024}
}

@inproceedings{gecontext,
  title={In-context Autoencoder for Context Compression in a Large Language Model},
  author={Ge, Tao and Jing, Hu and Wang, Lei and Wang, Xun and Chen, Si-Qing and Wei, Furu},
  booktitle={The Twelfth International Conference on Learning Representations},
year={2024}
}

@article{shao2024flexibly,
  title={Flexibly scaling large language models contexts through extensible tokenization},
  author={Shao, Ninglu and Xiao, Shitao and Liu, Zheng and Zhang, Peitian},
  journal={arXiv preprint arXiv:2401.07793},
  year={2024}
}

@inproceedings{hao2025omnikv,
  title={OmniKV: Dynamic context selection for efficient long-context LLMs},
  author={Hao, Jitai and Zhu, Yuke and Wang, Tian and Yu, Jun and Xin, Xin and Zheng, Bo and Ren, Zhaochun and Guo, Sheng},
  booktitle={The Thirteenth International Conference on Learning Representations},
  year={2025}
}

@inproceedings{yuan2024efficient,
  title={Efficient transformer adaptation with soft token merging},
  author={Yuan, Xin and Fei, Hongliang and Baek, Jinoo},
  booktitle={Proceedings of the IEEE/CVF Conference on Computer Vision and Pattern Recognition},
  pages={3658--3668},
  year={2024}
}

@article{ryoo2021tokenlearner,
  title={Tokenlearner: Adaptive space-time tokenization for videos},
  author={Ryoo, Michael and Piergiovanni, AJ and Arnab, Anurag and Dehghani, Mostafa and Angelova, Anelia},
  journal={Advances in neural information processing systems},
  volume={34},
  pages={12786--12797},
  year={2021}
}

@article{shin2025atom,
  title={AToM: Adaptive Token Merging for Efficient Acceleration of Vision Transformer},
  author={Shin, Jaekang and Kang, Myeonggu and Han, Yunki and Park, Junyoung and Kim, Lee-Sup},
  journal={IEEE Transactions on Computers},
  year={2025},
  publisher={IEEE}
}

@inproceedings{jiang2024longllmlingua,
  title={LongLLMLingua: Accelerating and Enhancing LLMs in Long Context Scenarios via Prompt Compression},
  author={Jiang, Huiqiang and Wu, Qianhui and Luo, Xufang and Li, Dongsheng and Lin, Chin-Yew and Yang, Yuqing and Qiu, Lili},
  booktitle={Proceedings of the 62nd Annual Meeting of the Association for Computational Linguistics (Volume 1: Long Papers)},
  pages={1658--1677},
  year={2024}
}

@inproceedings{xu2024recomp,
  title={RECOMP: Improving retrieval-augmented LMs with context compression and selective augmentation},
  author={Xu, Fangyuan and Shi, Weijia and Choi, Eunsol},
  booktitle={The Twelfth International Conference on Learning Representations},
year={2024}
}

@misc{grattafiori2024llama3herdmodels,
      title={The Llama 3 Herd of Models}, 
      author={Aaron Grattafiori and Abhimanyu Dubey and Abhinav Jauhri and Abhinav Pandey and Abhishek Kadian and Ahmad Al-Dahle and Aiesha Letman and Akhil Mathur and Alan Schelten and Alex Vaughan and Amy Yang and Angela Fan and Anirudh Goyal and Anthony Hartshorn and Aobo Yang and Archi Mitra and Archie Sravankumar and Artem Korenev and Arthur Hinsvark and Arun Rao and Aston Zhang and Aurelien Rodriguez and Austen Gregerson and Ava Spataru and Baptiste Roziere and Bethany Biron and Binh Tang and Bobbie Chern and Charlotte Caucheteux and Chaya Nayak and Chloe Bi and Chris Marra and Chris McConnell and Christian Keller and Christophe Touret and Chunyang Wu and Corinne Wong and Cristian Canton Ferrer and Cyrus Nikolaidis and Damien Allonsius and Daniel Song and Danielle Pintz and Danny Livshits and Danny Wyatt and David Esiobu and Dhruv Choudhary and Dhruv Mahajan and Diego Garcia-Olano and Diego Perino and Dieuwke Hupkes and Egor Lakomkin and Ehab AlBadawy and Elina Lobanova and Emily Dinan and Eric Michael Smith and Filip Radenovic and Francisco Guzmán and Frank Zhang and Gabriel Synnaeve and Gabrielle Lee and Georgia Lewis Anderson and Govind Thattai and Graeme Nail and Gregoire Mialon and Guan Pang and Guillem Cucurell and Hailey Nguyen and Hannah Korevaar and Hu Xu and Hugo Touvron and Iliyan Zarov and Imanol Arrieta Ibarra and Isabel Kloumann and Ishan Misra and Ivan Evtimov and Jack Zhang and Jade Copet and Jaewon Lee and Jan Geffert and Jana Vranes and Jason Park and Jay Mahadeokar and Jeet Shah and Jelmer van der Linde and Jennifer Billock and Jenny Hong and Jenya Lee and Jeremy Fu and Jianfeng Chi and Jianyu Huang and Jiawen Liu and Jie Wang and Jiecao Yu and Joanna Bitton and Joe Spisak and Jongsoo Park and Joseph Rocca and Joshua Johnstun and Joshua Saxe and Junteng Jia and Kalyan Vasuden Alwala and Karthik Prasad and Kartikeya Upasani and Kate Plawiak and Ke Li and Kenneth Heafield and Kevin Stone and Khalid El-Arini and Krithika Iyer and Kshitiz Malik and Kuenley Chiu and Kunal Bhalla and Kushal Lakhotia and Lauren Rantala-Yeary and Laurens van der Maaten and Lawrence Chen and Liang Tan and Liz Jenkins and Louis Martin and Lovish Madaan and Lubo Malo and Lukas Blecher and Lukas Landzaat and Luke de Oliveira and Madeline Muzzi and Mahesh Pasupuleti and Mannat Singh and Manohar Paluri and Marcin Kardas and Maria Tsimpoukelli and Mathew Oldham and Mathieu Rita and Maya Pavlova and Melanie Kambadur and Mike Lewis and Min Si and Mitesh Kumar Singh and Mona Hassan and Naman Goyal and Narjes Torabi and Nikolay Bashlykov and Nikolay Bogoychev and Niladri Chatterji and Ning Zhang and Olivier Duchenne and Onur Çelebi and Patrick Alrassy and Pengchuan Zhang and Pengwei Li and Petar Vasic and Peter Weng and Prajjwal Bhargava and Pratik Dubal and Praveen Krishnan and Punit Singh Koura and Puxin Xu and Qing He and Qingxiao Dong and Ragavan Srinivasan and Raj Ganapathy and Ramon Calderer and Ricardo Silveira Cabral and Robert Stojnic and Roberta Raileanu and Rohan Maheswari and Rohit Girdhar and Rohit Patel and Romain Sauvestre and Ronnie Polidoro and Roshan Sumbaly and Ross Taylor and Ruan Silva and Rui Hou and Rui Wang and Saghar Hosseini and Sahana Chennabasappa and Sanjay Singh and Sean Bell and Seohyun Sonia Kim and Sergey Edunov and Shaoliang Nie and Sharan Narang and Sharath Raparthy and Sheng Shen and Shengye Wan and Shruti Bhosale and Shun Zhang and Simon Vandenhende and Soumya Batra and Spencer Whitman and Sten Sootla and Stephane Collot and Suchin Gururangan and Sydney Borodinsky and Tamar Herman and Tara Fowler and Tarek Sheasha and Thomas Georgiou and Thomas Scialom and Tobias Speckbacher and Todor Mihaylov and Tong Xiao and Ujjwal Karn and Vedanuj Goswami and Vibhor Gupta and Vignesh Ramanathan and Viktor Kerkez and Vincent Gonguet and Virginie Do and Vish Vogeti and Vítor Albiero and Vladan Petrovic and Weiwei Chu and Wenhan Xiong and Wenyin Fu and Whitney Meers and Xavier Martinet and Xiaodong Wang and Xiaofang Wang and Xiaoqing Ellen Tan and Xide Xia and Xinfeng Xie and Xuchao Jia and Xuewei Wang and Yaelle Goldschlag and Yashesh Gaur and Yasmine Babaei and Yi Wen and Yiwen Song and Yuchen Zhang and Yue Li and Yuning Mao and Zacharie Delpierre Coudert and Zheng Yan and Zhengxing Chen and Zoe Papakipos and Aaditya Singh and Aayushi Srivastava and Abha Jain and Adam Kelsey and Adam Shajnfeld and Adithya Gangidi and Adolfo Victoria and Ahuva Goldstand and Ajay Menon and Ajay Sharma and Alex Boesenberg and Alexei Baevski and Allie Feinstein and Amanda Kallet and Amit Sangani and Amos Teo and Anam Yunus and Andrei Lupu and Andres Alvarado and Andrew Caples and Andrew Gu and Andrew Ho and Andrew Poulton and Andrew Ryan and Ankit Ramchandani and Annie Dong and Annie Franco and Anuj Goyal and Aparajita Saraf and Arkabandhu Chowdhury and Ashley Gabriel and Ashwin Bharambe and Assaf Eisenman and Azadeh Yazdan and Beau James and Ben Maurer and Benjamin Leonhardi and Bernie Huang and Beth Loyd and Beto De Paola and Bhargavi Paranjape and Bing Liu and Bo Wu and Boyu Ni and Braden Hancock and Bram Wasti and Brandon Spence and Brani Stojkovic and Brian Gamido and Britt Montalvo and Carl Parker and Carly Burton and Catalina Mejia and Ce Liu and Changhan Wang and Changkyu Kim and Chao Zhou and Chester Hu and Ching-Hsiang Chu and Chris Cai and Chris Tindal and Christoph Feichtenhofer and Cynthia Gao and Damon Civin and Dana Beaty and Daniel Kreymer and Daniel Li and David Adkins and David Xu and Davide Testuggine and Delia David and Devi Parikh and Diana Liskovich and Didem Foss and Dingkang Wang and Duc Le and Dustin Holland and Edward Dowling and Eissa Jamil and Elaine Montgomery and Eleonora Presani and Emily Hahn and Emily Wood and Eric-Tuan Le and Erik Brinkman and Esteban Arcaute and Evan Dunbar and Evan Smothers and Fei Sun and Felix Kreuk and Feng Tian and Filippos Kokkinos and Firat Ozgenel and Francesco Caggioni and Frank Kanayet and Frank Seide and Gabriela Medina Florez and Gabriella Schwarz and Gada Badeer and Georgia Swee and Gil Halpern and Grant Herman and Grigory Sizov and Guangyi and Zhang and Guna Lakshminarayanan and Hakan Inan and Hamid Shojanazeri and Han Zou and Hannah Wang and Hanwen Zha and Haroun Habeeb and Harrison Rudolph and Helen Suk and Henry Aspegren and Hunter Goldman and Hongyuan Zhan and Ibrahim Damlaj and Igor Molybog and Igor Tufanov and Ilias Leontiadis and Irina-Elena Veliche and Itai Gat and Jake Weissman and James Geboski and James Kohli and Janice Lam and Japhet Asher and Jean-Baptiste Gaya and Jeff Marcus and Jeff Tang and Jennifer Chan and Jenny Zhen and Jeremy Reizenstein and Jeremy Teboul and Jessica Zhong and Jian Jin and Jingyi Yang and Joe Cummings and Jon Carvill and Jon Shepard and Jonathan McPhie and Jonathan Torres and Josh Ginsburg and Junjie Wang and Kai Wu and Kam Hou U and Karan Saxena and Kartikay Khandelwal and Katayoun Zand and Kathy Matosich and Kaushik Veeraraghavan and Kelly Michelena and Keqian Li and Kiran Jagadeesh and Kun Huang and Kunal Chawla and Kyle Huang and Lailin Chen and Lakshya Garg and Lavender A and Leandro Silva and Lee Bell and Lei Zhang and Liangpeng Guo and Licheng Yu and Liron Moshkovich and Luca Wehrstedt and Madian Khabsa and Manav Avalani and Manish Bhatt and Martynas Mankus and Matan Hasson and Matthew Lennie and Matthias Reso and Maxim Groshev and Maxim Naumov and Maya Lathi and Meghan Keneally and Miao Liu and Michael L. Seltzer and Michal Valko and Michelle Restrepo and Mihir Patel and Mik Vyatskov and Mikayel Samvelyan and Mike Clark and Mike Macey and Mike Wang and Miquel Jubert Hermoso and Mo Metanat and Mohammad Rastegari and Munish Bansal and Nandhini Santhanam and Natascha Parks and Natasha White and Navyata Bawa and Nayan Singhal and Nick Egebo and Nicolas Usunier and Nikhil Mehta and Nikolay Pavlovich Laptev and Ning Dong and Norman Cheng and Oleg Chernoguz and Olivia Hart and Omkar Salpekar and Ozlem Kalinli and Parkin Kent and Parth Parekh and Paul Saab and Pavan Balaji and Pedro Rittner and Philip Bontrager and Pierre Roux and Piotr Dollar and Polina Zvyagina and Prashant Ratanchandani and Pritish Yuvraj and Qian Liang and Rachad Alao and Rachel Rodriguez and Rafi Ayub and Raghotham Murthy and Raghu Nayani and Rahul Mitra and Rangaprabhu Parthasarathy and Raymond Li and Rebekkah Hogan and Robin Battey and Rocky Wang and Russ Howes and Ruty Rinott and Sachin Mehta and Sachin Siby and Sai Jayesh Bondu and Samyak Datta and Sara Chugh and Sara Hunt and Sargun Dhillon and Sasha Sidorov and Satadru Pan and Saurabh Mahajan and Saurabh Verma and Seiji Yamamoto and Sharadh Ramaswamy and Shaun Lindsay and Shaun Lindsay and Sheng Feng and Shenghao Lin and Shengxin Cindy Zha and Shishir Patil and Shiva Shankar and Shuqiang Zhang and Shuqiang Zhang and Sinong Wang and Sneha Agarwal and Soji Sajuyigbe and Soumith Chintala and Stephanie Max and Stephen Chen and Steve Kehoe and Steve Satterfield and Sudarshan Govindaprasad and Sumit Gupta and Summer Deng and Sungmin Cho and Sunny Virk and Suraj Subramanian and Sy Choudhury and Sydney Goldman and Tal Remez and Tamar Glaser and Tamara Best and Thilo Koehler and Thomas Robinson and Tianhe Li and Tianjun Zhang and Tim Matthews and Timothy Chou and Tzook Shaked and Varun Vontimitta and Victoria Ajayi and Victoria Montanez and Vijai Mohan and Vinay Satish Kumar and Vishal Mangla and Vlad Ionescu and Vlad Poenaru and Vlad Tiberiu Mihailescu and Vladimir Ivanov and Wei Li and Wenchen Wang and Wenwen Jiang and Wes Bouaziz and Will Constable and Xiaocheng Tang and Xiaojian Wu and Xiaolan Wang and Xilun Wu and Xinbo Gao and Yaniv Kleinman and Yanjun Chen and Ye Hu and Ye Jia and Ye Qi and Yenda Li and Yilin Zhang and Ying Zhang and Yossi Adi and Youngjin Nam and Yu and Wang and Yu Zhao and Yuchen Hao and Yundi Qian and Yunlu Li and Yuzi He and Zach Rait and Zachary DeVito and Zef Rosnbrick and Zhaoduo Wen and Zhenyu Yang and Zhiwei Zhao and Zhiyu Ma},
      year={2024},
      eprint={2407.21783},
      archivePrefix={arXiv},
      primaryClass={cs.AI},
      url={https://arxiv.org/abs/2407.21783}, 
}

@article{radford2019language,
  title={Language models are unsupervised multitask learners},
  author={Radford, Alec and Wu, Jeffrey and Child, Rewon and Luan, David and Amodei, Dario and Sutskever, Ilya and others},
  journal={OpenAI blog},
  volume={1},
  number={8},
  pages={9},
  year={2019}
}

@misc{merity2016pointersentinelmixturemodels,
      title={Pointer Sentinel Mixture Models}, 
      author={Stephen Merity and Caiming Xiong and James Bradbury and Richard Socher},
      year={2016},
      eprint={1609.07843},
      archivePrefix={arXiv},
      primaryClass={cs.CL},
      url={https://arxiv.org/abs/1609.07843}, 
}

@misc{pan2024llmlingua2datadistillationefficient,
      title={LLMLingua-2: Data Distillation for Efficient and Faithful Task-Agnostic Prompt Compression}, 
      author={Zhuoshi Pan and Qianhui Wu and Huiqiang Jiang and Menglin Xia and Xufang Luo and Jue Zhang and Qingwei Lin and Victor Rühle and Yuqing Yang and Chin-Yew Lin and H. Vicky Zhao and Lili Qiu and Dongmei Zhang},
      year={2024},
      eprint={2403.12968},
      archivePrefix={arXiv},
      primaryClass={cs.CL},
      url={https://arxiv.org/abs/2403.12968}, 
}

@article{li2023compressing,
  title={Compressing context to enhance inference efficiency of large language models},
  author={Li, Yucheng and Dong, Bo and Lin, Chenghua and Guerin, Frank},
  journal={arXiv preprint arXiv:2310.06201},
  year={2023}
}

@inproceedings{zhu2015aligning,
  title={Aligning books and movies: Towards story-like visual explanations by watching movies and reading books},
  author={Zhu, Yukun and Kiros, Ryan and Zemel, Rich and Salakhutdinov, Ruslan and Urtasun, Raquel and Torralba, Antonio and Fidler, Sanja},
  booktitle={Proceedings of the IEEE international conference on computer vision},
  pages={19--27},
  year={2015}
}

@misc{Gokaslan2019OpenWeb,  
	title={OpenWebText Corpus},
	author={Aaron Gokaslan and Vanya Cohen},
	howpublished={\url{http://Skylion007.github.io/OpenWebTextCorpus}}, 
	year={2019}
}

@inproceedings{bisk2020piqa,
  title={Piqa: Reasoning about physical commonsense in natural language},
  author={Bisk, Yonatan and Zellers, Rowan and Gao, Jianfeng and Choi, Yejin and others},
  booktitle={Proceedings of the AAAI conference on artificial intelligence},
  volume={34},
  number={05},
  pages={7432--7439},
  year={2020}
}

@inproceedings{gordon-etal-2012-semeval,
    title = "{S}em{E}val-2012 Task 7: Choice of Plausible Alternatives: An Evaluation of Commonsense Causal Reasoning",
    author = "Gordon, Andrew  and
      Kozareva, Zornitsa  and
      Roemmele, Melissa",
    editor = "Agirre, Eneko  and
      Bos, Johan  and
      Diab, Mona  and
      Manandhar, Suresh  and
      Marton, Yuval  and
      Yuret, Deniz",
    booktitle = "*{SEM} 2012: The First Joint Conference on Lexical and Computational Semantics {--} Volume 1: Proceedings of the main conference and the shared task, and Volume 2: Proceedings of the Sixth International Workshop on Semantic Evaluation ({S}em{E}val 2012)",
    month = "7-8 " # jun,
    year = "2012",
    address = "Montr{\'e}al, Canada",
    publisher = "Association for Computational Linguistics",
    url = "https://aclanthology.org/S12-1052/",
    pages = "394--398"
}

@misc{mihaylov2018suitarmorconductelectricity,
      title={Can a Suit of Armor Conduct Electricity? A New Dataset for Open Book Question Answering}, 
      author={Todor Mihaylov and Peter Clark and Tushar Khot and Ashish Sabharwal},
      year={2018},
      eprint={1809.02789},
      archivePrefix={arXiv},
      primaryClass={cs.CL},
      url={https://arxiv.org/abs/1809.02789}, 
}

@inproceedings{nallapati-etal-2016-abstractive,
    title = "Abstractive Text Summarization using Sequence-to-sequence {RNN}s and Beyond",
    author = "Nallapati, Ramesh  and
      Zhou, Bowen  and
      dos Santos, Cicero  and
      Gu{\ensuremath{\dot{}}}l{\c{c}}ehre, {\c{C}}a{\u{g}}lar  and
      Xiang, Bing",
    editor = "Riezler, Stefan  and
      Goldberg, Yoav",
    booktitle = "Proceedings of the 20th {SIGNLL} Conference on Computational Natural Language Learning",
    month = aug,
    year = "2016",
    address = "Berlin, Germany",
    publisher = "Association for Computational Linguistics",
    url = "https://aclanthology.org/K16-1028/",
    doi = "10.18653/v1/K16-1028",
    pages = "280--290"
}

@misc{ganesan2018rouge20updatedimproved,
      title={ROUGE 2.0: Updated and Improved Measures for Evaluation of Summarization Tasks}, 
      author={Kavita Ganesan},
      year={2018},
      eprint={1803.01937},
      archivePrefix={arXiv},
      primaryClass={cs.IR},
      url={https://arxiv.org/abs/1803.01937}, 
}

@misc{zhang2020bertscoreevaluatingtextgeneration,
      title={BERTScore: Evaluating Text Generation with BERT}, 
      author={Tianyi Zhang and Varsha Kishore and Felix Wu and Kilian Q. Weinberger and Yoav Artzi},
      year={2020},
      eprint={1904.09675},
      archivePrefix={arXiv},
      primaryClass={cs.CL},
      url={https://arxiv.org/abs/1904.09675}, 
}

@misc{maa2024aime,
  author       = {{Mathematical Association of America}},
  title        = {{American Invitational Mathematics Examination (AIME) 2024}},
  month        = feb,
  year         = {2024},
  howpublished = {\url{https://maa.org/math-competitions/american-invitational-mathematics-examination-aime}},
  note         = {Accessed: 2025-05-16}
}

@misc{clark2018thinksolvedquestionanswering,
      title={Think you have Solved Question Answering? Try ARC, the AI2 Reasoning Challenge}, 
      author={Peter Clark and Isaac Cowhey and Oren Etzioni and Tushar Khot and Ashish Sabharwal and Carissa Schoenick and Oyvind Tafjord},
      year={2018},
      eprint={1803.05457},
      archivePrefix={arXiv},
      primaryClass={cs.AI},
      url={https://arxiv.org/abs/1803.05457}, 
}

@misc{geng2025zip2zipinferencetimeadaptivetokenization,
      title={zip2zip: Inference-Time Adaptive Tokenization via Online Compression}, 
      author={Saibo Geng and Nathan Ranchin and Yunzhen yao and Maxime Peyrard and Chris Wendler and Michael Gastpar and Robert West},
      year={2025},
      eprint={2506.01084},
      archivePrefix={arXiv},
      primaryClass={cs.CL},
      url={https://arxiv.org/abs/2506.01084}, 
}

@misc{li2024500xcompressorgeneralizedpromptcompression,
      title={500xCompressor: Generalized Prompt Compression for Large Language Models}, 
      author={Zongqian Li and Yixuan Su and Nigel Collier},
      year={2024},
      eprint={2408.03094},
      archivePrefix={arXiv},
      primaryClass={cs.CL},
      url={https://arxiv.org/abs/2408.03094}, 
}

@misc{liskavets2024promptcompressioncontextawaresentence,
      title={Prompt Compression with Context-Aware Sentence Encoding for Fast and Improved LLM Inference}, 
      author={Barys Liskavets and Maxim Ushakov and Shuvendu Roy and Mark Klibanov and Ali Etemad and Shane Luke},
      year={2024},
      eprint={2409.01227},
      archivePrefix={arXiv},
      primaryClass={cs.CL},
      url={https://arxiv.org/abs/2409.01227}, 
}

@misc{xu2025procutllmpromptcompression,
      title={ProCut: LLM Prompt Compression via Attribution Estimation}, 
      author={Zhentao Xu and Fengyi Li and Albert Chen and Xiaofeng Wang},
      year={2025},
      eprint={2508.02053},
      archivePrefix={arXiv},
      primaryClass={cs.CL},
      url={https://arxiv.org/abs/2508.02053}, 
}

@misc{fei2025efficientpromptcompressionevaluator,
      title={Efficient Prompt Compression with Evaluator Heads for Long-Context Transformer Inference}, 
      author={Weizhi Fei and Xueyan Niu and Guoqing Xie and Yingqing Liu and Bo Bai and Wei Han},
      year={2025},
      eprint={2501.12959},
      archivePrefix={arXiv},
      primaryClass={cs.CL},
      url={https://arxiv.org/abs/2501.12959}, 
}

@misc{chuang2024learningcompresspromptnatural,
      title={Learning to Compress Prompt in Natural Language Formats}, 
      author={Yu-Neng Chuang and Tianwei Xing and Chia-Yuan Chang and Zirui Liu and Xun Chen and Xia Hu},
      year={2024},
      eprint={2402.18700},
      archivePrefix={arXiv},
      primaryClass={cs.CL},
      url={https://arxiv.org/abs/2402.18700}, 
}

@misc{zakazov2026cmprsrabstractivetokenlevelquestionagnostic,
      title={Cmprsr: Abstractive Token-Level Question-Agnostic Prompt Compressor}, 
      author={Ivan Zakazov and Berke Argin and Oussama Gabouj and Kamel Charaf and Alexander Sharipov and Alexi Semiz and Lorenzo Drudi and Nicolas Baldwin and Robert West},
      year={2026},
      eprint={2511.12281},
      archivePrefix={arXiv},
      primaryClass={cs.CL},
      url={https://arxiv.org/abs/2511.12281}, 
}

@inproceedings{li2025text,
  title={Text or Pixels? Evaluating Efficiency and Understanding of LLMs with Visual Text Inputs.},
  author={Li, Yanhong and Lan, Zixuan and Zhou, Jiawei},
  booktitle={EMNLP (Findings)},
  pages={10564--10578},
  year={2025}
}

@misc{lan2026reducedmatrixmultiplicationinputadaptive,
      title={Reduced Matrix Multiplication: Input-Adaptive Matrix-Product Reduction for LLM Inference}, 
      author={Zixuan Lan and Yanhong Li and Jiawei Zhou},
      year={2026},
      eprint={2608.13426},
      archivePrefix={arXiv},
      primaryClass={cs.LG},
      url={https://arxiv.org/abs/2608.13426}, 
}
\newpage
\appendix
\onecolumn
\clearpage

\newpage

\begin{center}
    \Large{\textbf{Technical Appendices}}
\end{center}

\startcontents[appendix]
\printcontents[appendix]{}{1}{}{\setcounter{tocdepth}{2}}

\vspace{1em}

\section{Empirical Study}
\label{Research Motivation}

\subsection{Empirical Summary}
\label{app:empirical_summary}

We will subsequently provide additional content to supplement the main text in the following sections for UMIM. The central question of this work is whether the sequential computation of multiple adjacent tokens in a frozen autoregressive language model can be approximated by a single learned input-level representation. This is different from token deletion, prompt rewriting, hidden-state pruning, or introducing a new tokenizer. UMIM keeps the pretrained language model unchanged: the tokenizer, vocabulary, model architecture, and backbone parameters are all fixed. Only a lightweight merge module is trained to map selected token spans into single surrogate embeddings.

The main empirical message of the paper is that input-level token-span merging can substantially reduce the effective sequence length while preserving the behavior of the original model to a large extent. The merge module operates before Transformer computation, using only static token embeddings as input. After training, the resulting surrogate embedding can replace a short multi-token span as a single computational unit. This allows the model to process compressed prompts and compressed decoding contexts without changing the underlying language model. In this sense, UMIM is not merely a heuristic for shortening text; it is a learned input-level approximation of multi-step token computation in frozen Transformer language models.

A key strength of UMIM is that the compression is achieved with strong performance preservation. Across language modeling and downstream tasks, the compressed model remains close to the uncompressed base model while reducing the effective sequence length. This is important because many efficient inference methods obtain speed or memory savings by deleting tokens, dropping context, or compressing information in a task-dependent way, often causing a visible accuracy degradation. UMIM instead attempts to preserve the functional role of the merged span: the span is still represented, but its multi-step computation is replaced by a single learned surrogate embedding. This design explains why UMIM can reduce sequence length and KV-cache usage while maintaining strong predictive and downstream performance.

Another important finding is that the merge module trained on a general corpus can transfer directly to downstream tasks. In the task-agnostic setting, the UMIM module trained on WikiText-103 is applied to unseen tasks without updating the backbone language model and without task-specific supervision. Even under this direct-transfer setting, UMIM achieves a strong accuracy--compression trade-off and consistently outperforms representative prompt-compression and context-reduction baselines at comparable token reduction rates. This result suggests that the learned module is not simply memorizing a fixed training corpus; rather, it learns a reusable span-to-surrogate mapping that can be consumed by frozen language models across different tasks.

The distinction between merge rules and the merge module is central to the method. Frequency-based rules decide where merging is attempted, but they do not define the compressed representation itself. The compressed representation is produced by the learned merge module. Therefore, UMIM should not be understood as only an n-gram compression heuristic. The rules provide stable and recurring candidate spans, while predictive distillation teaches the module how to replace the original multi-token computation with a single embedding that preserves the language model's output behavior. This separation makes the framework both reusable and adaptable: the pretrained module can be reused across domains, while the rule set can be reconstructed when a target task requires higher merge coverage.

The task-specific adaptation experiments further demonstrate this point. Starting from the WikiText-trained merge module, we can rebuild task-specific merge rules and then adapt only the merge module through lightweight fine-tuning and preference optimization. The backbone language model remains frozen throughout this process. This setting is substantially more practical than retraining or modifying the full language model. More importantly, the adapted module can achieve a stronger accuracy--compression trade-off than both the directly transferred module and external compression baselines. In some settings, the adapted UMIM variant can preserve or even exceed the accuracy of the original uncompressed base model while still reducing the effective sequence length by a large margin. This indicates that UMIM is not only a compression method, but also a practical framework for learning reusable and adaptable input-level approximations of sequential computation.

\subsection{LLM Background}

Modern language models generate text in an autoregressive manner, where tokens are processed and generated sequentially. This sequential process leads to high inference cost, especially when the input or generated sequence becomes long. Motivated by this practical limitation, we explore whether multiple tokens can be represented by a single learned embedding during inference. The goal is to reduce token-level computation while preserving the model's original predictive behavior as much as possible.

In the following subsections, we describe our early attempts and empirical observations, which may provide valuable experience.

\subsection{Early Attempt One: Embedding Similarity-Based Token Merging}

In the early stages of this work, we explored token merging from a purely representation-based and linguistically motivated perspective. Since our goal was to reduce redundant token-level computation without introducing additional training or modifying the language model, a natural first attempt was to identify semantically similar tokens and merge them directly at the embedding level.

Concretely, given a token sequence $(x_1, x_2, x_3, x_4)$, we considered computing the cosine similarity between adjacent token embeddings, such as $e(x_2)$ and $e(x_3)$. If the similarity exceeded a predefined threshold, the two tokens were regarded as semantically redundant and merged into a single representation, typically by simple averaging of their embeddings. This approach was motivated by the intuition that semantically similar tokens may play interchangeable roles in language understanding, and that their embeddings—learned by large language models—should reflect this similarity.

Importantly, this design deliberately avoided any form of training or supervision. The merging decision relied solely on static token embeddings and geometric similarity measures, making it lightweight and model-agnostic. From a linguistic and representation-learning standpoint, this approach appears reasonable and aligns with common practices in embedding-based clustering and semantic similarity analysis.

However, empirical results quickly revealed that this strategy failed to preserve model behavior in practice. Even when merging token pairs with relatively high cosine similarity, the resulting generations often deviated noticeably from the original outputs. In some cases, merging led to abrupt changes in the predicted next-token distribution, indicating that semantic proximity in embedding space was insufficient to guarantee computational substitutability within the language model.

Upon further analysis, we identified several fundamental limitations of this approach. First, cosine similarity in high-dimensional embedding spaces lacks a stable and transferable threshold. The similarity distribution is highly concentrated, making it difficult to distinguish genuinely mergeable token pairs from unrelated ones in a robust manner. Small variations in threshold choice often resulted in either excessive merging or negligible compression.

More fundamentally, semantic similarity between token embeddings does not imply equivalence in their computational roles within a Transformer. Token embeddings serve only as inputs to a highly nonlinear and context-dependent computation pipeline involving positional encodings, multi-head attention, and deep feed-forward transformations. Two tokens that appear similar in embedding space may interact very differently with surrounding context, attend to different positions, or trigger distinct internal computation patterns. As a result, replacing multiple tokens with an averaged embedding introduces a distribution shift that the model was never trained to handle.

\subsection{Early Attempt 2: Training-Based Pairwise Token Merging}

After observing that purely heuristic or geometry-based token merging strategies fail to reliably preserve model behavior, we naturally shifted to a training-based perspective. From this viewpoint, it is reasonable to hypothesize that there exists a learnable \emph{super embedding} in the embedding space that can replace multiple token embeddings and approximate their joint computational effect. The theoretical motivation behind this assumption will be discussed in later sections.

In this stage, we began with the simplest and most controlled setting: \emph{pairwise token merging}. Given an input sequence
$(x_1, x_2, x_3, x_4)$, we deterministically merged $(x_1, x_2)$ into a single token and $(x_3, x_4)$ into another, thereby reducing the sequence length by half while keeping the merging rule fixed. This design avoids the complexity of dynamic span selection and allows us to isolate the learnability of the merging operation itself.

As an initial sanity check, we froze a GPT-2 model and trained only a lightweight merging module on a small number of sampled sentences. Even under this minimal setup, we observed that the merged representations could approximate the original next-token predictive distributions after a few training steps. This result indicates that, in principle, such super embeddings do exist and can be learned under constrained conditions.

We then extended this experiment to a more realistic setting using the IMDB dataset. In this configuration, input sequences of length 512 were uniformly compressed via pairwise merging into 256 tokens. At this scale, the problem shifted from feasibility to stability and generalization. We found that the expressive capacity of the merging module became a critical factor. To achieve satisfactory distribution alignment with the frozen GPT-2 model, the merging module required substantially increased capacity. Consequently, we employed a deep, over-parameterized linear network with nonlinear activations, LayerNorm, and residual connections to transform the merged embeddings back to the original embedding dimension. While this design lacks theoretical elegance, it represented the most effective empirical solution under the constraints of this experimental setup.

Regarding the training objective, we used distribution-level alignment metrics consistent with those in the main experiments (e.g., Top-1 and Top-$p$ overlap). A notable empirical observation was that using KL divergence alone for distillation often resulted in unstable training or poor convergence. In this setting, the merging module was required to directly approximate the predictive distribution induced by two consecutive decoding steps, leading to noisy and weak optimization signals. To mitigate this issue, we introduced an additional cosine similarity constraint at the embedding or sentence level, providing a geometric regularization that significantly improved training stability.

It is important to emphasize that the design choices in this stage—including fixed pairwise merging, a highly over-parameterized merging network, and the combination of KL-based distillation with cosine regularization—were driven primarily by empirical considerations rather than theoretical optimality. Although this approach can yield reasonable results in controlled and moderate-scale settings, its reliance on fixed segmentation and excessive parameterization exposes clear limitations in scalability and generalization. These observations motivated the search for a more general and structurally grounded merging formulation, which we present in the subsequent sections.

\subsection{Foundation to our current Work}

After extensive exploration and empirical refinement, we arrive at the current merging formulation. The concrete architectural and implementation details have already been described thoroughly in the main text and appendices. In this section, we focus instead on a more fundamental question: \emph{why this approach is expected to work in principle}. Our goal is to clarify that the method is not a heuristic construction, but rather the result of parallel theoretical reasoning and engineering validation.

We begin with the construction of the merge set. Any token merging strategy must first address a basic question: \emph{which token pairs or spans should be merged}. This question is closely tied to the tokenization mechanisms employed by modern language models.

Most contemporary language models rely on Byte Pair Encoding (BPE) or its variants for tokenization. BPE is not designed to recover linguistically complete units; instead, it is a statistically driven compression algorithm. Starting from atomic units (characters or bytes), BPE repeatedly merges the most frequent adjacent token pairs in the training corpus, aiming to reduce sequence length while maintaining a compact vocabulary. As a result, the segmentation produced by BPE is governed by frequency statistics rather than semantic or linguistic completeness.

While highly effective in practice, this process often leads to the fragmentation of expressions that humans naturally perceive as single semantic units. For example, adverbs such as \emph{excitedly} may be decomposed into \emph{excited} and \emph{ly}, and common multi-word expressions such as \emph{machine learning} or \emph{love to do something} are represented as multiple tokens. From a linguistic or cognitive perspective, these expressions typically function as indivisible units, yet they are split due to the statistical nature of the tokenizer.

Our approach does not seek to revise or replace BPE, but rather to complement it at inference time. Specifically, we ask whether frequently occurring composite expressions that are split by the tokenizer can be re-aggregated into larger units during model execution. This motivates the use of high-frequency $n$-grams as candidates for merging. Such a choice can be interpreted both as compensating for the side effects of BPE segmentation and as explicitly modeling structures that the language model has already encountered repeatedly during training.

\paragraph{A functional approximation view.}
The existence of a useful surrogate embedding can be understood through the
same approximation principle that underlies ordinary model training. A
language model does not have direct access to the true data-generating
distribution. Instead, its parameters are learned from empirical observations:

\begin{equation}
\theta^*
=
\arg\min_\theta
\mathbb{E}_{(x,y)\sim p_{\mathrm{data}}}
\left[-\log p_\theta(y\mid x)\right].
\end{equation}

Once pretrained, the frozen model $p_\theta$ defines a stable and observable
mapping from an input sequence to a sequence of predictive distributions.
UMIM introduces a second approximation problem. Rather than learning
$p_\theta$ from data, it keeps $p_\theta$ fixed and learns a merge module
$M_\phi$ whose compressed inputs reproduce the behavior of $p_\theta$ on the
original sequence.

Let $\widetilde{x}=C_\phi(x;\mathcal{R})$ denote the sequence obtained by
replacing spans selected by the merge rules $\mathcal{R}$ with surrogate
embeddings
\begin{equation}
z_s=M_\phi(e(s)).
\end{equation}
If $\mathcal{A}(x)$ denotes the alignment between retained positions in the
original and compressed sequences, the merge module is trained according to

\begin{equation}
\phi^*
=
\arg\min_\phi
\mathbb{E}_{x\sim\mathcal{D}}
\left[
\sum_{(t,\widetilde{t})\in\mathcal{A}(x)}
D_{\mathrm{KL}}
\left(
p_\theta(\cdot\mid x_{\leq t})
\;\middle\|\;
p_\theta(\cdot\mid\widetilde{x}_{\leq\widetilde{t}})
\right)
\right].
\end{equation}

The underlying hypothesis is therefore not that a single embedding exactly
reconstructs every intermediate state produced by a multi-token span. Rather,
we hypothesize that, over the contexts in which a recurring span appears,
there exists a compact input representation that approximately preserves the
span's functional effect on the frozen model's subsequent predictive
distributions. UMIM learns this representation directly from the behavior of
the uncompressed model.

High-frequency spans are particularly suitable for this approximation because
they provide repeated observations across many contexts. For a span $s$, its
surrogate representation is effectively optimized over the empirical context
distribution in which $s$ occurs. Frequent spans provide more samples from this
distribution, making the expected behavioral effect of replacing $s$ better
specified and more reliably estimable. The shared merge module further
amortizes this approximation across different spans, rather than learning an
independent embedding for every entry in the rule set.

From this perspective, UMIM performs functional distillation at the input
level. It does not require the surrogate embedding to reproduce the original
token embeddings or all of their intermediate hidden states. It only requires
the compressed sequence to preserve the predictive behavior that matters to
the frozen language model.

\subsection{Information Preservation and Approximation Scope}
\label{app:information_preservation}

UMIM does not claim that a surrogate embedding can losslessly replace the original token span in every context. Nor does it require every hidden state, attention score, or KV state to remain identical before and after compression. We instead adopt an operational definition of information preservation: with the backbone language model frozen, the compressed sequence should preserve the predictive behavior of the original sequence at subsequent positions as closely as possible. Thus, UMIM targets end-to-end predictive behavior rather than exact layer-wise equivalence.

UMIM also differs from direct token deletion. For each merged span, the embeddings of all constituent tokens are provided to the merge module and jointly used to produce the surrogate embedding. Their information is therefore not discarded without processing, but compressed through a trainable mapping. During training, we compare the predictive distributions of the original and compressed sequences across many naturally occurring contexts. Distillation then teaches the surrogate to approximate the aggregate effect of the complete span on subsequent predictions. High-frequency $n$-grams are selected primarily because they provide more training observations across contexts, not because frequent spans are assumed to be unimportant or semantically empty.

Although the same $n$-gram produces the same static surrogate embedding at the input level, its role inside the model is not fully context-independent. Once the surrogate is passed into the frozen Transformer, it interacts with surrounding tokens through positional encoding, self-attention, MLP layers, and subsequent Transformer computation. Different contexts can therefore produce different hidden states, attention interactions, and downstream effects from the same initial surrogate. Context independence applies to the initial output of the merge module, not to the complete contextual computation performed by the Transformer.

Our experiments examine this preservation at several levels. The distribution-alignment metrics directly measure agreement between the original and compressed predictive distributions at aligned positions. Table~3 provides qualitative examples in which generation behavior is preserved after merging, while Figure~4 shows that surrogate states remain involved in subsequent attention rather than being ignored. The PPL, QA, and summarization experiments further evaluate preservation through downstream behavior. The task-adaptation results also show that when a distribution shift exists between the general-purpose merge module and a target task, applying SFT and RL only to the small merge module can further reduce the performance loss caused by merging while keeping the backbone LLM frozen.

At the same time, we acknowledge a structural limit of this approximation. After multiple token-level KV states are compressed into one surrogate state, a future query can no longer access every constituent token independently in exactly the same way as in the original model. We therefore do not claim that a context-independent surrogate preserves every internal computation or task-critical detail in every context. Highly ambiguous, strongly context-dependent, or task-critical spans may introduce greater approximation error. This helps explain the remaining performance gap in some task-agnostic evaluations and why the approximation can become more difficult under extreme length and domain shifts.

Whether a token is critical is itself dependent on the context, query, and downstream task, and frequency-based rules are not semantic-importance detectors. Our claim is therefore not lossless semantic compression, but an effective functional approximation over the observed data distribution. For settings requiring more conservative treatment of critical content, the framework can use stricter merge rules, a higher frequency threshold, task-defined protected spans, or task-specific rules. The merge module can also be further adapted through SFT and RL. These choices provide different ways to control the accuracy--compression trade-off.

In summary, UMIM does not guarantee lossless semantic replacement for every span and context. It processes all constituent token embeddings and distills their aggregate predictive effect under the frozen LLM, thereby providing practical information compression with substantial behavioral preservation. Our results show that this approximation is effective across the evaluated models and tasks, while the remaining errors define the current scope of the method and motivate conservative rule construction and task adaptation.

\subsection{Generalization of Context-Shared Surrogate Embeddings}
\label{app:surrogate_generalization}

The preceding discussion defines what UMIM aims to preserve and clarifies the scope of this approximation, but leaves open an important question: if a surrogate is optimized on a finite set of observed contexts, why should it preserve the frozen model's predictive behavior in previously unseen contexts? We provide a conditional
generalization analysis that relates this discrepancy to the frequency support
used during rule construction.

Let $g$ be a candidate $n$-gram and let $c$ denote a context in which $g$
occurs. The frozen teacher model produces the predictive distribution
$p_\theta(\cdot\mid c,g)$ from the original sequence. Replacing $g$ with
$M_\phi(e(g))$ produces the corresponding compressed-model distribution
$p_{\theta,\phi}(\cdot\mid c,g)$. We define the behavioral approximation loss as

\begin{equation}
\ell_\phi(c,g)
=
D_{\mathrm{KL}}
\left(
p_\theta(\cdot\mid c,g)
\;\middle\|\;
p_{\theta,\phi}(\cdot\mid c,g)
\right).
\end{equation}

When distillation is applied at multiple aligned positions, $\ell_\phi(c,g)$
denotes the average KL divergence over those positions. The soft-target
cross-entropy used during training satisfies

\begin{equation}
\operatorname{CE}
\left(p_\theta,p_{\theta,\phi}\right)
=
H(p_\theta)
+
D_{\mathrm{KL}}
\left(p_\theta\|p_{\theta,\phi}\right).
\end{equation}

Because the teacher and its entropy are fixed with respect to $\phi$,
minimizing the soft-target cross-entropy is equivalent to minimizing the
forward KL divergence.

For a fixed $g$, suppose that its $N_g$ observed training contexts are
$c_1,\ldots,c_{N_g}$. Its empirical and population risks are

\begin{equation}
\widehat{R}_g(\phi)
=
\frac{1}{N_g}
\sum_{i=1}^{N_g}
\ell_\phi(c_i,g),
\qquad
R_g(\phi)
=
\mathbb{E}_{c\sim P(c\mid g)}
\left[\ell_\phi(c,g)\right].
\end{equation}

We make the following standard assumptions for the analysis: for each fixed
$g$, the observed contexts are modeled as independent draws from
$P(c\mid g)$; the loss satisfies $0\leq\ell_\phi(c,g)\leq B$; and the
loss class induced by the capacity-controlled merge-module family $\Phi$
satisfies

\begin{equation}
\widehat{\mathfrak{R}}_{N_g}
\left(\mathcal{L}_{\Phi,g}\right)
\leq
\frac{C_\Phi}{\sqrt{N_g}}.
\end{equation}

The bounded-loss condition can be implemented through probability clipping, or
viewed as an explicit bounded-logit assumption over the distributions included
in the analysis. Let $U$ be a fixed finite universe of candidate $n$-grams
defined before observing the training corpus, and let

\begin{equation}
G_\tau=\{g\in U:N_g\geq\tau\}
\end{equation}

be the set retained by a frequency threshold $\tau$.

A standard Rademacher-complexity argument, followed by a union bound over
$U$, implies that, with probability at least $1-\delta$, simultaneously for
every $g\in G_\tau$ and every $\phi\in\Phi$,

\begin{equation}
R_g(\phi)
\leq
\widehat{R}_g(\phi)
+
2\widehat{\mathfrak{R}}_{N_g}
\left(\mathcal{L}_{\Phi,g}\right)
+
3B
\sqrt{
\frac{\log(2|U|/\delta)}
     {2N_g}
}.
\end{equation}

Using the assumed capacity bound and $N_g\geq\tau$ gives

\begin{equation}
R_g(\phi)
\leq
\widehat{R}_g(\phi)
+
\frac{
2C_\Phi+
3B\sqrt{\log(2|U|/\delta)/2}
}{
\sqrt{\tau}
}.
\label{eq:frequency_generalization}
\end{equation}

Equation~\ref{eq:frequency_generalization} separates two sources of error.
The empirical term $\widehat{R}_g(\phi)$ contains the approximation error
incurred when a multi-token computation is represented by a single surrogate.
This term may remain nonzero even with unlimited observations. The second term
captures statistical uncertainty from learning the surrogate using finitely
many contexts and decreases as the minimum frequency support $\tau$ increases.

Pinsker's inequality additionally gives

\begin{equation}
\operatorname{TV}
\left(
p_\theta,p_{\theta,\phi}
\right)
\leq
\sqrt{
\frac{1}{2}
D_{\mathrm{KL}}
\left(
p_\theta\|p_{\theta,\phi}
\right)
}.
\end{equation}

Together with Jensen's inequality, this implies

\begin{equation}
\mathbb{E}_{c}
\left[
\operatorname{TV}
\left(
p_\theta,p_{\theta,\phi}
\right)
\right]
\leq
\sqrt{\frac{R_g(\phi)}{2}}.
\end{equation}

The result is deliberately conditional. It does not establish that one
context-independent surrogate exactly reproduces every state associated with a
multi-token span, nor does it guarantee small approximation error for an
arbitrary span. Instead, it shows that when the merge module fits the observed
teacher behavior well, stronger frequency support provides tighter control of
its expected discrepancy on unseen contexts drawn from the same distribution.
The frequency-threshold ablation in
Section~\ref{sec:ablation_study} empirically evaluates the qualitative
prediction of this analysis.

\subsection{UMIM as Parameter-Efficient Task Adaptation}
\label{app:parameter_efficient_adaptation}

The preceding analysis considers generalization to unseen contexts drawn from a similar distribution. In downstream tasks, however, the context distribution, frequent token spans, and task objective may differ from those observed in WikiText-103. UMIM addresses this distribution shift through lightweight task adaptation. The WikiText-trained base merge module serves as a reusable initialization that has already learned a general span-to-surrogate mapping. Given a target task, we first reconstruct task-specific merge rules from its training split to increase merge coverage. If no additional training is desired, the new rules can be directly used with the fixed base merge module, corresponding to the \texttt{WT + Rules} setting.

For further adaptation, \texttt{WT + FT} initializes from the WikiText-trained base merge module and performs distillation-based SFT on the target-task training corpus. This is not conventional label-only fine-tuning. The frozen backbone processes the original uncompressed sequence to provide the teacher distribution and processes the compressed sequence to produce the student distribution. Their predictions are aligned at valid positions, and only the merge module is updated to minimize the distribution-matching loss. The \texttt{Task-Adapted} setting then starts from this SFT checkpoint and further applies RL to the merge module using the downstream task signal. The backbone LLM remains frozen throughout all stages.

This formulation makes UMIM a parameter-efficient task-adaptation method in addition to a token-compression method. Similar to parameter-efficient fine-tuning methods such as LoRA, UMIM updates only a small set of additional parameters while leaving the pretrained backbone unchanged. However, rather than inserting trainable parameters into the internal Transformer layers, UMIM adapts an external input-level merge module. It therefore provides two forms of efficiency simultaneously: only a small module is trained, and the adapted model can process a shorter effective sequence with reduced KV-cache usage. Task adaptation can consequently improve the accuracy--compression trade-off in both directions: task-specific rules increase the achievable merge ratio, while SFT and RL help preserve or improve task performance under this more aggressive compression.

The base merge module is important to this process. The \texttt{Task-trained} ablation in Figure~3 trains a new merge module directly from the downstream corpus, whereas \texttt{WT + FT} and \texttt{Task-Adapted} reuse the WikiText-trained initialization. Their stronger performance indicates that the base module provides reusable compression knowledge that cannot always be learned reliably from a smaller task-specific corpus alone. We do not claim that adaptation must outperform the uncompressed model on every task. Rather, Figure~3 and Appendix Figure~5 show that, across the three evaluated tasks, adapting only the merge module can support substantially higher merge ratios while preserving and, in some settings, improving downstream accuracy.

\subsection{Choice of $n$-gram Lengths}
\label{app:ngram_length}

This subsection clarifies why the current implementation of {\namem} restricts
candidate spans to $n$-grams of length 2--4. This range reflects a practical
choice based on statistical support, training feasibility, and the timing of
compression during autoregressive decoding. It is not a fundamental limitation
of the merge-module architecture.

Our base merge module is trained on WikiText-103, from which we extract
high-frequency contiguous token spans. For a fixed corpus, the number of
possible $n$-grams grows rapidly with $n$, while the number of repeated
occurrences of any particular span generally decreases. Longer spans therefore
tend to provide fewer contextual observations for learning a reliable
surrogate. Restricting the rule set to 2--4 grams allows each retained span to
satisfy a meaningful minimum-frequency threshold while keeping the number of
candidate rules manageable.

The current implementation also uses a single-stage, non-overlapping merge
assignment. Overlap is resolved during rule construction, so the merge rules
applied during training and inference do not compete for the same token
positions. For example, given a sequence
$(x_1,x_2,x_3,x_4)$, if $(x_2,x_3)$ is selected and replaced by a surrogate
$x_{23}$, the resulting surrogate is not recursively merged with $x_4$.
Representing $(x_2,x_3,x_4)$ with one surrogate instead requires the complete
3-gram to be selected and merged in a single operation.

Recursive or hierarchical merging is conceptually possible, but it would define
a different learning problem. The representation of a span would depend on the
order in which its sub-spans were merged, producing dynamic and path-dependent
merge structures. Supporting these structures would complicate rule assignment,
teacher--student alignment, and efficient batch construction. We therefore use
a fixed non-overlapping assignment so that every compressed sequence and every
distillation target can be determined before the forward pass.

Shorter spans also provide favorable statistical coverage. A longer recurring
expression may contain shorter sub-spans that already satisfy the merge
threshold. Consequently, parts of the same region may already be compressed
through shorter, non-overlapping matches without explicitly retaining the
complete longer expression as an additional rule. Extending the rule set to
higher-order $n$-grams may therefore provide diminishing additional coverage,
while increasing the number of candidates that must be counted, stored, and
observed sufficiently often during training. We treat this as a statistical
and computational prior rather than as a claim that all longer expressions can
be represented equivalently by shorter merges.

The timing of a match is also important during autoregressive decoding. A span
can be merged only after the token IDs of the complete span have been generated.
A bigram can therefore be detected immediately after its second token becomes
available. Once merged, it reduces the effective sequence length and KV-cache
length used by subsequent decoding steps. A 5-gram, by contrast, cannot be
matched until all five token IDs have been generated. Before that point, some
parts of the same region may already have become eligible for shorter,
non-overlapping matches. Longer rules can thus delay the point at which
compression begins, without necessarily providing a proportional increase in
token reduction.

The merge module itself can process variable-length spans and does not
conceptually require $n\leq4$. The framework could therefore be extended to
longer $n$-grams, or to recursive merging with an appropriately redesigned
training procedure. In the present work, however, 2--4 grams provide a balanced
trade-off among contextual support, rule-set size, batch-level training,
non-overlapping assignment, and early compression during decoding. We therefore
adopt this range as the default operating regime of {\namem}, rather than as a
theoretical upper bound on mergeable span length.

\subsection{Conclusion}

Having presented the methodology, empirical development, and theoretical analysis of {\namem}, we conclude by clarifying the broader perspective of this work. Our goal is not to claim a single definitive strategy for token merging, but to introduce and validate a different way of thinking about token-level computation in frozen language models. Our experiments show that a learned input-level surrogate can replace selected multi-token spans while approximately preserving the predictive behavior of the original model. This provides empirical evidence that part of the sequential computation associated with recurring token spans can be compressed into a single learned representation.

The merge module and frequency-based rule construction used in this paper represent one practical and empirically effective implementation of this idea, rather than its only possible realization. The rules identify spans with sufficient statistical support, while predictive distillation trains the merge module to approximate their functional effect on the frozen language model. Our theoretical discussion further clarifies why stronger frequency support can improve generalization across contexts, while also recognizing that a context-independent surrogate cannot guarantee exact equivalence for every span and context.

Beyond task-agnostic compression, {\namem} also provides a parameter-efficient approach to task adaptation. A base merge module trained on a general corpus can be reused directly or adapted to a downstream task by reconstructing task-specific merge rules, continuing distillation-based SFT, and optionally applying RL. Throughout this process, the backbone LLM remains frozen and only the lightweight merge module is updated. The resulting adaptation can support higher merge ratios while preserving or, in some settings, improving downstream accuracy. Thus, {\namem} provides efficiency in both training and inference: it requires updating only a small additional module while reducing the effective sequence length and KV-cache usage during inference.

Our current exploration remains limited by available training data and computational resources. Future work may investigate stronger merge criteria, context-dependent or hierarchical surrogate representations, long-context training, and more optimized inference implementations. Another promising direction is to incorporate token merging directly into language-model pretraining or to jointly optimize the merge mechanism and backbone, rather than treating merging only as a post-hoc component.

Overall, the contribution of this work is not limited to a particular merge module. It introduces a broader research direction in which token-level sequential computation is treated as a structure that can be learned, approximated, adapted, and compressed. We hope this perspective encourages further work on input-level computation reduction and parameter-efficient adaptation for more efficient and scalable language models.

\section{Merge-Rule Mining Details}
\label{sec:Data Preparation}

Given a corpus $\mathcal{C}$, we tokenize its documents using the tokenizer of the corresponding backbone model. The tokenized documents are concatenated and partitioned into non-overlapping segments for batched data processing. We use a segment length of $L=512$ for Llama-3.1-8B, Llama-3.2-1B, and GPT-2-XL, and $L=1024$ for DeepScaleR-1.5B-Preview. The following procedure constructs the merge rules used for training the base merge module.

\paragraph{N-gram Extraction and Frequency Counting}
Let $S=\{s_1,\ldots,s_N\}$ denote the resulting tokenized segments, where $s_i=(t_{i,1},\ldots,t_{i,L})$. For each segment, we enumerate all contiguous $n$-grams with $n\in\{2,3,4\}$. For an $n$-gram
\[
g=(t_{i,j},t_{i,j+1},\ldots,t_{i,j+n-1}),
\]
we compute its corpus-wide occurrence count
\[
f(g)=\sum_{i=1}^{N}\sum_{j=1}^{L-n+1}
\mathbbm{1}\!\left[
(t_{i,j},\ldots,t_{i,j+n-1})=g
\right].
\]
The frequency is aggregated across all segments in the training corpus.

\paragraph{Frequency-based Candidate Selection}
We retain an $n$-gram only if it occurs at least $\tau$ times:
\[
\mathcal{R}^{(n)}_{\tau}
=
\left\{
g:\lvert g\rvert=n,\; f(g)\geq\tau
\right\},
\qquad n\in\{2,3,4\}.
\]
Unless otherwise stated, the base merge rules use $\tau=5$. We vary this threshold in the frequency-threshold ablation to study the trade-off between merge coverage and the statistical support available for learning surrogate embeddings.

\paragraph{Intra-order Overlap Filtering}
Frequency thresholding may retain same-order rules that compete for overlapping positions. For example, two bigrams $(a,b)$ and $(b,c)$ can occur at consecutive positions and share token $b$. Analogously, two $n$-grams may overlap when the $(n-1)$-token suffix of one candidate equals the $(n-1)$-token prefix of another. For such competing candidates, we retain the rule with the larger corpus frequency. This filtering reduces systematic conflicts among rules of the same order while favoring candidates with stronger training support.

\paragraph{Cross-order Containment Filtering}
We next remove lower-order rules that are fully contained in retained higher-order rules. Specifically, a trigram is removed if it is a contiguous sub-span of a retained 4-gram, and a bigram is removed if it is a contiguous sub-span of a retained 3-gram or 4-gram. This gives priority to longer and more specific spans when the same local token pattern is represented at multiple orders. The resulting rule set is
\[
\mathcal{R}
=
\widetilde{\mathcal{R}}^{(2)}_{\tau}
\cup
\widetilde{\mathcal{R}}^{(3)}_{\tau}
\cup
\widetilde{\mathcal{R}}^{(4)}_{\tau},
\]
where $\widetilde{\mathcal{R}}^{(n)}_{\tau}$ denotes the filtered rules of order $n$.

\paragraph{Non-overlapping Assignment in a Sequence}
After constructing $\mathcal{R}$, we match its rules against each token sequence. All matched occurrences are sorted first by their starting position and then, for matches with the same starting position, by decreasing span length. We scan the sorted candidates from left to right and retain a match only if it does not overlap with a previously selected match. Therefore, every token belongs to at most one merged span, and all merge locations can be determined before the forward pass. The current implementation performs a single stage of merging and does not recursively merge previously constructed surrogate embeddings.

\paragraph{Task-specific Rule Adaptation}
The main task-agnostic experiments use the rules mined from WikiText-103 together with the WikiText-trained base merge module. No downstream-task data are used to reconstruct rules in this setting. In the task-adaptation experiments, we apply the same frequency-counting and non-overlapping assignment procedure to the training split of the target task. Reconstructing only the rule set corresponds to the \texttt{WT + Rules} setting, while \texttt{WT + FT} and \texttt{Task-Adapted} additionally adapt the WikiText-trained merge module through distillation-based SFT and RL, respectively.

\paragraph{Choice of 2- to 4-grams}
The restriction to 2- to 4-grams is a practical design choice rather than an architectural limitation. It reflects the statistical support of recurring spans, the additional coverage provided by longer rules, and the delay before a complete span becomes available during autoregressive decoding. A detailed discussion is provided in Section~\ref{app:ngram_length}.

\paragraph{Number of Retained Merge Rules}
Table~\ref{tab:merge_rule_counts} reports the final rule sets used to train the base merge modules. The rules for Llama-3.1-8B, Llama-3.2-1B, and GPT-2-XL are mined from WikiText-103. Llama-3.1-8B and Llama-3.2-1B use the same tokenizer and therefore share the same rule set. For DeepScaleR-1.5B-Preview, the rules are mined from our generated mathematical-reasoning corpus described in Section~\ref{Datasets}.

\begin{table}[t]
  \centering
  \caption{Number of retained merge rules used for base merge-module training. All rule sets use the frequency threshold $\tau=5$.}
  \label{tab:merge_rule_counts}
  \small
  \begin{tabular}{lrrr}
    \toprule
    \textbf{Model} & \textbf{2-grams} & \textbf{3-grams} & \textbf{4-grams} \\
    \midrule
    Llama-3.1-8B            & 216{,}700 & 684{,}793 & 1{,}266{,}080 \\
    Llama-3.2-1B            & 216{,}700 & 684{,}793 & 1{,}266{,}080 \\
    GPT-2-XL                & 218{,}415 & 702{,}570 & 1{,}241{,}657 \\
    DeepScaleR-1.5B-Preview &  35{,}691 & 212{,}367 & 1{,}665{,}423 \\
    \bottomrule
  \end{tabular}
\end{table}

\section{Module Architecture and Integration}
\label{sec:Implementation Details of Methods}

The {\namem} merge module is an attention-based pooling network that operates on static input embeddings before they enter the frozen Transformer. Given a selected span of $n$ token embeddings, the module produces one surrogate embedding with the same dimension as the backbone's input embeddings. The current experiments use $n\in\{2,3,4\}$, although the pooling architecture itself can process variable-length spans.

The merge module is trained separately for each backbone in the current work. Its architecture can be applied to different Transformer-based language models, but its parameters depend on the backbone's embedding dimension, embedding geometry, tokenizer, and predictive behavior. We therefore do not assume that one merge-module checkpoint transfers directly across different backbone models.

\subsection{Attention-based Pooling Architecture}

Let
\[
E_s =
[e(x_t),e(x_{t+1}),\ldots,e(x_{t+n-1})]^\intercal
\in\mathbb{R}^{n\times d}
\]
denote the static input embeddings of a selected token span
$s=(x_t,\ldots,x_{t+n-1})$, where $d$ is the embedding dimension of the frozen language model. Let $h$ be the number of pooling heads and $d_h=d/h$.

The span embeddings are first processed by a shared token projection:
\begin{equation}
P
=
\operatorname{LN}_{kv}
\left(
\operatorname{GELU}
\left(
\operatorname{Linear}_{kv}(E_s)
\right)
\right)
\in\mathbb{R}^{n\times d},
\end{equation}
where $\operatorname{Linear}_{kv}:\mathbb{R}^{d}\rightarrow\mathbb{R}^{d}$ has weight matrix
\[
W^{kv}\in\mathbb{R}^{d\times d}.
\]
The last dimension of $P$ is then divided across $h$ heads:
\[
P=[P_1,\ldots,P_h],
\qquad
P_i\in\mathbb{R}^{n\times d_h}.
\]

Each head has one input-independent learnable query
\[
q_i\in\mathbb{R}^{d_h}.
\]
Collectively, these queries form
\[
Q=[q_1;\ldots;q_h]\in\mathbb{R}^{h\times d_h}.
\]
For head $i$, the attention weights and pooled representation are
\begin{equation}
\alpha_i
=
\operatorname{softmax}
\left(
\frac{P_iq_i}{\sqrt{d_h}}
\right)
\in\mathbb{R}^{n},
\qquad
o_i
=
\alpha_i^\intercal P_i
\in\mathbb{R}^{d_h}.
\end{equation}

Although $Q$ is independent of the input embeddings, the attention weights remain input-dependent through $P_i$. Consequently, different token spans produce different attention distributions and different surrogate embeddings.

The outputs of all heads are concatenated and passed through an output processor:
\begin{equation}
z_s
=
\operatorname{LN}_{o}
\left(
\operatorname{GELU}
\left(
\operatorname{Linear}_{o}
\left(
[o_1;\ldots;o_h]
\right)
\right)
\right)
\in\mathbb{R}^{d},
\end{equation}
where the output projection has
\[
W^{o}\in\mathbb{R}^{d\times d}.
\]
Because $z_s$ has the same dimension as an ordinary token embedding, it can be directly passed to the frozen backbone as one input-level computational unit.

\subsection{Difference from Standard Self-Attention}

The merge module is better understood as multi-head attention pooling rather than a standard Transformer self-attention layer. Standard self-attention produces input-dependent queries, keys, and values for every token and returns an output sequence with the same length as its input. In contrast, the merge module uses one learned query per head and reduces an entire span to a single vector.

The projected span representation $P_i$ is shared as both the key and value for head $i$. This shared key-value design keeps the module compact and is sufficient for pooling short spans. In our empirical exploration, introducing separate key and value projections or additional linear layers did not provide a clear improvement, while increasing the number of trainable parameters. We therefore use a shared $W^{kv}$ followed by the output projection $W^o$.

These architectural choices are empirical rather than theoretically optimal. Among the alternatives we explored, the current design provided a favorable balance among predictive alignment, training stability, and parameter efficiency.

\subsection{Parameter Count}

The learnable parameters of the merge module are
\[
\phi
=
\left\{
Q,
W^{kv},
b^{kv},
W^o,
b^o,
\operatorname{LN}_{kv},
\operatorname{LN}_o
\right\}.
\]
The query matrix contains $d$ parameters. Each of the two linear layers contains $d^2+d$ parameters, and each LayerNorm contains $2d$ parameters. The total parameter count is therefore
\begin{equation}
|\phi|
=
2d^2+7d
\approx
2d^2.
\end{equation}
For Llama-3.1-8B, where $d=4096$, the merge module contains approximately $33.6$ million parameters, corresponding to about $0.42\%$ of an 8-billion-parameter backbone. During both base training and task adaptation, the backbone remains frozen and only these merge-module parameters are updated.

\subsection{Training-time Integration}

During training, the merge rules first identify a set of non-overlapping token spans. For every selected span $s$, the original embeddings $E_s$ are passed through the merge module and replaced by the resulting surrogate $z_s$. The remaining token embeddings retain their original order, producing a shorter embedding sequence that is passed to the frozen language model.

The uncompressed sequence is processed by the same frozen backbone to provide teacher predictive distributions. The compressed and uncompressed predictions are then aligned at valid positions after merging. The merge module is optimized through soft-target cross-entropy, which is equivalent to minimizing the forward KL divergence up to the fixed entropy of the teacher distribution. No backbone parameters are updated.

Because merge locations are non-overlapping and are determined before the forward pass, spans with the same length can be pooled in parallel. Training therefore does not require recursive merging or dynamic modification of intermediate Transformer hidden states.

\subsection{Prompt-time Integration}

For prompt processing, the input is tokenized using the unchanged backbone tokenizer. The resulting token IDs are matched against the merge-rule set, and non-overlapping occurrences are selected from left to right. If several rules begin at the same position, the longer span is preferred.

All selected spans are replaced at the embedding level before the prompt enters the Transformer. For example,
\[
[x_1,x_2,x_3,x_4]
\longrightarrow
[e(x_1),z_{23},e(x_4)],
\qquad
z_{23}=\merge(e(x_2),e(x_3)).
\]
The compressed sequence uses contiguous position indices. In this example, the three resulting input units receive positions $[0,1,2]$. The backbone then performs an ordinary forward pass over this shorter embedding sequence, producing a correspondingly shorter prompt KV cache.

\subsection{Decoding-time Integration}

During autoregressive decoding, {\namem} does not skip token-generation steps. After each new token ID is generated, we check whether the unmerged suffix of the generated token history matches a rule in $\mathcal{R}$. The matching procedure checks longer rules first. If no rule is matched, the new token embedding is processed normally and one KV state is appended to the cache.

If a newly completed suffix
\[
s=(x_{T-n+1},\ldots,x_T)
\]
matches a merge rule, the merge module computes
\[
z_s
=
\merge
\left(
e(x_{T-n+1}),\ldots,e(x_T)
\right).
\]
At this point, the first $n-1$ tokens of the suffix have already been processed and stored in the KV cache, while the newest token has just been generated. We therefore roll back the $n-1$ cached suffix states and process $z_s$ as one input embedding instead of processing the newest token separately. The frozen backbone then produces one surrogate KV state for the complete span.

This operation changes neither the generated token IDs nor the decoded text. It only shortens the effective sequence and KV-cache representation used by subsequent decoding steps. Previously created surrogate units are not recursively merged, ensuring that every merge corresponds directly to one rule-defined span of original token IDs.

\begin{lstlisting}[
language=Python,
caption={Multi-head pooling used by the merge module.},
label={lst:multiheadpool},
basicstyle=\ttfamily\footnotesize,
breaklines=true,
breakatwhitespace=true,
columns=fullflexible
]
# tokens: (batch_size, span_length, d_model)

processed = token_processor(tokens)
processed = processed.view(
    batch_size,
    span_length,
    num_heads,
    head_dim,
)
processed = processed.permute(0, 2, 1, 3)

# query: (num_heads, head_dim)
query = query.unsqueeze(0).unsqueeze(2)
query = query.expand(
    batch_size,
    num_heads,
    1,
    head_dim,
)

scores = einsum(
    "bhqd,bhkd->bhk",
    query,
    processed,
) / sqrt(head_dim)

weights = softmax(scores, dim=2).unsqueeze(-1)
pooled = (processed * weights).sum(dim=2)
pooled = pooled.reshape(batch_size, d_model)

output = output_processor(pooled)
\end{lstlisting}

In summary, the merge module provides an input-level mechanism for approximating the functional effect of selected multi-token spans. It produces backbone-compatible surrogate embeddings, requires no modification to the internal Transformer layers, and supports both prompt compression and dynamic KV-cache compression during autoregressive decoding.

\section{Predictive Distillation Loss and Alignment Metrics}
\label{Loss and Alignment Metric}

This section describes how we align the predictive distributions of the original and compressed sequences, optimize the merge module through distillation, and evaluate their predictive agreement.

\paragraph{Original and Compressed Prediction Paths.}
Let $F_{\theta}$ denote the frozen backbone language model and $M_{\phi}$ the trainable merge module. For an input sequence $X$, the teacher distribution is obtained from the original sequence:
\begin{equation*}
P^{T} = \operatorname{softmax}\left(F_{\theta}(X)\right).
\end{equation*}
The compressed path replaces each selected token span with a surrogate embedding produced by $M_{\phi}$:
\begin{equation*}
P^{S} = \operatorname{softmax}\left(F_{\theta}(M_{\phi}(X))\right).
\end{equation*}
The two paths share the same frozen backbone. Only $\phi$, the parameters of the merge module, is updated during training.

\paragraph{Mask Construction and Sequence Alignment.}
Token merging shortens the sequence and shifts subsequent positions. We therefore construct separate masks for the original teacher sequence and the compressed student sequence.

For a merged span of length $n$, the implementation first constructs an intermediate length-preserving layout containing $n-1$ inactive placeholders followed by one valid surrogate position. The surrogate is therefore aligned with the final position of the original span. The corresponding keep mask, denoted by $M^{\mathrm{keep}}$, marks retained tokens and surrogate positions as $1$, and inactive placeholders and padding positions as $0$.

Let $s_b$ be the start position of the first merged span in sample $b$. The teacher loss mask is
\begin{equation*}
M^{T}_{b,j}
=
M^{\mathrm{keep}}_{b,j}
\cdot
\mathbf{1}[j \geq s_b].
\end{equation*}
Thus, positions before the first merge are excluded because the original and compressed computations are identical before any surrogate embedding is introduced. Within each merged span, only its final position is used as the teacher-side alignment anchor.

The inactive placeholders are then removed to form the packed compressed sequence. Its attention mask, $M^{\mathrm{attn}}$, marks valid compressed positions. Let
\begin{equation*}
N_b = \sum_j M^{T}_{b,j},
\qquad
\widetilde{L}_b = \sum_j M^{\mathrm{attn}}_{b,j}.
\end{equation*}
The first $\widetilde{L}_b-N_b$ valid positions in the compressed sequence correspond to the unaffected prefix before the first merge. We exclude these positions to construct the student loss mask $M^{S}$. Consequently,
\begin{equation*}
\sum_j M^{T}_{b,j}
=
\sum_j M^{S}_{b,j}
=
N_b.
\end{equation*}

Teacher and student distributions are selected by their respective masks and packed in causal order:
\begin{equation*}
\overline{P}^{T}_{b}
=
\operatorname{Pack}(P^{T}_{b},M^{T}_{b}),
\qquad
\overline{P}^{S}_{b}
=
\operatorname{Pack}(P^{S}_{b},M^{S}_{b}).
\end{equation*}
Here, $\operatorname{Pack}$ selects all positions marked by the mask while preserving their order. Therefore, no explicit original-to-compressed index table is required. The two packed sequences contain the same number of predictive distributions, and their corresponding rows are used for training and evaluation.

\paragraph{Alignment Example.}
Consider the original sequence
\begin{equation*}
X=[x_1,x_2,x_3,x_4,x_5,x_6],
\end{equation*}
where $(x_2,x_3)$ and $(x_5,x_6)$ are replaced by surrogate embeddings $z_{23}$ and $z_{56}$. The masks and compressed sequence are
\begin{align*}
M^{T} &=[0,0,1,1,0,1],\\
\widetilde{X} &=[x_1,z_{23},x_4,z_{56}],\\
M^{\mathrm{attn}} &=[1,1,1,1],\\
M^{S} &=[0,1,1,1].
\end{align*}
The aligned prediction positions are therefore
\begin{equation*}
(x_3,z_{23}),\qquad
(x_4,x_4),\qquad
(x_6,z_{56}).
\end{equation*}
The notation above refers to the states at which the corresponding next-token distributions are produced, rather than requiring their hidden representations to be identical.

\paragraph{Distillation Objective.}
Let
\begin{equation*}
\mathcal{I}
=
\{(b,\ell)\mid 1\leq \ell\leq N_b\}
\end{equation*}
denote all aligned positions in a batch, and let
\begin{equation*}
N_{\mathrm{valid}}=\sum_b N_b.
\end{equation*}
We optimize the soft-target cross-entropy between the teacher and student distributions:
\begin{equation}
\mathcal{L}_{\mathrm{distill}}
=
-\frac{1}{N_{\mathrm{valid}}}
\sum_{(b,\ell)\in\mathcal{I}}
\sum_{v=1}^{|\mathcal{V}|}
\overline{P}^{T}_{b,\ell}(v)
\log\left(
\overline{P}^{S}_{b,\ell}(v)+\epsilon
\right),
\label{eq:masked_distillation_loss}
\end{equation}
where $\mathcal{V}$ is the vocabulary and $\epsilon$ is a small constant for numerical stability. The loss is normalized by the total number of aligned positions rather than by the padded sequence length.

The forward KL divergence satisfies
\begin{equation*}
D_{\mathrm{KL}}(P^{T}\|P^{S})
=
H(P^{T},P^{S})-H(P^{T}).
\end{equation*}
Because the teacher and backbone are frozen, $H(P^{T})$ is constant with respect to $\phi$. Minimizing the soft-target cross-entropy therefore gives the same gradient with respect to the merge-module parameters as minimizing the forward KL divergence.

\paragraph{Practical Note on Full-Vocabulary Distillation.}
\label{sec:full-vocab-distill}

Our implementation uses the complete teacher distribution over the vocabulary rather than a truncated Top-$k$ target. At each aligned position, the training objective is
\begin{equation*}
\mathcal{L}_{\mathrm{full}}
=
-\sum_{v\in\mathcal{V}}
P^{T}(v)\log P^{S}_{\phi}(v),
\end{equation*}
where $P^{T}$ is the predictive distribution of the original frozen model and $P^{S}_{\phi}$ is the distribution produced by the compressed path. The only trainable component in the compressed path is the merge module.

This choice preserves the teacher distribution without introducing a truncation threshold or modifying its probability mass. A Top-$k$ target retains only a subset $\mathcal{K}$ of the teacher vocabulary support and either removes or renormalizes the remaining probability mass. The resulting objective therefore matches the student to a modified target rather than to the original teacher distribution.

Importantly, our motivation is not that tokens outside the teacher's Top-$k$ set necessarily receive no gradient. If a sparse teacher target is combined with a full student softmax, an outside token with target probability zero still receives a gradient proportional to its student probability. The relevant difference is that the relative probabilities assigned by the teacher outside $\mathcal{K}$ are no longer preserved. Truncation therefore changes the desired distribution and removes part of the information available for matching the original model's predictive behavior.

This distinction is particularly relevant for UMIM because the compressed path can initially differ substantially from the original path. The merge module must learn an input embedding whose effect propagates through the entire frozen Transformer. Unlike conventional LM-to-LM distillation, the trainable component does not have its own language-model layers or intermediate supervision. Full-vocabulary matching provides the merge module with the complete output-level behavior of the frozen backbone at every aligned position.

During early development, we tested teacher Top-$k$ targets with $k\in\{128,256,512,1024\}$. None of these settings produced reliable convergence or useful teacher--student alignment in our preliminary runs, whereas full-vocabulary distillation trained successfully. These exploratory runs were used to select the final training configuration, but they were not designed or retained as a controlled quantitative ablation. We therefore report this observation as a practical implementation finding rather than as evidence that Top-$k$ distillation cannot work for UMIM.

Full-vocabulary distillation increases training-time memory and computation because teacher and student probabilities are retained over the complete vocabulary. This cost is limited to merge-module training and does not affect inference. More sophisticated sparse or adaptive distillation objectives may provide a better memory--quality trade-off, but we leave their systematic study to future work. In our experiments, full-vocabulary matching was the most reliable configuration among the variants we tested. Top-$k$ and Top-$p$ are used only as alignment metrics and are not training objectives.

\paragraph{Alignment Metrics.}
In addition to the distillation loss, we report four metrics over the same aligned positions. These metrics are used only to evaluate predictive agreement and are not additional training objectives.

\begin{itemize}
    \item \textbf{Top-1 Agreement.}
    Let
    \begin{equation*}
    y^{T}_{i}=\arg\max_v \overline{P}^{T}_{i}(v),
    \qquad
    y^{S}_{i}=\arg\max_v \overline{P}^{S}_{i}(v).
    \end{equation*}
    Top-1 agreement is
    \begin{equation*}
    \mathrm{Top1}
    =
    \frac{1}{N_{\mathrm{valid}}}
    \sum_{i=1}^{N_{\mathrm{valid}}}
    \mathbf{1}[y^{T}_{i}=y^{S}_{i}].
    \end{equation*}

    \item \textbf{Top-$k$ Overlap.}
    Let $\mathcal{K}^{T}_{i}$ and $\mathcal{K}^{S}_{i}$ be the sets of the $k$ most probable tokens under the teacher and student distributions. We use $k=3$ and $k=10$:
    \begin{equation*}
    \mathrm{Overlap}_{k}
    =
    \frac{1}{N_{\mathrm{valid}}}
    \sum_{i=1}^{N_{\mathrm{valid}}}
    \frac{
    |\mathcal{K}^{T}_{i}\cap\mathcal{K}^{S}_{i}|
    }{k}.
    \end{equation*}

    \item \textbf{Top-$p$ Overlap.}
    Let $\mathcal{P}^{T}_{i}(p)$ and $\mathcal{P}^{S}_{i}(p)$ be the smallest token sets whose cumulative probability reaches the threshold $p$. Their overlap is
    \begin{equation*}
    \mathrm{Overlap}_{p}
    =
    \frac{1}{N_{\mathrm{valid}}}
    \sum_{i=1}^{N_{\mathrm{valid}}}
    \frac{
    |\mathcal{P}^{T}_{i}(p)\cap\mathcal{P}^{S}_{i}(p)|
    }{
    \min\left(
    |\mathcal{P}^{T}_{i}(p)|,
    |\mathcal{P}^{S}_{i}(p)|
    \right)
    }.
    \end{equation*}
    The value of $p$ is specified by the corresponding experimental setting.

    \item \textbf{Mean Reciprocal Rank.}
    For each aligned position, we take the teacher's top-1 token $y^{T}_{i}$ and measure its rank $\operatorname{rank}_{S}(y^{T}_{i})$ under the student distribution:
    \begin{equation*}
    \mathrm{MRR}
    =
    \frac{1}{N_{\mathrm{valid}}}
    \sum_{i=1}^{N_{\mathrm{valid}}}
    \frac{1}{
    \operatorname{rank}_{S}(y^{T}_{i})
    }.
    \end{equation*}
\end{itemize}

\paragraph{Implementation Summary.}
Algorithm~\ref{alg:masked-cross-entropy} summarizes the alignment and loss computation. Padding is introduced only after teacher and student positions have been selected, and padded rows are excluded from both the loss and all alignment metrics.

\begin{algorithm}[H]
\caption{Aligned Predictive Distillation}
\label{alg:masked-cross-entropy}
\begin{algorithmic}[1]
\Require Teacher probabilities $P^{T}$, compressed probabilities $P^{S}$, keep mask $M^{\mathrm{keep}}$, compressed attention mask $M^{\mathrm{attn}}$, and first merge position $s_b$ for each sample
\Ensure Distillation loss $\mathcal{L}_{\mathrm{distill}}$

\State Initialize aligned teacher list $\mathcal{T}$ and student list $\mathcal{S}$

\State For each sample $b$:
\State \quad $M^{T}_{b}\gets M^{\mathrm{keep}}_{b}$
\State \quad Set $M^{T}_{b,j}\gets 0$ for all $j<s_b$
\State \quad $N_b\gets\sum_j M^{T}_{b,j}$
\State \quad $\widetilde{L}_b\gets\sum_j M^{\mathrm{attn}}_{b,j}$
\State \quad $M^{S}_{b}\gets M^{\mathrm{attn}}_{b}$
\State \quad Set the first $\widetilde{L}_b-N_b$ valid entries of $M^{S}_{b}$ to $0$
\State \quad $T_b\gets\operatorname{Pack}(P^{T}_{b},M^{T}_{b})$
\State \quad $S_b\gets\operatorname{Pack}(P^{S}_{b},M^{S}_{b})$
\State \quad Verify that $|T_b|=|S_b|=N_b$
\State \quad Append $T_b$ and $S_b$ to $\mathcal{T}$ and $\mathcal{S}$

\State Compute $\mathcal{L}_{\mathrm{distill}}$ from all aligned rows using Equation~\ref{eq:masked_distillation_loss}
\State \Return $\mathcal{L}_{\mathrm{distill}}$
\end{algorithmic}
\end{algorithm}

\section{Inference with {\namem}: Prompt and Decoding-Time Merging}
\label{Decoding Implementation}

{\namem} can be applied during prompt processing and autoregressive decoding. Prompt merging compresses the input before prefill, while decoding-time merging further compresses generated-token KV states. The two mechanisms can be used separately for ablation, while the full {\namem} pipeline applies both.

The merge decision depends only on realized token IDs and is independent of the token-selection rule. It can therefore be combined with greedy decoding, top-$k$ sampling, or top-$p$ sampling by changing the token-selection step. Our reference implementation uses greedy decoding. The implementation requires the backbone to accept input embeddings and expose a KV cache that can be cropped during generation.

\paragraph{Rule Matching.}
For prompt merging, we scan the tokenized input from left to right. At each position, candidate spans are checked in descending length order, i.e., 4-, 3-, and then 2-grams in our experiments. If a match is found, we select the longest matching span and advance to the first token after it. Otherwise, we advance by one token. This produces a set of non-overlapping merge spans.

The same longest-first policy is used during decoding, but only the unmerged raw-token suffix ending at the newly generated token needs to be checked. A surrogate created by an earlier merge is treated as one internal state and is not recursively included in another merge. This prevents overlapping or nested merges.

\paragraph{Prompt Merging.}
Given an input prompt with token IDs
\begin{equation*}
X=[x_1,\ldots,x_{L_p}],
\end{equation*}
we first identify a set of non-overlapping spans $\mathcal{S}_p$. For every span $(i,j)\in\mathcal{S}_p$, the original embeddings
\begin{equation*}
[e(x_i),\ldots,e(x_{j-1})]
\end{equation*}
are passed through the merge module to obtain one surrogate embedding. The original span is then replaced by this surrogate.

For efficiency, all selected spans of the same length can be processed together in one batched merge-module call. After replacing all selected spans, the compressed prompt length is
\begin{equation*}
\widetilde{L}_p
=
L_p-\sum_{(i,j)\in\mathcal{S}_p}(j-i-1).
\end{equation*}
The compressed embeddings use contiguous positions and are passed to the frozen backbone for prefill. Consequently, the initial KV cache contains $\widetilde{L}_p$ states rather than $L_p$ states.

If decoding-time merging is disabled, generation then proceeds normally without further cache modification. Prompt merging therefore directly reduces prefill sequence length and initial KV-cache length, while leaving the subsequent autoregressive procedure unchanged.

\begin{lstlisting}[
style=compactpython,
caption={Prompt-Only Merging},
label={lst:prompt_only_merging}
]
# x: input text
# R[n]: merge rules of length n
# M: trained merge module
# LM: frozen language model

token_ids = Tokenizer(x)
spans = []
i = 0

# Left-to-right, longest-first matching
while i < len(token_ids):
    matched = False

    for n in [4, 3, 2]:
        span = tuple(token_ids[i:i+n])

        if len(span) == n and span in R[n]:
            spans.append((i, i+n))
            i = i + n
            matched = True
            break

    if not matched:
        i = i + 1

# Spans of the same length can be pooled in batch
surrogates = {}
for n in [2, 3, 4]:
    spans_n = [
        (start, end) for start, end in spans
        if end - start == n
    ]

    if len(spans_n) > 0:
        span_embeddings = stack([
            model_embedding(token_ids[start:end])
            for start, end in spans_n
        ])
        pooled = M(span_embeddings)

        for (start, end), embedding in zip(
            spans_n, pooled
        ):
            surrogates[start] = (end, embedding)

# Construct the packed compressed prompt
compressed_embeddings = []
i = 0

while i < len(token_ids):
    if i in surrogates:
        end, embedding = surrogates[i]
        compressed_embeddings.append(embedding)
        i = end
    else:
        compressed_embeddings.append(
            model_embedding(token_ids[i])
        )
        i = i + 1

E = stack(compressed_embeddings)

generated_ids = autoregressive_decode(
    LM,
    prompt_embeddings=E,
    merge_during_decoding=False
)

return Tokenizer.decode(generated_ids)
\end{lstlisting}

\paragraph{Decoding-Time Merging and KV-Cache Update.}
Decoding-time merging is performed after each new token ID is selected. Suppose the newly generated token completes an unmerged suffix
\begin{equation*}
g=(x_{t-n+1},\ldots,x_t)
\end{equation*}
that appears in the merge rules. At this point, the newest token $x_t$ has been sampled from the current logits but has not yet been forwarded through the model. Therefore, only the preceding $n-1$ tokens in the matched suffix already have KV entries.

We crop these $n-1$ KV entries, compute the surrogate embedding
\begin{equation*}
m_g=M_{\phi}
\left(
e(x_{t-n+1}),\ldots,e(x_t)
\right),
\end{equation*}
and forward $m_g$ through the frozen backbone using the cropped cache. This forward pass produces one new key and value state for the surrogate at every Transformer layer.

Thus, {\namem} does not directly construct or average the original KV states. It removes the available raw-token KV entries and lets the frozen backbone compute a new KV state from the surrogate embedding. If the cache length before processing $x_t$ is $L_C$, the cache length after merging an $n$-gram becomes
\begin{equation*}
L_C' = L_C-(n-1)+1.
\end{equation*}
By comparison, processing $x_t$ without merging would produce a cache of length $L_C+1$. The matched span therefore occupies one KV position instead of $n$ positions.

If no rule matches the suffix, the newly generated token is forwarded normally and one KV entry is appended. In either case, one token is selected during every autoregressive step. {\namem} does not skip token-generation steps; it reduces the effective sequence and cache lengths used by later steps.

\begin{lstlisting}[
style=compactpython,
caption={Prompt and Decoding-Time Merging},
label={lst:dynamic_online_merging}
]
# x: input text
# R[n]: merge rules of length n
# M: trained merge module
# LM: frozen language model

token_ids = Tokenizer(x)

# Apply prompt merging and prefill the model
prompt_spans = find_nonoverlapping_spans(
    token_ids,
    R,
    longest_first=True
)
E = build_compressed_embeddings(
    token_ids,
    prompt_spans,
    M
)
logits, cache = LM.forward(
    inputs_embeds=E,
    use_cache=True
)

# generated_ids preserves the actual output tokens
generated_ids = []

# generated_state is used only for suffix matching.
# A merged marker prevents overlapping merges.
generated_state = []

while not finished:
    y = select_next_token(logits)
    generated_ids.append(y)

    if y is an end-of-sequence token:
        break

    generated_state.append(y)

    span = longest_matching_raw_suffix(
        generated_state,
        R,
        lengths=[4, 3, 2]
    )

    if span is None:
        step_embedding = model_embedding(y)

    else:
        n = len(span)

        # The newest token has not entered the cache.
        # Only the previous n-1 entries are removed.
        cache = crop_cache(
            cache,
            cache_length(cache) - (n - 1)
        )

        span_embeddings = model_embedding(span)
        step_embedding = M(span_embeddings)

        # Replace the raw suffix with an internal marker.
        generated_state = (
            generated_state[:-n]
            + [MERGED(span)]
        )

    logits, cache = LM.forward(
        inputs_embeds=step_embedding,
        past_key_values=cache,
        use_cache=True
    )

return Tokenizer.decode(generated_ids)
\end{lstlisting}

\paragraph{Output Preservation.}
Surrogate embeddings and merged markers are used only for the model's internal computation. We always preserve the complete sequence of generated token IDs separately. The final output is decoded from these original generated IDs, not from the compressed internal representation. Therefore, decoding-time merging reduces the internal sequence and KV-cache lengths without removing tokens from the returned text.

\paragraph{Computational Scope.}
Prompt merging provides an immediate reduction in prefill length, and this shorter prompt cache is retained throughout generation. Decoding-time merging further reduces generated-token KV-cache growth whenever a generated suffix matches the merge rules. However, it cannot skip autoregressive generation steps and introduces additional work for suffix lookup, merge-module evaluation, and cache cropping. This explains why the realized decoding speedup can be smaller than the prefill speedup even when the cache length is reduced.

The reported implementation uses standard Python and PyTorch operations without custom CUDA or Triton kernels. Since the surrogate embedding for a fixed token span is context-independent, frequently used surrogates could also be precomputed and retrieved through a lookup table. Suffix matching, surrogate lookup, and KV-cache updates could further be fused into optimized kernels. These engineering optimizations are not included in the reported implementation and are left for future work.

\section{Task Adaptation with SFT and RL}
\label{sec:task_adaptation_details}

This section describes how UMIM is adapted to downstream tasks while keeping the backbone LLM fully frozen. Task adaptation is distinct from the task-agnostic evaluation in the main experiments. In the task-agnostic setting, we directly use the merge rules mined from WikiText-103 and the corresponding WikiText-trained base merge module, without using downstream training data. Task adaptation instead reconstructs merge rules from the target task's training split and optionally updates the merge module through SFT and RL.

The complete adaptation pipeline contains three stages:
\begin{enumerate}
    \item Construct task-specific merge rules from the downstream training split.
    \item Initialize from the WikiText-trained base merge module and perform distillation-based SFT.
    \item Initialize from the SFT checkpoint and further optimize the merge module through RL.
\end{enumerate}
The backbone LLM remains frozen throughout all stages.

\subsection{Task-Specific Merge-Rule Construction}

Given the target-task training set $\mathcal{D}_{\mathrm{task}}$, we tokenize its examples using the backbone LLM's tokenizer and count adjacent token spans. In the reported task-adaptation experiments, we use task-specific bigrams and vary the frequency threshold to control the rule-set size and resulting merge ratio.

For a multiple-choice task, each training sequence is constructed by concatenating the task input with its correct answer or completion. We count the bigrams in these sequences and retain candidates whose frequency is no smaller than a threshold $\tau$:
\begin{equation*}
\mathcal{R}_{\mathrm{task}}^{(\tau)}
=
\left\{
g:
\operatorname{count}_{\mathcal{D}_{\mathrm{task}}}(g)
\geq \tau
\right\}.
\end{equation*}

When applying the rules, we scan each token sequence from left to right using non-overlapping matching. If the bigram beginning at the current position belongs to $\mathcal{R}_{\mathrm{task}}^{(\tau)}$, it is selected and the scan advances by two tokens. Otherwise, the scan advances by one token.

A lower threshold retains more task-specific bigrams and therefore increases the achievable merge ratio, but it may also introduce spans with less training support. A higher threshold gives a more conservative rule set. The different thresholds produce the accuracy--compression curves reported in Figure~3 and Appendix Figure~5.

If no additional training is desired, the task-specific rules can be used directly with the fixed WikiText-trained base merge module. This corresponds to the \texttt{WT + Rules} setting.

\subsection{Distillation-Based SFT}

To further adapt the merge module to the task-specific rules, we initialize it from the WikiText-trained base merge module $M_{\phi_0}$ and perform SFT only on the merge module. This SFT stage is not conventional label-only fine-tuning and does not update the backbone LLM. Instead, it continues to use predictive distillation.

For a task-training sequence $X$, the frozen backbone first processes the original uncompressed sequence and produces the teacher distribution:
\begin{equation*}
P^{T}
=
\operatorname{softmax}
\left(
F_{\theta}(X)
\right).
\end{equation*}

The task-specific rules $\mathcal{R}_{\mathrm{task}}$ determine the spans to be merged. Replacing these spans with surrogate embeddings produces the compressed sequence
\begin{equation*}
\widetilde{X}
=
C_{\phi}
\left(
X;\mathcal{R}_{\mathrm{task}}
\right).
\end{equation*}
The same frozen backbone then processes the compressed sequence and produces the student distribution:
\begin{equation*}
P^{S}_{\phi}
=
\operatorname{softmax}
\left(
F_{\theta}(\widetilde{X})
\right).
\end{equation*}

The original and compressed distributions are aligned using the mask-and-pack procedure described in Section~\ref{Loss and Alignment Metric}. The merge module is optimized using the aligned soft-target cross-entropy:
\begin{equation*}
\phi_{\mathrm{SFT}}
=
\arg\min_{\phi}
\mathcal{L}_{\mathrm{distill}}
\left(
P^{T},P^{S}_{\phi}
\right).
\end{equation*}

Only $\phi$ receives gradients, while the backbone parameters $\theta$ remain frozen. For example, in HellaSwag, we construct each SFT sequence by concatenating the context with the correct ending. The merge module therefore learns to preserve the original model's predictive behavior under the target-task distribution and task-specific rules.

This stage corresponds to \texttt{WT + FT}. It combines the general span-to-surrogate initialization learned from WikiText-103 with domain-specific distillation signals from the downstream task.

\subsection{RL Adaptation}

Distillation-based SFT aims to preserve the original model's overall predictive behavior, but it does not directly optimize the downstream decision objective. We therefore initialize a policy merge module from the SFT checkpoint $M_{\phi_{\mathrm{SFT}}}$ and further optimize it through RL.

For a multiple-choice example, let $x$ denote the task context, $y^{+}$ the correct ending, and $y^{-}$ an incorrect ending. We construct
\begin{equation*}
X^{+}=[x;y^{+}],
\qquad
X^{-}=[x;y^{-}].
\end{equation*}

During this stage, the merge rules are applied to the task context shared by the correct and incorrect sequences. Both candidates therefore use the same compressed context, and the optimization signal is determined by their different endings.

We compute the mean token log-probability of each candidate ending:
\begin{equation*}
s_{\phi}(x,y)
=
\frac{1}{|y|}
\sum_{t=1}^{|y|}
\log
p_{\theta,\phi}
\left(
y_t
\mid
C_{\phi}(x;\mathcal{R}_{\mathrm{task}}),
y_{<t}
\right).
\end{equation*}

The policy merge module and a frozen reference merge module are initialized from the same checkpoint. In the final \texttt{Task-Adapted} setting, both are initialized from $M_{\phi_{\mathrm{SFT}}}$. The reference merge module remains fixed, while only the policy merge module receives gradients.

We use the following reference-regularized pairwise RL objective:
\begin{equation}
\begin{aligned}
\mathcal{L}_{\mathrm{RL}}
=
-\log \sigma
\Big(
\beta
\big[
&
\left(
s_{\phi}(x,y^{+})
-
s_{\mathrm{ref}}(x,y^{+})
\right)
\\
-&
\left(
s_{\phi}(x,y^{-})
-
s_{\mathrm{ref}}(x,y^{-})
\right)
\big]
\Big),
\end{aligned}
\label{eq:task_rl_objective}
\end{equation}
where $\beta$ controls the strength of the update relative to the frozen reference. When a question contains multiple incorrect candidates, we first average the pairwise losses within that question and then average across questions in the batch.

This stage does not retrain the language model. It only changes how the merge module constructs surrogate embeddings, encouraging the compressed frozen LLM to assign higher scores to correct answers. Distillation-based SFT first preserves general predictive behavior under the target-task distribution, while RL subsequently optimizes the task-level decision objective. Their combination forms the final \texttt{Task-Adapted} setting.

\subsection{Adaptation and Ablation Settings}

Table~\ref{tab:task_adaptation_settings} summarizes the task-adaptation and ablation settings used in Figure~3 and Figure~4.

\begin{table*}[t]
\centering
\caption{Task-adaptation and ablation settings. The backbone LLM remains frozen in every setting.}
\label{tab:task_adaptation_settings}
\small
\begin{tabular}{l l l c c c}
\toprule
\textbf{Setting}
& \textbf{Initialization}
& \textbf{Merge Rules}
& \textbf{Distillation SFT}
& \textbf{RL}
& \textbf{Backbone} \\
\midrule
\texttt{WT + Rules}
& WikiText base
& Task-specific
& No
& No
& Frozen \\
\texttt{WT + FT}
& WikiText base
& Task-specific
& Yes
& No
& Frozen \\
\texttt{Task-Adapted}
& WikiText base
& Task-specific
& Yes
& Yes
& Frozen \\
\texttt{Task-trained}
& Random
& Task-specific
& From scratch
& No
& Frozen \\
\texttt{WT + DPO}
& WikiText base
& Task-specific
& No
& Yes
& Frozen \\
\bottomrule
\end{tabular}
\end{table*}

In \texttt{Task-trained}, a new merge module is randomly initialized and trained directly on the target-task corpus using the same distillation objective. This setting measures the benefit of the WikiText-trained initialization. In \texttt{WT + DPO}, we skip distillation-based SFT and apply the RL objective directly to the WikiText-trained base merge module. These two settings are ablations rather than the final task-adaptation pipeline.

\subsection{Task-Adaptation Experimental Configuration}
\label{app:task_adaptation_configuration}

This subsection provides the experimental configurations used for the task-adaptation results in Section~\ref{sec:Task_Specific}. All experiments use Llama-3.1-8B. Except for the \texttt{Task-trained} ablation, the merge module is initialized from the base merge module trained on WikiText-103. Throughout adaptation, the tokenizer, input embedding matrix, and backbone LLM remain frozen; only the merge module parameters are updated.

\paragraph{Task-specific rule construction.}

For HellaSwag, task-specific merge rules are constructed using only the training split. Each training sequence is formed by concatenating the context with its correct ending and is truncated or padded to 128 tokens. We count adjacent token pairs and retain bigrams whose frequencies exceed thresholds in
$\{2000, 200, 50, 5, 1\}$. These thresholds produce the operating points reported in Figure~\ref{fig:hellaswag_progress}, with realized merge rates ranging from 12.10\% to 43.09\%. Candidate bigrams are applied from left to right using non-overlapping matching, so each token belongs to at most one merged span.

For ARC-Easy and ARC-Challenge, rules are constructed from the question tokens in each task's training split. We use bigram rules with frequency threshold 5 and a maximum sequence length of 128 tokens. The training sequence contains the question and its correct answer, while merging is applied to the question portion. Validation examples are used for model selection but not for rule construction or parameter updates.

\paragraph{SFT configuration.}

The SFT stage retains the predictive-distillation objective used to train the base merge module. For each training sequence, the frozen uncompressed LLM produces the teacher next-token distributions, while the compressed model produces the student distributions after applying the task-specific merge rules. After aligning the corresponding predictive positions, we minimize cross-entropy from the teacher distributions to the student distributions. Thus, although the adaptation data come from the downstream task, the backbone LLM is not trained with task labels; gradients are applied only to the merge module.

The \texttt{WT + FT} setting initializes the merge module from the WikiText-trained checkpoint. The \texttt{Task-trained} ablation uses the same task data and distillation objective but trains a newly initialized merge module instead of starting from the WikiText checkpoint. SFT checkpoints are selected according to validation distillation loss, with early stopping.

\paragraph{RL configuration.}

The RL stage is implemented using DPO. For each multiple-choice example, the correct option is used as the preferred continuation, and each incorrect option forms a rejected continuation. The preferred and rejected sequences share the same compressed context or question, ensuring that the optimization targets how the adapted merge module affects downstream predictions.

Both the policy merge module and the frozen reference merge module are initialized from the same SFT checkpoint. We compute the mean token log-probability of each answer continuation and optimize the DPO objective to increase the policy's relative preference for the correct continuation. The backbone LLM and reference merge module remain frozen, and only the policy merge module is updated. The \texttt{WT + DPO} ablation uses the same RL procedure but initializes directly from the WikiText-trained merge module, without the preceding SFT stage. RL checkpoints are selected according to validation accuracy.

\paragraph{Optimization settings.}

All adaptation runs use eight 48\,GB GPUs, DeepSpeed ZeRO stage 2, and bf16 precision. We use AdamW with gradient accumulation set to 1, gradient clipping at 1.0, and a warmup-decay schedule with 10\% of the optimization steps used for warmup. Weight decay is $10^{-3}$ for SFT and $10^{-4}$ for RL. The remaining task-specific configurations are summarized in Table~\ref{tab:task_adaptation_configurations}.

\begin{table*}[t]
\centering
\caption{Task-adaptation training configurations. Batch denotes the micro-batch size per GPU. The RL stage is implemented using DPO.}
\label{tab:task_adaptation_configurations}
\small
\setlength{\tabcolsep}{5pt}
\resizebox{\textwidth}{!}{
\begin{tabular}{llcccccc}
\toprule
\textbf{Task}
& \textbf{Stage}
& \textbf{Rule Threshold}
& \textbf{Batch/GPU}
& \textbf{Learning Rate}
& \textbf{Epochs}
& \textbf{Patience}
& $\boldsymbol{\beta}$ \\
\midrule
HellaSwag
& SFT
& $\{2000,200,50,5,1\}$
& 32
& $1\times10^{-4}$
& 5
& 3
& -- \\
HellaSwag
& RL
& $\{2000,200,50,5,1\}$
& 32
& $5\times10^{-5}$
& 2
& 2
& 0.1 \\
\midrule
ARC-Easy
& SFT
& 5
& 2
& $5\times10^{-4}$
& 12
& 3
& -- \\
ARC-Easy
& RL
& 5
& 4
& $5\times10^{-5}$
& 10
& 3
& 0.1 \\
\midrule
ARC-Challenge
& SFT
& 5
& 4
& $5\times10^{-4}$
& 10
& 5
& -- \\
ARC-Challenge
& RL
& 5
& 2
& $5\times10^{-4}$
& 10
& 3
& 0.1 \\
\bottomrule
\end{tabular}
}
\end{table*}

\subsection{Implementation Procedures}

SFT and RL are implemented as two separate training stages rather than a joint optimization procedure. The SFT stage first produces a task-specific merge-module checkpoint. The RL stage then loads this checkpoint and uses it to initialize both the trainable policy merge module and the frozen reference merge module.

\paragraph{SFT implementation.}
Listing~\ref{lst:task_sft} summarizes the distillation-based SFT stage.

\begin{lstlisting}[
style=compactpython,
caption={Distillation-Based SFT for Task Adaptation},
label={lst:task_sft}
]
# D_task: downstream training split
# F_theta: frozen backbone LLM
# M_base: WikiText-trained base merge module

# Construct task-specific merge rules
R_task = mine_frequent_rules(D_task)

# Initialize from the WikiText checkpoint
M_sft = load_wikitext_merge_module()
freeze(F_theta)

for original_sequence in task_training_sequences:
    # Teacher path: original sequence
    with no_gradient():
        teacher_probs = F_theta(
            original_sequence
        )

    # Student path: compressed sequence
    compressed_sequence = merge(
        original_sequence,
        R_task,
        M_sft
    )
    student_probs = F_theta(
        compressed_sequence
    )

    # Align teacher and student positions
    teacher_aligned, student_aligned = align(
        teacher_probs,
        student_probs
    )

    loss = soft_target_cross_entropy(
        teacher_aligned,
        student_aligned
    )

    # The backbone remains frozen
    update_only(M_sft, loss)

save_merge_module(
    M_sft,
    "task_sft_checkpoint"
)
\end{lstlisting}

The SFT checkpoint is saved independently after distillation training. Only the merge-module parameters are stored and passed to the next stage; the frozen backbone is not modified.

\paragraph{RL implementation.}
Listing~\ref{lst:task_rl} summarizes the subsequent RL stage. It is launched as a separate training run after the SFT checkpoint has been produced.

\begin{lstlisting}[
style=compactpython,
caption={RL Adaptation from the SFT Checkpoint},
label={lst:task_rl}
]
# F_theta: frozen backbone LLM
# R_task: task-specific merge rules
# D_pairs: correct/incorrect task pairs

M_sft = load_merge_module(
    "task_sft_checkpoint"
)

# Both modules start from the same SFT weights
M_policy = copy(M_sft)
M_reference = frozen_copy(M_sft)

freeze(F_theta)
freeze(M_reference)

for question in D_pairs:
    context = question.context
    correct_answer = question.correct_answer
    wrong_answers = question.wrong_answers

    pair_losses = []

    for wrong_answer in wrong_answers:
        # The same compressed context is used for
        # the correct and incorrect candidates.
        policy_positive = ending_score(
            F_theta,
            M_policy,
            context,
            correct_answer,
            R_task
        )
        policy_negative = ending_score(
            F_theta,
            M_policy,
            context,
            wrong_answer,
            R_task
        )

        with no_gradient():
            reference_positive = ending_score(
                F_theta,
                M_reference,
                context,
                correct_answer,
                R_task
            )
            reference_negative = ending_score(
                F_theta,
                M_reference,
                context,
                wrong_answer,
                R_task
            )

        pair_loss = rl_objective(
            policy_positive,
            policy_negative,
            reference_positive,
            reference_negative
        )
        pair_losses.append(pair_loss)

    # Average over candidates from the same question
    loss = mean(pair_losses)

    # Only the policy merge module is updated
    update_only(M_policy, loss)

save_merge_module(
    M_policy,
    "task_adapted_checkpoint"
)
\end{lstlisting}

The two stages are therefore connected only through the merge-module checkpoint. Distillation-based SFT first adapts the WikiText-trained module to the target-task distribution and merge rules. RL then starts from this adapted checkpoint and further optimizes task-level decisions. At no point are the SFT and RL objectives optimized jointly, and the backbone LLM remains frozen throughout both stages.

For the \texttt{WT + DPO} ablation, the RL procedure is unchanged, but both the policy and reference merge modules are initialized directly from the WikiText-trained base checkpoint rather than the SFT checkpoint. For \texttt{Task-trained}, the SFT procedure is initialized randomly instead of using the WikiText-trained merge module.

\section{Base Merge-Module Training Setup}
\label{sec:Implementation Details}
\label{sec:base_training_setup}

This section describes the corpora, optimization settings, hardware, and model-specific configurations used to train the base merge modules. The task-adaptation stages are described separately in Section~\ref{sec:task_adaptation_details}.

\subsection{Training Corpora}
\label{Datasets}
\label{sec:training_corpora}

\paragraph{WikiText-based training.}
For Llama-3.1-8B, Llama-3.2-1B, and GPT-2-XL, we train the base merge modules using WikiText-103 (\texttt{wikitext-103-raw-v1}). Each corpus split is tokenized using the corresponding backbone tokenizer. We concatenate the tokenized documents within each split, discard the incomplete remainder, and partition the resulting token stream into non-overlapping sequences of 512 tokens.

We use the official WikiText-103 training, validation, and test splits. The merge rules are mined only from the training split and are then applied to all three splits. The resulting token IDs, attention masks, and merge positions are precomputed and stored for distributed training.

Although Llama-3.1-8B and Llama-3.2-1B use different backbone sizes, they have compatible tokenization in our experiments and therefore produce the same WikiText-derived rule counts. GPT-2-XL uses its own tokenizer and a separately constructed rule set.

\paragraph{Math-reasoning training corpus.}
For DeepScaleR-1.5B-Preview, we construct a separate math-reasoning corpus to demonstrate that a merge module can also be trained for long-form reasoning domains. The source questions are taken from \texttt{KbsdJames/Omni-MATH} and \texttt{RUC-AIBOX/STILL-3-Preview-RL-Data}. We retain questions of at most 200 tokens and use \texttt{agentica-org/DeepScaleR-1.5B-Preview} to sample three reasoning trajectories for each question. Generation uses nucleus sampling with $p=0.7$, temperature $1.0$, and a maximum total sequence length of 2048 tokens.

The generated texts, including the source questions and sampled reasoning, are tokenized and concatenated into a continuous token stream. This stream is then divided into non-overlapping sequences of 1024 tokens. Thus, 1024 is the fixed training-sequence length after corpus construction; individual generated trajectories are not required to contain exactly 1024 tokens.

We divide the resulting sequences into 90\% training, 5\% validation, and 5\% test splits. The DeepScaleR merge rules are mined only from the training portion of this generated reasoning corpus. This setup is separate from the WikiText-based training used for the Llama and GPT-2 backbones.

\subsection{Optimization and Distributed Training}
\label{Parameter Settings and Training Implementation Choices}
\label{sec:base_optimization}

All base merge modules are trained on eight NVIDIA GPUs. Depending on machine availability, the experiments use NVIDIA RTX A6000, RTX 6000 Ada, or L40S GPUs, each with 48\,GB of memory. Llama-3.1-8B and DeepScaleR-1.5B-Preview use DeepSpeed ZeRO stage 2, while Llama-3.2-1B and GPT-2-XL use PyTorch DistributedDataParallel. All merge-module training is performed in \texttt{bfloat16} precision.

We use AdamW with weight decay $1\times10^{-3}$. The learning-rate schedule contains a linear warmup over the first 10\% of training steps, followed by linear decay. Each run is configured for at most 15 epochs. We select the best checkpoint according to validation distillation loss and use early stopping with patience 3. Training can be resumed from a saved checkpoint when interrupted by the compute-time limit.

The original and compressed prediction paths use the same frozen backbone checkpoint. The original sequence produces the teacher distribution, while the compressed sequence produces the student distribution through the merge module and the frozen backbone. Gradients are propagated only to the merge-module parameters; all backbone parameters remain frozen.

The preprocessed token IDs and attention masks are stored as \texttt{.pt} files, while variable-length merge positions are stored as \texttt{.pkl} files. They are loaded through a custom dataset and distributed using \texttt{DistributedSampler}.

During training, we record the distillation cross-entropy, Top-1 agreement, Top-3 overlap, Top-10 overlap, Top-$p$ overlap with $p=0.9$, MRR, and the average number of merged spans. These quantities are evaluation and monitoring metrics; only the distillation cross-entropy is optimized. Training and validation statistics are logged with Weights \& Biases.

\subsection{Model-Specific Configurations}
\label{sec:model_specific_training}

Table~\ref{tab:base_training_configurations} reports the model-specific training configurations. Batch size is reported per GPU. All merge modules use four pooling heads.

\begin{table*}[t]
\centering
\caption{Base merge-module training configurations. All runs use eight 48\,GB GPUs with the backbone frozen.}
\label{tab:base_training_configurations}
\small
\setlength{\tabcolsep}{4pt}
\resizebox{\textwidth}{!}{%
\begin{tabular}{@{}l l c c c c c@{}}
\toprule
\textbf{Backbone}
& \textbf{Corpus}
& \textbf{Method}
& \textbf{Length}
& \textbf{Batch/GPU}
& \textbf{LR}
& $\boldsymbol{d}$ \\
\midrule
Llama-3.1-8B
& WT-103
& ZeRO-2
& 512
& 3
& $8\times10^{-4}$
& 4096 \\
Llama-3.2-1B
& WT-103
& DDP
& 512
& 4
& $8\times10^{-4}$
& 2048 \\
GPT-2-XL
& WT-103
& DDP
& 512
& 4
& $7\times10^{-4}$
& 1600 \\
DeepScaleR-1.5B-Preview
& Math
& ZeRO-2
& 1024
& 3
& $8\times10^{-4}$
& 1536 \\
\bottomrule
\end{tabular}%
}
\end{table*}

Here, $d$ denotes the backbone input-embedding dimension and therefore also the input and output dimension of the corresponding merge module. A separate base merge module is trained for each backbone because the module operates directly in that backbone's embedding space.

\section{Additional Experiments and Evaluation}
\label{sec:Additional Experiments and Evaluation}

This section provides additional evaluation protocols and results for perplexity, open-ended generation, downstream tasks, long-form mathematical reasoning, and specialized-domain adaptation.

\subsection{Perplexity Evaluation}
\label{sec:additional_ppl}

We evaluate language-modeling performance on WikiText-103, BookCorpus, and OpenWebText. The WikiText-derived merge rules and base merge module are applied directly to these datasets without task-specific training. Thus, BookCorpus and OpenWebText evaluate task-agnostic transfer beyond the merge module's WikiText training corpus.

A direct perplexity calculation on only the compressed sequence would omit the original tokens that are absorbed into surrogate embeddings. We therefore use a two-stage accounting procedure that assigns likelihood to all $N-1$ next-token targets in an original sequence of length $N$.

In the first stage, the sequence is partitioned into unmerged tokens and merged spans. A single compressed forward pass evaluates transitions between consecutive compressed elements. When the next element is a merged span, its first constituent token is used as the next-token target. In the second stage, auxiliary prefix forwards evaluate the remaining constituent tokens inside every merged span. The two losses are summed and normalized by the original number of next-token targets. This procedure makes UMIM perplexity comparable to the original token-level perplexity while retaining the compressed sequence in the main forward pass.

\begin{lstlisting}[
style=compactpython,
caption={Two-Stage Token-Level PPL Accounting},
label={lst:two_stage_ppl}
]
# x: original token IDs of length N
# R: merge rules
# M: trained merge module
# LM: frozen language model

elements = merge_longest_first(x, R)

def embed(element):
    if isinstance(element, tuple):
        token_embs = stack([
            model_embedding(token_id)
            for token_id in element
        ])
        return M(token_embs)
    return model_embedding(element)

def first_token(element):
    if isinstance(element, tuple):
        return element[0]
    return element

# Stage 1: transitions between compressed elements
main_inputs = elements[:-1]
main_targets = [
    first_token(element)
    for element in elements[1:]
]

main_embeddings = stack([
    embed(element)
    for element in main_inputs
])
main_logits = LM.forward(main_embeddings)

total_nll = 0
target_count = 0

for position, target in enumerate(main_targets):
    total_nll += -log_prob(
        main_logits[position],
        target
    )
    target_count += 1

# Stage 2: tokens skipped inside merged spans
for index, element in enumerate(elements):
    if not isinstance(element, tuple):
        continue

    prefix = elements[:index]

    for offset in range(1, len(element)):
        # Use the compressed preceding context, followed
        # by the observed raw prefix of the current span.
        context = (
            prefix
            + list(element[:offset])
        )
        context_embeddings = stack([
            embed(item)
            for item in context
        ])
        logits = LM.forward(context_embeddings)

        target = element[offset]
        total_nll += -log_prob(
            logits[-1],
            target
        )
        target_count += 1

assert target_count == len(x) - 1
PPL = exp(total_nll / target_count)
\end{lstlisting}

This two-stage procedure is used only for token-level likelihood accounting. It is different from decoding-time KV-cache rollback, which is described in Section~\ref{Decoding Implementation}.

\subsection{Text Generation}
\label{Text Generation}

\subsubsection{Generation Setup}

We conduct the open-ended generation experiments using Llama-3.1-8B. UMIM applies prompt merging before prefill and continues to check generated raw-token suffixes for decoding-time merging. Surrogate states are used only internally; the returned text always preserves the complete generated token-ID sequence.

For qualitative greedy-decoding results, we randomly select five prompts from the WikiText-103 test split. Depending on the length needed to present a complete example, we generate either 20 or 30 new tokens. We additionally use the prompt ``Tell me something about Lakers.'' as an illustrative Top-$p$ example. For this prompt, we draw three independent samples with $p=0.9$ and temperature $0.8$.

For quantitative evaluation, examples are constructed from the WikiText-103 test split. The first 100 tokens of each evaluated sequence are used as the prompt, and tokens 101--200 are used as the reference. Each model generates 100 new tokens conditioned on the prompt.

We evaluate three decoding strategies:
\begin{itemize}
    \item greedy decoding;
    \item Top-$p$ sampling with $p=0.9$ and temperature $0.8$;
    \item random sampling from the full next-token distribution without nucleus truncation.
\end{itemize}
For Top-$p$ and random sampling, we perform five independent generation runs for each input and report the average results.

Generation quality is evaluated using repetition and diversity metrics (Rep-2/3/4 and Dist-2/3/4), as well as ROUGE-2, ROUGE-L, ROUGE-Lsum, BERTScore, chrF, and Mauve.

\subsubsection{Qualitative Analysis}
\label{sematic}
\label{sec:qualitative_generation}

Tables~\ref{tab:qualitative-generation_greedy} and~\ref{tab:qualitative-generation_top} provide qualitative examples under greedy and Top-$p$ decoding. The brackets indicate token spans merged internally by UMIM. These brackets are visual annotations only and are not part of the returned text.

The examples show that UMIM can produce fluent continuations while repeatedly replacing multi-token spans with surrogate states. They should be interpreted as qualitative illustrations rather than evidence of exact semantic equivalence in every context.

\begin{figure}[t]
  \centering
  \includegraphics[width=\linewidth]{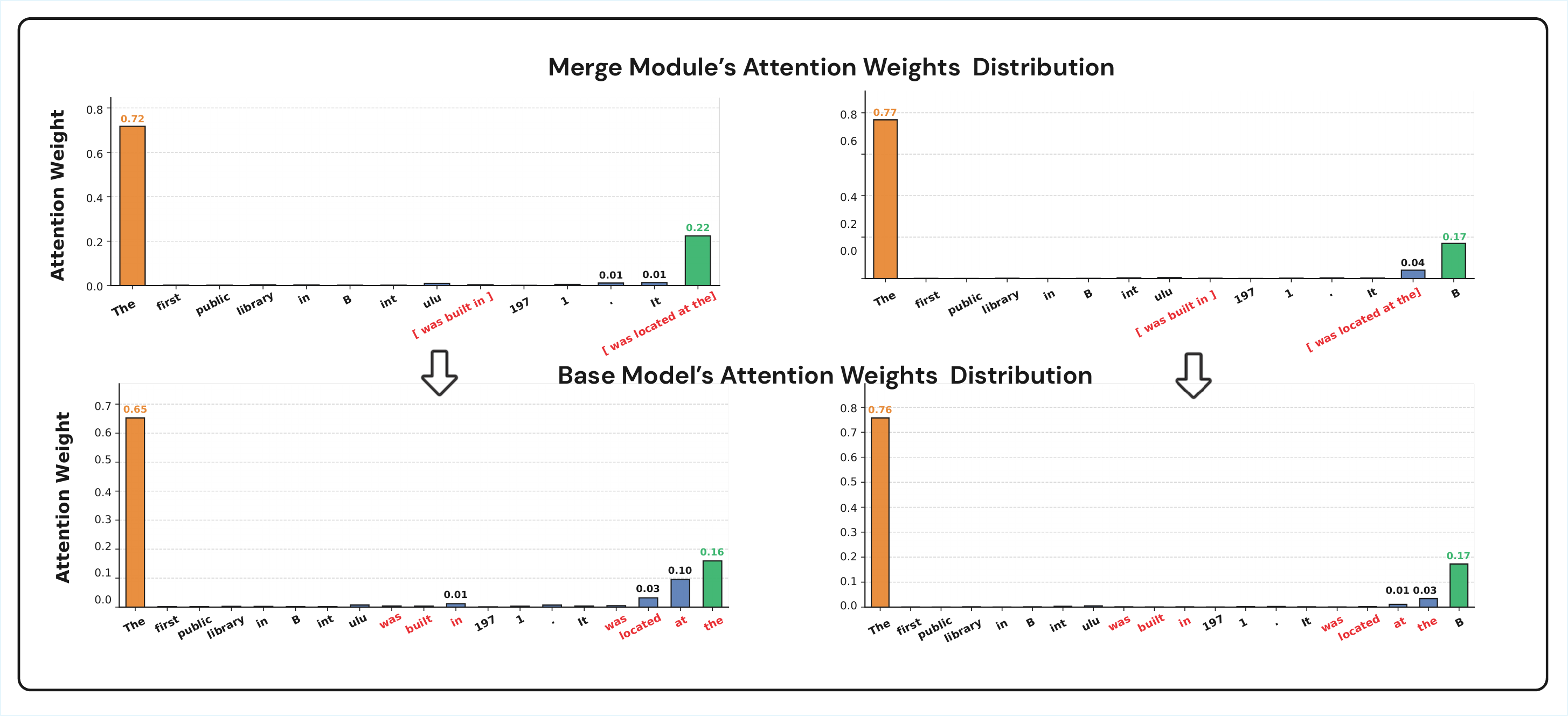}
  \caption{Attention distributions before and after applying UMIM. The visualization shows that surrogate states remain involved in subsequent attention, but does not imply exact attention or semantic equivalence.}
  \label{fig:attention}
\end{figure}

Figure~\ref{fig:attention} compares attention distributions before and after merging. The visualization shows that surrogate states continue to receive attention from later tokens rather than being ignored. However, attention weight alone does not prove complete semantic preservation or layer-wise equivalence.

We also observe that merged embeddings have near-zero cosine similarity with the static embeddings of their individual constituent tokens. This indicates that the merge module learns representations distinct from simply copying or averaging the constituent embeddings. It does not, by itself, establish that the surrogate is context-dependent or perfectly preserves the original span.

\newcolumntype{L}{>{\raggedright\arraybackslash}X}

\begin{table*}[t]
  \centering
  \caption{Qualitative generation results with greedy decoding. Brackets mark spans merged internally by UMIM.}
  \label{tab:qualitative-generation_greedy}
  \scriptsize
  \setlength{\tabcolsep}{5pt}
  \renewcommand{\arraystretch}{1.25}
  \begin{tabularx}{\textwidth}{@{}L L L@{}}
    \toprule
    \textbf{Prompt} & \textbf{Base Model} & \textbf{UMIM} \\
    \midrule
    Traditional Chinese literary criticism emphasized \textbf{[the life of the]} author when interpreting a work
    & The author's life was seen as the key to understanding the work.
    & The author's life \textbf{[was seen as]} the key \textbf{[to understanding the work]}. \\
    \midrule
    In \textbf{[2025]}, \textbf{[he moved to the]} institute \textbf{[in an attempt]} to study.
    & He was a very good student, and he was very good at his job.
    & He \textbf{[was a student of]} \textbf{[the institute and]} \textbf{[was a member of]} the institute's football team. \\
    \midrule
    In the spring, \textbf{[his youngest son]}, Ian, was born.
    & He was a beautiful baby boy, and he was the light of his father's life.
    & He \textbf{[was a beautiful baby]}, \textbf{[with a head of thick]}, dark \textbf{[hair and]} a round, chubby face. \\
    \midrule
    The first public library in Bintulu \textbf{[was built in]} 1971.
    & It was located at the Bintulu District Office. The library was later moved to the Bintulu Town Council building in 1973.
    & It \textbf{[was located at]} the Bintulu District Office. The library was \textbf{[later moved to the]} Bintulu Town Council building \textbf{[in 1973]}. \\
    \midrule
    Everyone considered the reviewers \textbf{[an improvement on the]} conference.
    & The reviewers were very helpful and the reviews were very thorough. The reviewers were very helpful and the reviews were very thorough.
    & The reviewers were very professional \textbf{[and the conference]} was very well organized. The conference \textbf{[was a great success]} and we are looking forward to next year's conference. \\
    \bottomrule
  \end{tabularx}
\end{table*}

\begin{table*}[t]
  \centering
  \caption{Qualitative generation results with Top-$p$ decoding. Brackets mark spans merged internally by UMIM.}
  \label{tab:qualitative-generation_top}
  \scriptsize
  \setlength{\tabcolsep}{5pt}
  \renewcommand{\arraystretch}{1.2}
  \begin{tabularx}{\textwidth}{@{}l L L L@{}}
    \toprule
    \multicolumn{4}{l}{\textbf{Prompt:} Tell me something about Lakers.} \\
    \midrule
    \textbf{Method} & \textbf{Sample 1} & \textbf{Sample 2} & \textbf{Sample 3} \\
    \midrule
    Base
    & Why did they change their name from Minneapolis Lakers to Los Angeles Lakers? What are their nicknames? What are their logos? Who are their rivals? Who are their greatest players? What are their colors? What is their motto?
    & What do you think of Kobe? Lakers are the best team in the league, I believe. I think they will win the championship again. Kobe is the best player in the league, and I think he will win the MVP this year.
    & I've heard that they are the best team in the NBA, but I'm not sure about that. Who are the best players on the team? What are their chances to win the NBA championship this season? I want to see them in action. \\
    \midrule
    UMIM
    & Tell me something about Lakers. I never \textbf{[been to a]} game. I want \textbf{[to know everything]} about \textbf{[the team and the]} players \textbf{[and the history of]} the team. I'm like \textbf{[a kid in]} \textbf{[a candy store]}. \textbf{[I just want to]} know everything.
    & What \textbf{[do you think of]} their chances \textbf{[to make the playoffs]} this year? What \textbf{[do you think of]} \textbf{[the Clippers]}? Do you think they'll be \textbf{[able to win the]} championship this year? What \textbf{[do you think of]} the Lakers' chances?
    & What is \textbf{[the most exciting thing]} about Lakers? Why do you love them? What makes \textbf{[them special]}? The Lakers \textbf{[are a very]} special team. They have had \textbf{[some of the greatest]} players \textbf{[to ever play the]} \textbf{[game of basketball]}. \\
    \bottomrule
  \end{tabularx}
\end{table*}

\subsubsection{Quantitative Results}
\label{sec:generation_quantitative}

\begin{table*}[t]
  \centering
  \caption{N-gram repetition and diversity results. Lower Rep-$n$ and higher Dist-$n$ are better. Bold indicates the better result within each decoding pair.}
  \label{tab:ngram-metrics}
  \small
  \setlength{\tabcolsep}{7pt}
  \resizebox{\textwidth}{!}{%
  \begin{tabular}{lcccccc}
    \toprule
    \textbf{Method}
    & \textbf{Rep-2} $\downarrow$
    & \textbf{Dist-2} $\uparrow$
    & \textbf{Rep-3} $\downarrow$
    & \textbf{Dist-3} $\uparrow$
    & \textbf{Rep-4} $\downarrow$
    & \textbf{Dist-4} $\uparrow$ \\
    \midrule
    Top-$p$ (Base) & 11.76 & 39.58 & 10.93 & 70.14 & 6.78 & 85.59 \\
    \rowcolor{gray!20}
    Top-$p$ (UMIM) & \textbf{11.28} & \textbf{43.49} & \textbf{9.08} & \textbf{75.59} & \textbf{4.55} & \textbf{89.92} \\
    \midrule
    Greedy (Base) & 14.03 & \textbf{40.99} & 15.32 & \textbf{60.38} & 13.97 & \textbf{69.75} \\
    \rowcolor{gray!20}
    Greedy (UMIM) & \textbf{13.34} & 39.04 & \textbf{14.44} & 58.96 & \textbf{12.74} & 69.30 \\
    \midrule
    Sampling (Base) & 11.06 & \textbf{50.72} & \textbf{7.13} & \textbf{83.10} & \textbf{2.84} & \textbf{94.68} \\
    \rowcolor{gray!20}
    Sampling (UMIM) & \textbf{10.77} & 33.96 & 12.02 & 59.34 & 9.76 & 74.25 \\
    \bottomrule
  \end{tabular}%
  }
\end{table*}

\begin{table*}[t]
  \centering
  \caption{Text-similarity results against the held-out WikiText continuation. Higher is better. Bold indicates the better result within each decoding pair.}
  \label{tab:similarity-metrics}
  \small
  \setlength{\tabcolsep}{6pt}
  \resizebox{\textwidth}{!}{%
  \begin{tabular}{lcccccc}
    \toprule
    \textbf{Method}
    & \textbf{ROUGE-2}
    & \textbf{ROUGE-L}
    & \textbf{ROUGE-Lsum}
    & \textbf{BERTScore}
    & \textbf{chrF}
    & \textbf{Mauve} \\
    \midrule
    Top-$p$ (Base) & \textbf{5.49} & \textbf{17.85} & \textbf{19.66} & \textbf{83.75} & \textbf{28.53} & \textbf{34.87} \\
    \rowcolor{gray!20}
    Top-$p$ (UMIM) & 4.87 & 17.17 & 19.14 & 83.57 & 28.39 & 33.90 \\
    \midrule
    Greedy (Base) & \textbf{5.82} & \textbf{18.89} & \textbf{20.03} & \textbf{83.35} & \textbf{26.99} & 64.28 \\
    \rowcolor{gray!20}
    Greedy (UMIM) & 5.38 & 18.44 & 19.47 & 82.98 & 26.08 & \textbf{67.21} \\
    \midrule
    Sampling (Base) & 4.12 & 16.23 & 18.24 & 83.25 & \textbf{28.28} & \textbf{36.81} \\
    \rowcolor{gray!20}
    Sampling (UMIM) & \textbf{5.45} & \textbf{18.09} & \textbf{19.58} & \textbf{83.44} & 27.67 & 27.41 \\
    \bottomrule
  \end{tabular}%
  }
\end{table*}

The results depend on the decoding strategy. Under Top-$p$ sampling, UMIM reduces repetition and improves all reported diversity metrics, while the reference-based similarity metrics remain close to the base model. Under greedy decoding, repetition is slightly reduced and the remaining metrics are generally close, with UMIM obtaining a higher Mauve score.

Under unrestricted random sampling, the results are more mixed. UMIM improves several reference-based similarity metrics, including ROUGE and BERTScore, but produces lower Dist-$n$ and Mauve scores and higher Rep-3/4. We therefore do not claim uniform preservation across every decoding distribution. Instead, these results show that the effect of compression interacts with the decoding strategy, with greedy and Top-$p$ decoding providing the most stable generation behavior in our experiments.

\subsection{Downstream Evaluation Protocols}
\label{Downstream Tasks}

\subsubsection{Multiple-Choice QA}

For PIQA, COPA, OpenBookQA, ARC-Easy, and ARC-Challenge, we follow a likelihood-based multiple-choice evaluation protocol. For each candidate answer, we concatenate the task prompt, question, and candidate into one sequence and compute its length-normalized negative log-likelihood. The candidate with the lowest loss is selected as the model prediction.

In the task-agnostic setting, every task uses the fixed merge rules mined from WikiText-103 and the WikiText-trained base merge module. No task-specific rules, SFT, or RL are used for these main QA results. The reported UMIM TR is the naturally realized reduction produced by the fixed WikiText-derived rules on each task.

\subsubsection{CNN/DailyMail Summarization}

We evaluate CNN/DailyMail using Llama-3.1-8B. The merge rules are mined from the WikiText-103 training split, and the WikiText-trained base merge module is applied directly without summarization-specific training. Generation uses greedy decoding with a fixed output length of 80 tokens.

In this experiment, token merging is applied only to the input prompt. It therefore evaluates task-agnostic prompt and prefill compression rather than decoding-time KV-cache compression.

For Selective Context, we adjust token-level prompt reduction to match UMIM's naturally realized TR. For LLMLingua-1, we use its default compression method, while LLMLingua-2 uses its small-model variant. For Sink Attention, we use four sink tokens and a sliding window equal to 16\% of the total prompt-plus-generation length. Because LLMLingua and Selective Context do not always permit exact control of the retained token count, their realized reductions may differ slightly from the target budget.

The results show that the fixed WikiText-trained UMIM module remains competitive with, and in the reported metrics outperforms, the evaluated baselines at comparable token-reduction levels on an unseen summarization domain.

\subsubsection{Long-Form Mathematical Reasoning}

We evaluate AIME2024 and AMC using DeepScaleR-1.5B-Preview and the merge module trained on the math-reasoning corpus described in Section~\ref{Datasets}. Because these tasks may generate very long reasoning traces, token merging is applied only to the input prompts. Generation is allowed to continue for up to 15,000 tokens, and performance is evaluated using pass@16.

This experiment evaluates prompt compression for long-form mathematical reasoning. It does not evaluate decoding-time merging or generated-token KV-cache reduction. Under this setting, UMIM achieves strong performance and outperforms Selective Context and LLMLingua in the reported results.

\subsubsection{Long-Context Evaluation}
\label{sec:long_context_evaluation}

Beyond the long-form mathematical reasoning evaluation with DeepScaleR, we further evaluate UMIM on RULER four-needle retrieval, multi-document question answering, and long-document summarization. These experiments examine UMIM under long-range retrieval, cross-document information integration, and long-sequence generation settings.

All experiments use Llama-3.1-8B and its base merge module trained only on 512-token WikiText-103 sequences. The backbone LLM and merge module remain frozen, without long-context SFT, RL, or other parameter updates. Because the original WikiText-derived merge rules have limited coverage on these datasets, directly applying them produces too few merges for a meaningful compression evaluation. We therefore follow the \texttt{WT + Rules} setting: high-frequency $n$-gram rules are re-mined from each evaluation corpus and paired with the frozen WikiText-trained merge module. This changes only the set of eligible merge spans; neither the merge module nor the backbone LLM is retrained.

\paragraph{RULER Four-Needle Retrieval.}

We first evaluate UMIM on RULER four-needle retrieval. This task requires the model to retrieve multiple pieces of information located at different positions in a long context, directly testing whether token merging preserves information needed for long-range retrieval. TR denotes the percentage of input tokens removed.

\begin{table}[t]
\centering
\caption{Results on RULER four-needle retrieval.}
\label{tab:ruler_long_context}
\small
\setlength{\tabcolsep}{4pt}
\begin{tabular}{ccccc}
\toprule
\textbf{Context}
& \textbf{Baseline}
& \textbf{UMIM}
& $\boldsymbol{\Delta}$
& \textbf{TR} \\
& \textbf{Acc. (\%)}
& \textbf{Acc. (\%)}
& \textbf{Acc. (pp)}
& \textbf{(\%)} \\
\midrule
1K  & 99.8  & 99.2 & -0.6  & 33.3 \\
2K  & 100.0 & 99.2 & -0.8  & 35.5 \\
4K  & 99.6  & 92.6 & -7.0  & 34.9 \\
8K  & 99.0  & 76.8 & -22.2 & 34.6 \\
16K & 96.6  & 33.8 & -62.8 & 33.7 \\
\bottomrule
\end{tabular}
\end{table}

\paragraph{Multi-Document Question Answering.}

We next evaluate whether UMIM can preserve and integrate evidence distributed across multiple documents. We use similar token-reduction ratios across context lengths to examine how performance changes as the input becomes longer.

\begin{table}[t]
\centering
\caption{Results on multi-document question answering.}
\label{tab:multidoc_long_context}
\small
\setlength{\tabcolsep}{4pt}
\begin{tabular}{ccccc}
\toprule
\textbf{Length}
& \textbf{Baseline}
& \textbf{UMIM}
& $\boldsymbol{\Delta}$
& \textbf{TR} \\
& \textbf{Acc. (\%)}
& \textbf{Acc. (\%)}
& \textbf{Acc. (pp)}
& \textbf{(\%)} \\
\midrule
2K  & 74.8 & 70.4 & -4.4  & 33.96 \\
4K  & 74.2 & 59.8 & -14.4 & 34.89 \\
8K  & 60.6 & 50.2 & -10.4 & 33.79 \\
16K & 58.0 & 19.4 & -38.6 & 34.29 \\
\bottomrule
\end{tabular}
\end{table}

\paragraph{Long-Context Summarization.}

Finally, we evaluate UMIM on GovReport and QMSum. In these experiments, merging is applied during both prompt processing and autoregressive decoding. Prompt TR denotes input-token reduction, Gen-KV TR denotes the reduction in generated-token KV-cache length at the end of generation, and Cumulative KV TR denotes the reduction in cumulative generated-token KV-cache usage throughout autoregressive decoding. All lengths are measured using the Llama-3.1-8B tokenizer.

GovReport has a mean input length of 10,275 tokens, a median of 8,712, and a range of 2,054--51,427 tokens. QMSum has a mean input length of 13,917 tokens, a median of 13,002, and a range of 2,633--30,438 tokens. We evaluate 200 examples from each dataset.

\begin{table*}[t]
\centering
\caption{Long-context summarization results. Prompt TR measures input-token reduction, while Gen-KV TR and Cumulative KV TR measure generated-token KV-cache reduction. All values are percentages.}
\label{tab:long_context_summarization}
\small
\setlength{\tabcolsep}{3.5pt}
\resizebox{\textwidth}{!}{
\begin{tabular}{llccccccccc}
\toprule
\textbf{Dataset}
& \textbf{Method}
& \textbf{Prompt TR}
& \textbf{Gen-KV TR}
& \textbf{Cumulative KV TR}
& \textbf{R-1}
& \textbf{R-2}
& \textbf{R-L}
& \textbf{R-Lsum}
& \textbf{ROUGE Avg.}
& \textbf{BERTScore} \\
\midrule
GovReport & Baseline & --    & --    & --    & 46.28 & 17.47 & 21.51 & 43.82 & 32.27 & 82.88 \\
GovReport & UMIM     & 14.50 & 12.59 & 13.04 & 46.60 & 17.18 & 21.98 & 44.05 & 32.45 & 84.75 \\
\midrule
QMSum & Baseline & --    & --    & --    & 21.29 & 6.55 & 15.07 & 18.59 & 15.37 & 82.82 \\
QMSum & UMIM     & 10.90 & 15.51 & 16.84 & 18.13 & 5.54 & 13.20 & 15.61 & 13.12 & 81.40 \\
\bottomrule
\end{tabular}
}
\end{table*}

\paragraph{Discussion.}

On RULER, UMIM remains close to the uncompressed baseline at context lengths of 1K and 2K while removing approximately one-third of the input tokens. As the context becomes longer and relevant information must be preserved over greater distances, the performance gap increases. Multi-document question answering shows a similar overall trend: UMIM retains performance relatively well at shorter lengths while achieving substantial compression, but preserving evidence becomes more difficult as context length increases.

For long-document summarization, UMIM maintains comparable overall quality on GovReport while reducing both prompt length and generated-token KV-cache usage, with slight improvements on several metrics. On QMSum, UMIM provides both prompt and decoding-time KV compression, although with some reduction in summarization quality. These results indicate that long-context tasks differ in their sensitivity to local token-span compression.

Importantly, these experiments directly transfer a base merge module trained only on 512-token WikiText-103 sequences to retrieval and question-answering contexts of up to 16K tokens and real documents containing up to 51K tokens. No long-context training, SFT, or RL is performed. The results therefore characterize how the base merge module generalizes well beyond its training length and domain, while also showing that training sequence length affects the preservation of long-range dependencies.

With sufficient training resources, long-context data should be incorporated from the beginning when training a base merge module for very long-context retrieval, multi-document question answering, long-form reasoning, or long-document summarization. Systematically training the merge module on long-context corpora with corresponding merge rules should further improve long-range performance while retaining UMIM's benefits in effective sequence length and KV-cache compression.

\subsection{Specialized-Domain Evaluation: Code-Adapted UMIM}
\label{sec:code_adapted_umim}

The main task-agnostic experiments use WikiText-derived merge rules and a WikiText-trained merge module. Code generation represents a more specialized domain with different syntax, symbolic structure, and token-frequency statistics. We therefore include a supplementary experiment to test whether the UMIM formulation remains effective when both the merge rules and merge module are constructed from in-domain code data.

We mine a code-specific merge set from the DeepMind CodeContests corpus and train a code-specific UMIM module. This is a domain-specific training experiment rather than direct transfer from the WikiText base module. Because the amount of code training data and available compute are limited, we treat this result as a supplementary validation rather than a fully optimized code-compression system.

We evaluate the code-adapted merge module on HumanEval using Llama-3.1-8B at a 25\% token-reduction level. The same compression level is used for LLMLingua-2 and Selective Context. HumanEval contains 164 tasks, and pass@1 is reported together with the number of passed and failed tasks.

\begin{table}[t]
\centering
\caption{HumanEval results on 164 tasks using Llama-3.1-8B at a 25\% token-reduction level.}
\label{tab:humaneval}
\small
\setlength{\tabcolsep}{5pt}
\begin{tabular}{lccc}
\toprule
\textbf{Method} & \textbf{pass@1 (\%)} & \textbf{Passed} & \textbf{Failed} \\
\midrule
Baseline & 62.2 & 102 & 62 \\
UMIM (code-trained) & 50.6 & 83 & 81 \\
LLMLingua-2 & 1.2 & 2 & 162 \\
Selective Context & 0.0 & 0 & 164 \\
\bottomrule
\end{tabular}
\end{table}

Despite the limited in-domain training data, code-adapted UMIM retains approximately 81\% of the original pass@1 performance, from 62.2\% to 50.6\%. Under the same token-reduction level, LLMLingua-2 and Selective Context retain substantially less performance.

This result does not imply that domain-specific training is required for every downstream task. Rather, it complements the task-agnostic experiments by showing that when a target domain has substantially different token statistics, UMIM can also be instantiated with domain-specific rules and training data. The remaining gap to the uncompressed baseline indicates that structurally sensitive code generation remains challenging and that larger in-domain training corpora or task adaptation may further improve performance.

\section{Efficiency Analysis}
\label{sec:efficiency_analysis}

\subsection{Empirical Efficiency Evaluation and Throughput Analysis}
\label{sec:efficiency}

During autoregressive decoding, each step generates one new token while attending to the accumulated KV cache. Under common small-batch and long-context inference settings, this process is often strongly affected by memory bandwidth and repeated KV-cache access. Reducing the effective sequence and KV-cache lengths can therefore reduce both attention computation and memory traffic.

\paragraph{Evaluation protocol.}

To examine whether token reduction translates into actual inference speedup, we measure prefill and decoding throughput separately, following the evaluation protocol used by Zip2Zip. All experiments use Llama-3.1-8B on a single NVIDIA A100 GPU with batch size 1 and bf16 precision.

We evaluate a controlled 50\% merge ratio while varying the prompt length. This controlled setting isolates the throughput effect of sequence-length reduction and is separate from the downstream evaluations, where TR is naturally determined by merge-rule coverage. A setting denoted by ``$256+256$'' contains a 256-token prompt followed by 256 generated tokens; the remaining settings follow the same notation. We keep the generation length fixed at 256 tokens and vary the prompt length from 256 to 2,048 tokens.

Each configuration is first executed for 10 warm-up runs, followed by 10 timed runs. GPU synchronization is performed around each timed region, and we report the mean throughput across the 10 measured runs.

Throughput is computed with respect to the original, uncompressed token count. For UMIM, prefill throughput therefore measures how many original prompt tokens are covered per second after compression, rather than the number of compressed embeddings passed to the backbone LLM. Decoding throughput measures the number of autoregressively generated tokens per second. Our current implementation is a non-fused Python-level prototype and does not use customized CUDA or Triton kernels for rule matching, merge-module execution, or KV-cache rollback.

\begin{table*}[t]
\centering
\caption{Throughput comparison for Llama-3.1-8B under a controlled 50\% merge ratio. Throughput is measured with respect to the original token count. A setting denotes prompt length $+$ generation length. Relative improvement is computed as $(\mathrm{UMIM}/\mathrm{Base}-1)\times100\%$.}
\label{tab:throughput}
\small
\setlength{\tabcolsep}{7pt}
\begin{tabular}{lcccccc}
\toprule
\multirow{2}{*}{\textbf{Setting}} &
\multicolumn{3}{c}{\textbf{Prefill Throughput (tokens/sec)}} &
\multicolumn{3}{c}{\textbf{Decode Throughput (tokens/sec)}} \\
\cmidrule(lr){2-4}
\cmidrule(lr){5-7}
& \textbf{Base} & \textbf{UMIM} & \textbf{Relative}
& \textbf{Base} & \textbf{UMIM} & \textbf{Relative} \\
\midrule
$256 + 256$  & 10,531.8 & 15,410.8 & $+46.3\%$  & 64.7 & 65.6 & $+1.3\%$ \\
$512 + 256$  & 14,210.6 & 30,978.5 & $+118.0\%$ & 65.3 & 65.9 & $+0.9\%$ \\
$1024 + 256$ & 16,138.3 & 50,516.0 & $+213.0\%$ & 65.1 & 66.4 & $+1.9\%$ \\
$2048 + 256$ & 16,592.5 & 62,124.7 & $+274.4\%$ & 64.8 & 66.4 & $+2.5\%$ \\
\bottomrule
\end{tabular}
\end{table*}

\paragraph{Results.}

UMIM improves prefill throughput across all evaluated context lengths. As the prompt length increases from 256 to 2,048 tokens, the prefill speedup grows from approximately $1.46\times$ to $3.74\times$. Prompt merging shortens the embedding sequence before it enters the backbone LLM, so every subsequent Transformer layer operates on the compressed sequence. The resulting computational benefit becomes larger as the original context length increases.

During autoregressive decoding, UMIM improves throughput by 0.9\%--2.5\% across the evaluated settings. These improvements are substantially smaller than the prefill gains but remain positive under the current non-fused implementation.

\paragraph{Interpretation.}

The difference between prefill and decoding follows from their different computational processes. During prefill, all matching prompt spans can be merged before the backbone LLM is executed. The reduced embedding sequence therefore lowers computation and memory use throughout the complete prefill pass.

During decoding, UMIM cannot skip autoregressive generation steps. After each new token is generated, UMIM checks whether the current sequence suffix matches a merge rule. If a match is found, KV-cache rollback replaces the corresponding KV entries with one merged KV state. The resulting benefit comes from reducing the effective sequence and KV-cache lengths used by subsequent decoding steps, rather than from reducing the number of generated tokens or autoregressive forward passes.

Our current decoding prototype performs suffix matching, merge-module execution, KV-cache rollback, and cache updates through Python-level control flow. These operations introduce scheduling overhead and additional small GPU kernel launches, offsetting part of the benefit from the shorter KV cache. The current results nevertheless show that the reduction in subsequent attention computation is sufficient to produce positive decoding throughput gains. They should be interpreted as evidence of initial practical feasibility rather than the performance of a fully optimized inference system.

\paragraph{Potential engineering optimizations.}

Under the current context-independent merge rules, a surrogate embedding depends only on the identities of the tokens in its span. Surrogate embeddings can therefore be precomputed and stored in a lookup table, avoiding repeated merge-module execution during inference. A more optimized implementation could further fuse suffix matching, embedding lookup, KV-cache rollback, and cache updates into CUDA or Triton kernels.

These optimizations would not change the underlying UMIM algorithm, but they require additional systems engineering and evaluation. The reported experiments are intended to establish that UMIM's reduction in effective sequence length can translate into actual inference speedup, rather than to present a fully optimized production inference system.

\paragraph{Summary.}

The throughput results confirm that UMIM reduces not only the nominal sequence length but also the realized inference cost. Prompt merging provides substantial prefill acceleration, particularly for longer contexts. Decoding-time merging cannot bypass autoregressive steps, but it continually shortens the KV cache used by subsequent attention operations and provides modest positive throughput improvements even in the current Python-level prototype.

\section{Ablation Study}
\label{sec:ablation_study}

We study three aspects of UMIM: the frequency threshold used to select merge candidates, the capacity of the merge module, and an alternative merge-rule construction strategy. Unless otherwise specified, all results are measured on held-out validation data using distribution-preservation metrics between the original and compressed models. Top-$p$ uses $p=0.9$.

\subsection{Frequency Threshold and Statistical Support}
\label{sec:threshold_ablation}

The frequency threshold $\tau$ controls which $n$-grams are included in the merge-rule set. An $n$-gram is retained only if it appears at least $\tau$ times in the training corpus. A lower threshold admits a larger and less frequently observed candidate set, generally increasing potential merge coverage. A higher threshold retains spans with stronger statistical support, allowing the merge module to observe each candidate in more training contexts.

To evaluate this trade-off, we construct merge rules exclusively from the training split and evaluate the resulting merge modules on unseen validation sequences. Table~\ref{tab:threshold_ablation} reports the results for Llama-3.2-1B and GPT-2-XL under thresholds $\tau\in\{500,50,5\}$.

\begin{table*}[t]
\centering
\caption{Validation distribution-preservation metrics under different training-corpus $n$-gram frequency thresholds. Higher values indicate closer agreement with the original uncompressed model.}
\label{tab:threshold_ablation}
\small
\setlength{\tabcolsep}{7pt}
\begin{tabular}{lcccccc}
\toprule
\textbf{Backbone}
& $\boldsymbol{\tau}$
& \textbf{Top-1}
& \textbf{Top-3}
& \textbf{Top-10}
& \textbf{Top-$p$}
& \textbf{MRR} \\
\midrule
Llama-3.2-1B & 500 & 0.8437 & 0.8356 & 0.8433 & 0.9504 & 0.9078 \\
Llama-3.2-1B & 50  & 0.7883 & 0.7840 & 0.7943 & 0.9320 & 0.8691 \\
Llama-3.2-1B & 5   & 0.7410 & 0.7414 & 0.7528 & 0.9155 & 0.8297 \\
\midrule
GPT-2-XL & 500 & 0.8414 & 0.8456 & 0.8579 & 0.9548 & 0.9048 \\
GPT-2-XL & 50  & 0.7588 & 0.7707 & 0.7893 & 0.9297 & 0.8447 \\
GPT-2-XL & 5   & 0.6955 & 0.7128 & 0.7359 & 0.9065 & 0.7928 \\
\bottomrule
\end{tabular}
\end{table*}

Across both backbones, all alignment metrics improve consistently as $\tau$ increases. For Llama-3.2-1B, Top-1 increases from 0.7410 at $\tau=5$ to 0.8437 at $\tau=500$, while MRR increases from 0.8297 to 0.9078. GPT-2-XL shows the same pattern.

These results indicate that frequency serves not only as a rule-selection criterion but also as a minimum statistical-support condition. Frequently occurring spans provide more natural contexts from which the shared merge module can learn their surrogate representations. Lower thresholds expand the candidate set and potential compression coverage, but they also introduce spans supported by fewer training examples, reducing the reliability of the learned approximation. The threshold therefore provides a practical control over the compression--alignment trade-off.

\subsection{Merge-Module Capacity}
\label{sec:merge_capacity_ablation}

We next examine whether increasing the capacity of the merge module substantially improves predictive alignment. We train one-layer and three-layer attention-pooling variants on WikiText-103 using the same merge rules with $\tau=5$. Both variants are evaluated on the same held-out validation data.

\begin{table}[t]
\centering
\caption{Merge-module capacity ablation on Llama-3.2-1B with $\tau=5$.}
\label{tab:merge_capacity_ablation}
\small
\setlength{\tabcolsep}{4pt}
\begin{tabular}{lccccc}
\toprule
\textbf{Architecture}
& \textbf{Top-1}
& \textbf{Top-3}
& \textbf{Top-10}
& \textbf{Top-$p$}
& \textbf{MRR} \\
\midrule
One attention layer
& 0.7410 & 0.7414 & 0.7528 & 0.9155 & 0.8330 \\
Three attention layers
& 0.7427 & 0.7415 & 0.7534 & 0.9158 & 0.8341 \\
\bottomrule
\end{tabular}
\end{table}

Increasing the module depth from one to three attention layers produces only marginal improvements. Top-1 increases from 0.7410 to 0.7427, while the remaining metrics change by at most approximately 0.0011. These results suggest that the shallow attention-pooling module already provides sufficient capacity for learning the surrogate embeddings in this setting. We therefore use the lighter one-layer design in the main experiments.

\subsection{Alternative Merge-Rule Construction}
\label{sec:alternative_rule_ablation}

High-frequency $n$-grams provide a simple and statistically supported way to instantiate UMIM, but frequency is not the only possible merge-rule criterion. We therefore evaluate an alternative strategy based on the language model's token-level confidence.

For each input chunk, we compute the autoregressive probability assigned by Llama-3.2-1B to every observed token. We identify consecutive spans in which all token probabilities exceed 0.5 and retain spans containing at least two tokens. To remain compatible with the current online merging implementation, candidates longer than four tokens are discarded. The resulting 2- to 4-gram spans are used as merge rules, and a merge module is trained using the same predictive-distillation procedure as in the main method.

\begin{table}[t]
\centering
\caption{Validation results using LM-confidence-based merge rules on Llama-3.2-1B. TR denotes the realized token reduction.}
\label{tab:confidence_rule_ablation}
\small
\setlength{\tabcolsep}{4pt}
\begin{tabular}{cccccc}
\toprule
\textbf{TR}
& \textbf{Top-1}
& \textbf{Top-3}
& \textbf{Top-10}
& \textbf{Top-$p$}
& \textbf{MRR} \\
\midrule
10.3\%
& 0.9107
& 0.9104
& 0.9152
& 0.9717
& 0.9496 \\
\bottomrule
\end{tabular}
\end{table}

The LM-confidence-based rules achieve strong distribution-preservation metrics at a realized TR of 10.3\%. Because this operating point has a relatively modest token reduction, the result should be interpreted as a feasibility study rather than a matched-TR comparison with frequency-based rules. It nevertheless demonstrates that UMIM is not tied to a single rule-construction heuristic. Other criteria can also provide effective merge candidates when the selected spans are sufficiently predictable and supported by the training data.

Overall, these ablations support three conclusions. First, the frequency threshold controls a practical trade-off between candidate coverage and statistical support. Second, increasing merge-module depth provides little benefit in the evaluated setting, supporting the lightweight architecture used in the main experiments. Third, alternative merge-rule criteria can also work, showing that high-frequency $n$-grams are an effective implementation of UMIM rather than a fixed limitation of the framework.

\subsection{Positional Encoding under Token-Span Merging}
\label{app:position_encoding}

\paragraph{Compressed-position convention.}
\namem{} does not modify the positional-encoding formulation or parameters of the backbone LLM, nor does it introduce position interpolation or an additional position-reindexing mechanism. A merged span occupies one effective position in the compressed sequence, and subsequent positions are assigned contiguously according to the effective sequence length.

For example, consider the original sequence
\[
[x_1,x_2,x_3,x_4]
\]
with positions
\[
[0,1,2,3].
\]
After merging \(x_2,x_3\) into a surrogate token \(x_{23}\), the input becomes
\[
[x_1,x_{23},x_4],
\]
with compressed positions
\[
[0,1,2].
\]
Thus, \namem{} does not preserve the original position indices of the merged tokens or manually compensate the positions of subsequent tokens. Training and inference consistently use this compressed-position convention.

\paragraph{Functional approximation rather than embedding equivalence.}
It is intuitive to view the surrogate token as using one token to represent the information originally carried by multiple tokens. More precisely, however, \namem{} does not require the surrogate embedding to be semantically identical to any constituent embedding or to reproduce every intermediate state of the original span. The surrogate is a learned continuous soft token whose purpose is to induce similar predictive behavior after being processed by the frozen LLM.

Let \(F_\theta\) denote the frozen language model, \(X\) and \(P\) the original token embeddings and positions, and \(\widetilde{X}_\phi\) and \(\widetilde{P}\) the compressed embeddings and positions produced using the merge module \(M_\phi\). The optimization target can be summarized as
\[
F_\theta(X,P)
\approx
F_\theta(\widetilde{X}_\phi,\widetilde{P})
\]
at aligned prediction positions. Importantly, \(\widetilde{P}\) contains the contiguous compressed positions. The merge module therefore learns to approximate the joint predictive effect of
\[
\text{span replacement}
+
\text{position compaction},
\]
rather than first reproducing the original span and then separately removing the effect of the position shift.

This explains why using the same position convention during training and inference is important. During predictive distillation, the compressed sequence is passed through the frozen LLM with the same positions that will be used at inference time. Gradients are propagated through the complete frozen computation to the merge module, allowing the surrogate embedding to compensate empirically for both replacing the span and compacting later positions. This consistency does not imply exact equivalence, but it avoids an additional train--test mismatch and directly includes position compaction in the optimization objective.

\paragraph{Why one surrogate token can still be effective.}
A \(d\)-dimensional surrogate embedding is not restricted to any discrete token in the original vocabulary. It is a learnable point in the continuous embedding space. After entering the frozen Transformer, it interacts with the preceding context, positional encoding, attention layers, and MLP layers. Consequently, although the input-side surrogate for a fixed \(n\)-gram is context-independent, its hidden representations inside the Transformer remain context-dependent.

A single surrogate KV state cannot generally reproduce every possible behavior of multiple original KV states under all future queries. However, \namem{} does not optimize for such layer-wise equivalence. Next-token prediction only requires preserving the information that materially affects the final predictive distribution. Frequent spans are observed in many training contexts, allowing the merge module to learn a surrogate that minimizes the expected predictive discrepancy over those contexts. Thus, \namem{} learns a distribution-dependent functional approximation rather than claiming an exact replacement for every possible context.

\paragraph{Structural effect under RoPE.}
For models using RoPE, position compaction does not arbitrarily perturb the relative position of every token pair. Let the effective compressed position of an original token at position \(t\) be
\[
\pi(t)=t-r(t),
\]
where \(r(t)\) is the number of token positions removed before \(t\). For two original positions \(i\) and \(j\), their compressed relative distance is
\[
\pi(j)-\pi(i)
=
(j-i)-\bigl(r(j)-r(i)\bigr).
\]
If both tokens have the same number of removed positions before them, such that
\[
r(i)=r(j),
\]
then their relative distance is unchanged:
\[
\pi(j)-\pi(i)=j-i.
\]
Therefore, token pairs within a region that does not cross a merged span preserve their original RoPE relative offsets, even though their absolute indices may shift. The relationships that change are primarily those crossing a merged span and those involving the surrogate token itself. For a pair crossing one or more merged spans, the relative distance is shortened by the number of removed positions between them.

This effect can also be characterized locally at the attention-logit level. For fixed query and key vectors, a RoPE interaction with relative distance \(\Delta\) can be written as
\[
\ell_{ij}=q_i^\top R_{\Delta}k_j,
\]
where \(R_{\Delta}\) is the corresponding RoPE rotation. If compression shortens the relative distance by \(\delta\), the positional contribution becomes
\[
\widetilde{\ell}_{ij}
=
q_i^\top R_{\Delta-\delta}k_j.
\]
The resulting difference satisfies
\[
\left|
\ell_{ij}-\widetilde{\ell}_{ij}
\right|
\leq
\lVert q_i\rVert_2
\lVert k_j\rVert_2
\left\|
R_{\Delta}-R_{\Delta-\delta}
\right\|_2.
\]
For the blockwise rotations used by RoPE,
\[
\left\|
R_{\Delta}-R_{\Delta-\delta}
\right\|_2
=
2\max_m
\left|
\sin\left(\frac{\delta\omega_m}{2}\right)
\right|,
\]
where \(\omega_m\) is the angular frequency of the \(m\)-th RoPE component. This shows that the positional change is a structured perturbation controlled by the amount of distance compression and the RoPE frequencies.

This local relation is not an end-to-end error bound. In a multi-layer Transformer, the queries, keys, values, and hidden states also change after merging. Nevertheless, it explains why compressed positions do not destroy all relative-position structure: within-region offsets remain unchanged, while cross-span distances are shortened in a systematic manner.

\paragraph{Learned absolute positional embeddings.}
For models using learned absolute positional embeddings, such as GPT-2-XL, the RoPE relative-offset property above does not apply. After a merge, subsequent tokens receive different absolute positional embeddings. \namem{} does not assume that this difference disappears. Instead, the merge module is trained under the same compressed-position convention used at inference time, allowing predictive distillation to jointly account for surrogate replacement and absolute-position shifting.

The effectiveness of \namem{} on both RoPE-based Llama models and GPT-2-XL provides empirical evidence that the method is not tied to one positional-encoding architecture. However, these results demonstrate the viability of the complete method under different positional mechanisms; they do not imply that position shifts have no effect.

\paragraph{Positions during autoregressive decoding.}
The same convention is used during autoregressive decoding. When a newly generated token completes a mergeable suffix, \namem{} rolls back the KV-cache entries corresponding to that span and replaces them with one surrogate state. Subsequent positions are assigned according to the updated effective cache length. The physical KV-cache length therefore becomes shorter than the number of tokens generated in the original uncompressed token stream.

This design allows \namem{} to retain the standard positional computation of the frozen backbone without modifying its positional-encoding architecture. At the same time, it shortens relative distances across previously merged spans. The merge module is trained to compensate for this effect empirically, but exact compensation is not guaranteed for every context or sequence length.

\paragraph{Empirical evidence and scope.}
All \namem{} results reported in this paper use the compressed-position convention without any additional positional adaptation. The language-modeling, QA, summarization, and generation results therefore provide end-to-end evidence that the frozen LLM can effectively use the learned surrogate embeddings despite position compaction. The qualitative generations and attention analysis further show that merged representations remain actively involved in the model computation rather than being ignored. Moreover, the task-adaptation experiments show that, while keeping the backbone completely frozen, adapting only the small merge module can further reduce the predictive discrepancy caused by compression.

These experiments do not separately identify how much error comes from surrogate replacement and how much comes from position compaction, since the two effects are jointly optimized in the current design. They nevertheless demonstrate that, without changing the backbone positional mechanism, \namem{} can achieve meaningful compression while retaining strong predictive performance across multiple models, tasks, and positional-encoding architectures.

We do not claim positional invariance or exact internal equivalence between the compressed and uncompressed computations. Preserving the original position indices is a meaningful alternative design, but a controlled comparison would require separately training
\[
\text{compressed-position training}
+
\text{compressed-position inference}
\]
and
\[
\text{preserved-position training}
+
\text{preserved-position inference}.
\]
For dynamic decoding, the preserved-position variant would also need to maintain both the logical position in the original token stream and the physical storage position in the compressed KV cache, together with corresponding changes to generation, cache rollback, attention masking, and position bookkeeping. We leave the systematic implementation and evaluation of this alternative positional design to future work.

\newpage

\end{document}